\documentclass[twocolumn,11pt]{article}
\usepackage{apacite}
\usepackage[utf8]{inputenc}
\DeclareUnicodeCharacter{00A0}{}
\DeclareUnicodeCharacter{00A0}{~}
\usepackage{lmodern}
\usepackage{mathtools, cuted}
\usepackage{titlesec}
\usepackage{blindtext}
\usepackage{rotating}
\usepackage{varwidth} 
\newlength\titleindent
\usepackage[top=2cm, bottom=3cm, left=1.5cm, right=1.5cm]{geometry}

\usepackage{import}
\usepackage{graphicx}
\usepackage{color}
\usepackage[table,xcdraw,svgnames]{xcolor}
\usepackage{tabularx,booktabs}
\usepackage{pgfplots}
\usepackage{pgfplotstable}
\usepackage{subcaption}
\usepackage{graphicx}
\usepackage{multirow}
\usepackage{mathtools}
\usepackage{algorithm, algorithmic}
\usepackage{makeidx}
\usepackage{soul}
\usepackage{xcolor}
\usepackage{amsthm,amssymb,amsmath}
\usepackage{pifont}

\usepackage{tabularx}
\usepackage{lscape}
\usepackage{pdflscape}
\usepackage{multirow}
\usepackage{tabulary}
\usepackage{supertabular,booktabs}
\usepackage{ragged2e}
\newcolumntype{L}[1]{>{\raggedright\arraybackslash}p{#1}}
\newcolumntype{C}[1]{>{\centering\arraybackslash}p{#1}}
\newcolumntype{R}[1]{>{\raggedleft\arraybackslash}p{#1}}
\usepackage{fourier, heuristica}
\usepackage{array, booktabs}
\usepackage{graphicx}
\usepackage{amsmath}

\usepackage[font={small,it}]{caption}
\usepackage{ltablex}
\DeclareCaptionFont{gray}{\color{gray}}
\usepackage{natbib}

\usepackage{amsmath}

\usepackage{amssymb}
\usepackage{latexsym}

\usepackage{url}
\AtBeginDocument{%
  \hypersetup{
    citecolor=blue!60!black,
    linkcolor=blue!60!black,   
    urlcolor=blue!60!black}}

\usepackage{tikz}
\usetikzlibrary{arrows.meta}
\usetikzlibrary{through}
\usetikzlibrary{matrix}
\usetikzlibrary{decorations.pathreplacing}
\tikzset{%
	>={Latex[width=2mm,length=2mm]},
	base/.style = {rectangle, rounded corners, draw=black,
		minimum width=4cm, minimum height=1cm,
		text centered, font=\sffamily},
	new/.style = {rectangle, draw=black,
		minimum width=4cm, minimum height=1cm,
		text centered, font=\sffamily},
	activityStarts/.style = {base, fill=cyan!90!blue},
	startstop/.style = {base, fill=red!40},
	activityRuns/.style = {base, fill=green!70!black},
	rec/.style = {new, fill=white},
	process/.style = {base, minimum width=2.5cm, fill=orange!15,
		font=\ttfamily},
	data/.style = {base, minimum width=2.5cm, fill=cyan!80!blue,
		font=\ttfamily},
}
\usetikzlibrary{mindmap,shadows}

\usepackage[hidelinks, pdfencoding=auto,colorlinks]{hyperref}
\title{
{\LARGE \bfseries FakeNews: GAN-based generation of realistic 3D volumetric data \\ A systematic review and taxonomy
}
}

\author{André Ferreira$^{1,2,3,8,9}$ \and Jianning Li$^{2,3,4}$ \and Kelsey L. Pomykala$^{3}$ \and Jens Kleesiek$^{3,4,5,7}$ \and Victor Alves$^1$ \and Jan Egger$^{2,3,4,6}$ }

\date{$^1$\small{Center Algoritmi, University of Minho, Braga, Portugal} \\
$^2$\small{Computer Algorithms for Medicine Laboratory, Graz, Austria} \\
$^3$\small{Institute for AI in Medicine (IKIM), University Medicine Essen, Girardetstraße 2, 45131 Essen, Germany} \\
$^4$\small{Cancer Research Center Cologne Essen (CCCE), University Medicine Essen, Hufelandstraße 55, 45147 Essen, Germany} \\
$^5$\small{German Cancer Consortium (DKTK), Partner Site Essen, Hufelandstraße 55, 45147 Essen, Germany} \\
$^6$\small{Institute of Computer Graphics and Vision, Graz University of Technology, Inffeldgasse 16, 8010 Graz, Austria} \\
$^7$\small{TU Dortmund University, Department of Physics, Otto-Hahn-Straße 4, 44227 Dortmund, Germany}\\
$^8$\small{Department of Oral and Maxillofacial Surgery, University Hospital RWTH Aachen, 52074 Aachen, Germany}\\
$^9$\small{Institute of Medical Informatics, University Hospital RWTH Aachen, 52074 Aachen, Germany}\\
}

\begin{document}
\twocolumn[\maketitle \begin{@twocolumnfalse} 
\hrule
\begin{abstract}
\large
With the massive proliferation of data-driven algorithms, such as deep learning-based approaches, the availability of high-quality data is of great interest. Volumetric data is very important in medicine, as it ranges from disease diagnoses to therapy monitoring. When the dataset is sufficient, models can be trained to help doctors with these tasks. Unfortunately, there are scenarios where large amounts of data is unavailable. For example, rare diseases and privacy issues can lead to restricted data availability. In non-medical fields, the high cost of obtaining enough high-quality data can also be a concern. A solution to these problems can be the generation of realistic synthetic data using Generative Adversarial Networks (GANs). The existence of these mechanisms is a good asset, especially in healthcare, as the data must be of good quality, realistic, and without privacy issues. Therefore, most of the publications on volumetric GANs are within the medical domain. In this review, we provide a summary of works that generate realistic volumetric synthetic data using GANs. We therefore outline GAN-based methods in these areas with common architectures, loss functions and evaluation metrics, including their advantages and disadvantages. We present a novel taxonomy, evaluations, challenges, and research opportunities to provide a holistic overview of the current state of volumetric GANs.\\ \leavevmode 
\hrule
\end{abstract}
\end{@twocolumnfalse}]

\section{Introduction}\label{sec1}
In this systematic review, we survey works that generate realistic synthetic 3D volumetric data with Generative Adversarial Networks (GANs) \citep{Goodfellow2014}. With the massive increase of data-driven algorithms, such as deep learning-based approaches, during the last years \citep{egger2021deep, egger2022medical}, data is of great interest. In this context, high-quality training, validation and testing datasets are required. Unfortunately, there are scenarios and applications where large amounts of these data are unavailable. Examples can come from the medical domain, with rare diseases, leading to an insufficient amount of initial training data. Moreover, additionally in the medical field, when dealing with real patient data, privacy issues can also limit the amount of available data. This problem does not only affect the medical field, as the cost of obtaining high-quality labelled data is very high in many other fields, such as object recognition and the study of porous media \citep{Mosser2017, Muzahid202120}. A solution to this problem can be the generation of synthetic data to perform data augmentation, along with additional novel data augmentation mechanisms \citep{shorten2019survey}.
Therefore, we outline GAN-based methods in this area with common architectures, loss functions and evaluation metrics, pros, cons, challenges, research opportunities for a holistic overview of the state-of-the-art, and we also present a novel taxonomy. Throughout this review, the term "realistic" will be used very frequently. This term means that it looks like real data, i.e. that it is capable of fooling experts in the field, and that it looks realistic enough to be used as real data. This type of data is in high demand, as data that mimics reality is needed, especially in the medical field. For brevity, "3D" is also used as a substitute for "volumetric" when not otherwise stated.

The amount of volumetric data has been increasing, as this type of data is important to represent volumetric object in an accurate way not achievable by 2D image representation. In medical imaging analysis, 3D data has shown to be essential for patient motoring, disease detection and treatment, drug research, and many more. Radiologists usually analyse these data slice by slice, as it is believed that experienced radiologists are able to mentally visualise the volume. However, this might bring inconsistencies among radiologist interpretations. On the other hand, the visualisation of the whole volume at once provides an overview which makes it easier to understand unfamiliar shapes. The 2D slice inspection might be more beneficial for disease detection, however, physicians and other professionals that do not have the same expertise as radiologists highly benefit from the volumetric visualisation of such data for treatment planning and better spatial understanding \citep{preim2007visualization}.

With enough volumetric data, it might be possible to, for example, train models for augmented realities to real visualisation of patient's lesions in the correct space, or to see the placement of simulated objects in the real world in the most organised way possible for organisational efficiency, or for construction purposes. 

\subsection{Manuscript Outline}

\label{sec:Manuscript_Outline}
We present a systematic review on the use of GANs for the generation of volumetric data, which includes general methods such as denoising, reconstruction, segmentation, classification, and image translation, as well as specific applications such as nuclei counting. This section presents the manuscript outline, the search strategy, the research questions, an insight of the background on volumetric data generation, and the types of 3D data representation. The rest of this review is organized as follows.

Section \ref{sec:GANs} provides a brief explanation of GAN architecture, main purposes, advantages, and disadvantages. This section is intended to provide an easy-to-understand insight into GANs, how they work, and also to describe possible applications.

Section \ref{sec:Generating_realistic_synthetic_3D_data:_a_review_of_works} provides an overview of the works done on generating realistic 3D data with GANs and a description of the main information and statistics related to modalities, application and metrics used. 
In sections \ref{sec:Loss_Functions} and \ref{sec:Evaluation_Metrics}, summary and insights
are provided about each group of loss functions and evaluation metrics used in the papers, with more detailed explanations in the Appendices \ref{app:loss_functions} and \ref{app:sec_Evaluation_Metrics}, with relevant references if more in-depth knowledge is required.

Section \ref{sec:3D_data_generation} presents the works that are considered relevant, i.e. when the output of the networks are synthetic volumetric data generated by GANs or when the generation of volumetric data is crucial for the downstream task. 
All these papers are summarized in Tables \ref{tab:table2}, \ref{tab:table3} and \ref{tab:table4} with respect to \textit{Modality}, \textit{Medical}, \textit{Dataset}, \textit{Network}, \textit{Loss function}, \textit{Evaluation Metric}, and \textit{Comments}. Wherever possible, these tables contain references to the datasets used as well as references to lesser known concepts and architectures. All acronyms and abbreviations, if used more than once, appear in section \ref{subsub:Acronyms_and_Abbreviations}. If used only in a particular table, the acronyms appear before that table.

Section \ref{sec:visual_results}  contains a closer look at relevant work from various fields that the reader should explore in more detail. Section \ref{sec:Conclusion_and_Discussion} provides a general discussion of the current state of use of GANs, the main problems and possible solutions, tendencies in volumetric data generation with GANs, conclusions that emerge from the review, and research opportunities that researchers could and/or should take. In section \ref{sec:Applications} we discuss the applications in the referenced papers, divided into medical and non-medical applications (Tables \ref{tab:table5} and \ref{tab:table6}). The material in the Appendix is also important for a deeper understanding of the review, especially for less experienced readers. It is therefore recommended to access it while reading the main manuscript.

To clarify, it should be noted that this is an overview focusing mainly on the use of GANs to generate volumetric data. The reason for this choice is the will to improve 3D data generation with GANs, an underdeveloped topic with great potential but which still needs further development. The target audience of this review is researchers who want to enter the field of volumetric synthetic data generation, with or without much experience with GANs.

\subsubsection{\textbf{Search Strategy}}
We performed a search in the IEEE Xplore Digital Library, Scopus, PubMed, and Web of Science with the search query \textit{\lq (("Generative Adversarial Network" OR "Generative Adversarial Networks" OR gan OR gans) AND (generation OR generative) AND (3d OR three-dimensional OR volumetric) AND data AND synthetic)\rq} to find specific papers on the use of GANs for volumetric data generation. Since GANs were presented in 2014 by \cite{Goodfellow2014}, all papers prior to 2014 were excluded.

During the search we found $317$ non-unique records, of which $161$ were duplicated and $1$ was published before 2014 shown by PubMed in relation to three-dimensional multicellular tumour spheroids, leaving $155$ remaining papers. 
Based on the titles and abstracts, we excluded $82$ records that did not mention volumetric generation with GANs.
We assessed the resulting $73$ different papers and excluded $1$ of them, which was a review paper on deep learning in pore imaging and modelling.

\begin{figure}[!t]
\centering
\includegraphics[width=0.8\columnwidth]{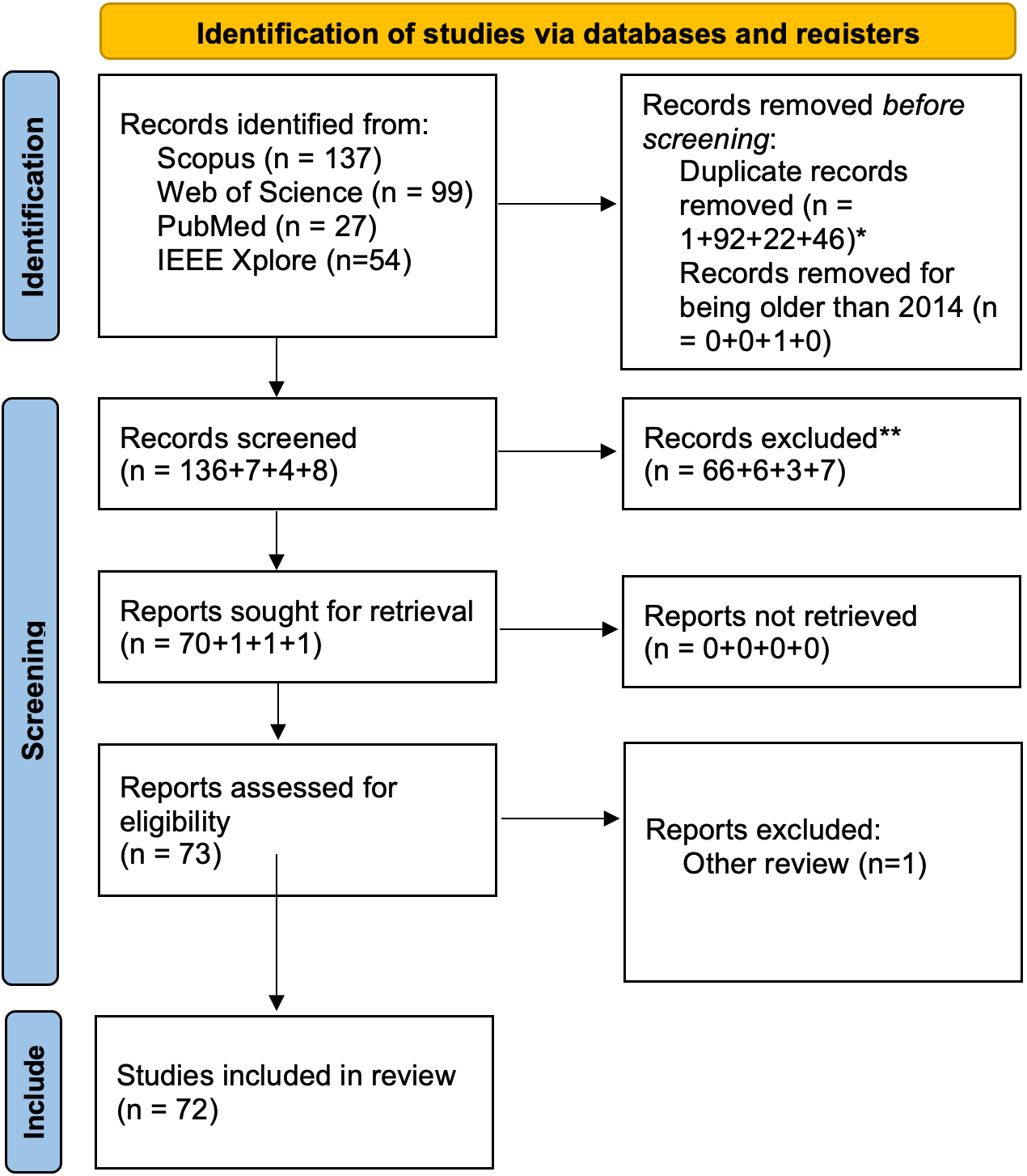}
\caption{PRISMA diagram. *Removed using an automation tool (python script).
** Excluded because they did not involve the generation of volumetric data.}
\label{prisma}
\end{figure}

After further reading, it turned out that $14$ articles did not actually use volumetric data. This resulted in a total of $58$ \textit{core papers} about generation of volumetric data using GANs, which will be covered in depth in our review. To the best of our knowledge, this is the first review that provides a detailed analysis of the published papers on the use of GANs for the generation of volumetric data.  The PRISMA diagram in Figure \ref{prisma} provides a summary overview of our screening.

Note that we include all published research, not just medical applications, which is beneficial for readers from all fields who want an overview of the potential applications of volumetric data generation using GANs. We distinguish between medical and non-medical applications, which makes it easier for the reader to focus on the desired field.

\subsubsection{\textbf{Research Questions}}
The overall aim of this systematic review is to analyse works published between 2014 and January 2022 on the generation of volumetric data with GANs.
In this regard, we defined the following main research questions for our study:
1) What are the different applications of GANs in the generation of volumetric data?
2) What are the methods most frequently or successfully employed by GANs in the generation of volumetric data?
3) What are the strengths and limitations of these methods?
4) What improvements are sought through the use of this technology?

\subsubsection{\textbf{Acronyms and Abbreviations}}
\label{subsub:Acronyms_and_Abbreviations}
The following list shows the abbreviations that are used more than once throughout the review. Other abbreviations are defined directly before each table, when used only once.
\begin{itemize}
\setlength\itemsep{-0.5em}
\item	Acc — Accuracy;
\item   ADNI — Alzheimer’s Disease Neuroimaging Initiative;
\item   Adv — Adversarial loss;
\item	AUC — Area Under the Curve;
\item	CAD — Computer-Aided Design;
\item	CBCT — Cone-Beam Computed Tomography;
\item   CE — Cross-Entropy;
\item   cGAN — conditional GAN;
\item   CT — Computed Tomography;
\item   DCGAN — Deep Convolutional Generative Adversarial Networks;
\item	DSC — Dice Similarity Coefficient;
\item   ED-GAN — Encoder-Decoder GAN;
\item	FID — Fréchet Inception Distance;
\item   HD — Hausdorff Distance;
\item   HU — Hounsfield Unit;
\item	IoU — Intersection-over-Union;
\item   KL — Kullback–Leibler;
\item   LiDAR — Light Detection And Ranging; 
\item	LIDC — Lung Image Database Consortium;
\item   LSGAN — Least Squares GAN;
\item	MAE — Mean Absolute Error;
\item   Minkowski functional — Porosity, specific surface area, average width, Euler number, Permeability;
\item   MRI — Magnetic Resonance Imaging;
\item	MSE — Mean Squared Error;
\item	NCC — Normalized Correlation Coefficient;
\item   NMSE — Normalized Mean Squared Error; 
\item   PC — Point Cloud;
\item   PET — Positron Emission Tomography;
\item   PGGAN — Progressive Growing GANs;
\item   Pre — Precision;
\item   PSNR — Peak Signal-to-Noise Ratio;
\item   RGB-D — Red, Green, Blue image with Depth;
\item   SEM — Scanning Electron Microscope;
\item	Sen — Sensitivity;
\item	Spe — Specificity;
\item   SSIM — Structural Similarity Index Measure;
\item   VTT — Visual Turing Test;
\item   WGAN — Wasserstein GAN;
\item   WGAN-GP — Wasserstein GAN with Gradient Penalty;
\end{itemize}

\subsection{Background on volumetric data generation}

Synthetic data are artificially generated by applying a sampling procedure to real data or by simulation. They must be realistic enough, but still different from the real data, i.e. they do not come directly from the real world. Synthetic data are created when the amount of available real data is insufficient for the task at hand, or they are unbalanced or incomplete. In particular, they are used when it is impossible or too expensive to obtain further data or when the real data is protected by data protection regulations.

Machine and deep learning solutions have gained prominence over the last decade. These technologies are state-of-the-art in a variety of image processing tasks, ranging from medical to non-medical applications. However, large datasets are needed to achieve good performance. In the case of volumetric data, the costs associated with their acquisition and processing lead to an intense search for ways to generate them synthetically.

Computer-aided design (CAD) has been used to create synthetic volumes as it allows the simulation of a wide range of real volumetric objects, e.g. to simulate material properties, to train segmentation and classification models, to be used in virtual or augmented realities and much more. This type of synthetic data can be created completely in a virtual mode, e.g. the creation of objects, virtual scenes and virtual worlds, which can be done manually when no other strategy is possible, or based on real-world data, i.e. point clouds or voxel grids  \citep{man2022review}. \cite{kohtala2021leveraging} trains an object recognition model using synthetic objects created with CAD. \cite{wong2019synthetic} create a computer vision system to recognise supermarket products in a warehouse environment using synthetic data from CAD. \cite{marcu2018safeuav} use a synthetic 3D aerial dataset created from 3D meshes to train a model to estimate depth and safe landing areas for unmanned aerial vehicles. These techniques can be used to capture multiple 2D images from different angles of objects or the entire 3D object.

Although there are not many works that use statistical shape models (SSM) or principal component analysis (PCA) to generate synthetic volumetric data, they may also be an option. \cite{li2022back} use SSM for synthetic correction of skull defects, and \cite{heimann2009statistical} give an overview of papers using SSM in medical image segmentation. \cite{blanz1999morphable} develop 3D morphable models based on SSM to generate facial shapes.
\cite{yu2021pca} apply PCA to model the shape of healthy human skulls and to synthetically correct defective skulls.

Autoencoders and variational autoencoders are one of the first deep learning architectures that have had a real impact on the generation of synthetic volumetric data in both medical and non-medical applications. Autoencoders and variational autoencoders have been used for various tasks, such as denoising \citep{kascenas2022denoising}, feature extraction \citep{huang2022biomarkers} and image compression \citep{tudosiu2020neuromorphologicaly}, and also in synthetic volume generation, e.g., \cite{zhang20193d} use a variational autoencoder approach for 3D shape synthesis,  more specifically for aircraft models, and \cite{saha2020quantifying} generate 3D car shapes from point clouds.

GANs are an improvement over existing deep learning architectures for generating synthetic data, as this technique can achieve greater realism and diversity compared to the other approaches. Recently, diffusion models have been able to outperform GANs in image synthesis \citep{dhariwal2021diffusion}. They are GANs' biggest competitor in image generation, and their research has grown exponentially \citep{croitoru2022diffusion}. However, GANs are distinct enough from diffusion models, and none of them can completely replace the other. Therefore, a systematic review of the existing methodologies for generating synthetic volumes using GANs is presented. It is important to note that some of the papers presented in this review use some of the above methods to obtain the dataset for training the GAN, e.g., \cite{greminger2020generative, Kniaz20203, Yang2017679}, as the use of one method does not preclude the other. 

Although synthetic volumetric data have many applications, their generation presents some challenges. Synthetic data aim to overcome several problems previously mentioned, such as unbalanced datasets, the time and cost of obtaining real data, or even the impossibility of obtaining such data. However, some synthetic data must be created manually because it cannot be done otherwise, which requires time, effort and skilled labour, and is therefore expensive and inefficient. Rendering volumetric data is also very computationally intensive, and processing this type of data requires significant memory and computing power. In addition, the generation of volumetric data is still very under-researched compared to the generation of 2D images, resulting in a lack of conventional metrics for assessing the quality of synthetic data and a complete pipeline for generating this type of data. The lack of large datasets such as ImageNet \citep{russakovsky2015imagenet} also delays research and development of approaches to generate volumetric synthetic data. Therefore, this review aims to provide an overview of works on the generation of synthetic volumetric data using a promising technology, GANs. A further discussion of the main problems and solutions can be found in \ref{sec:Conclusion_and_Discussion}.

\subsection{Types of 3D data representation}
\label{sub:Types_of_3D_data_representation}

3D geometry data are usually divided into three main groups: Point clouds, meshes and voxel grids. 

Meshes are representations of 3D objects using polygons, e.g. triangles or quadrilaterals, that form a mesh of faces in a 3D space (X, Y and Z axes). Each polygon consists of vertices, edges and an orientation vector connected to its immediate neighbour (without overlap) to form objects. This type of representation allows for fast processing as simple shapes are used to represent complex objects. Usually, meshes are obtained from point clouds after they have been processed by computer software, or they can be created manually using CAD software, but this is tedious and sometimes ineffective for some applications involving complex objects. Meshes can also be created using voxel grids through the use of appropriate software.

Point clouds are data points in a three-dimensional space, i.e. measurement points in the X, Y and Z axes. Each individual point represents a spatial measurement on the surface of the object. To represent an object, multiple points are acquired. 
Point clouds are permutation invariant, unlike voxel grids. They may or may not contain RGB, if so they also contain information about the colour of the object. They can also contain intensity information representing the strength of the reflection of the laser pulse. These points are created with special tools, namely laser scanners. The best known laser scanner is the  Light Detection and Ranging (LiDAR) sensor, which uses rapid laser pulses to measure multiple distances between the sensor and surfaces. These sensors provide an accurate representation of real world space, surfaces and objects, making this data suitable for examining objects in the real world. The denser the points, the more detailed the representation, allowing the study of textures or other smaller features. This technology can be used to represent small objects such as chairs or manufacturing parts, or larger objects such as historical monuments or entire representations of urban environments. It can also be used for autonomous vehicles by collecting multiple distances/point clouds that serve as input for machine and deep learning algorithms, allowing the vehicle to make fast decisions.

However, such data cannot be used as input for convolutional networks because they are irregular graph data. An example of regular data are voxel grids. Voxel grids are 3D grids organised in layers, rows and columns. Each intersection between a layer, row and column is called a voxel, which is assigned an intensity value. Voxel grids can be thought of as fixed-size point clouds, where each voxel has a fixed size and discrete coordinates, but point clouds can have an infinite number of points for each space. Voxels are often used to represent medical imaging, such as MRI, CT and other modalities.

Point clouds can be converted into 3D CAD models through a process called surface reconstruction \citep{berger2017survey}, or even used to create meshes or voxel grids. Point clouds can be used to represent volumetric data, such as in medical imaging, for multi-sampling and data compression, as point clouds are more memory efficient than voxel grids \citep{sitek2006tomographic}. Any data format can be converted to another, but information is always lost when converting point clouds to another format, as usually not all points are represented during the conversion, e.g. when converting a point cloud to a voxel grid.

Building computer vision pipelines for 3D data is not a mere extension of traditional deep learning techniques that work perfectly in 2D. 3D datasets are more complex, which leads to higher algorithm complexity, more instability and higher computational capacity requirements. Therefore, traditional tasks such as object recognition and segmentation are more challenging when using 3D data. Some works try to circumvent the volumetric aspect of the data by using multiple 2D views of the object as input \citep{su2015multi}. However, important information is lost, such as depth information, and multiple frames of the same object are not able to truly represent the object, because the network processes them as individual pieces of information rather than collective information. Also, using the depth information as an additional channel (in RGB-D images) does not represent the entire object under investigation, limiting the amount of information that can be fed to the algorithm.

The use of voxel grids can bridge this gap between 2D and 3D vision and enables the adaptation of some 2D image processing concepts to 3D. \cite{maturana2015voxnet} was one of the first works to use deep learning on voxel grids constructed from point clouds. With such a representation, the use of 3D convolutions is possible, and it can be processed more easily and efficiently than point clouds. Voxel grids are richer in information compared to point clouds in some situations. The representation of voxel grids is valuable for the detection of high level features such as shapes and whole objects, which is more difficult to achieve with point clouds.

Learning directly from point clouds requires the use of specialized networks, such as PointNet \citep{qi2017pointnet}. Since point clouds are simply an individual set of points represented in a 3D space, the order of input should be irrelevant to the network as it does not affect the geometry of the object. Therefore, multi-layer perceptron is used instead of convolutions. However, such a representation does not necessarily ensure that the network learns dependencies between points in the neighbourhood, which is easily captured by convolutions.  Due to the complexity of using point clouds directly as input to deep learning algorithms, this is still an under-researched topic, making it easier for researchers to convert point clouds into voxel grids and use better developed computer vision mechanisms \citep{gutierrez2021discriminative}. New research areas are being developed to address the complexity and problems of existing point cloud processing algorithms \citep{GarimellaNaidu2018Gradient}.

In GANs, point clouds are usually converted into voxel grids before being fed into the GAN. The convolution architecture requires regular inputs to function properly. Since meshes and point clouds are not equivalent to regular voxel grids, most researchers choose to convert the data into 3D grid-like structures or a collection of multiple views (2D images), e.g. \cite{Li20195530}. As mentioned earlier, using a 2D view of 3D data is suboptimal and memory inefficient. Conversion to 3D voxel grids is also memory inefficient and imposes spatial constraints on each point in the point cloud. However, such a constraint can be beneficial for some tasks, e.g. brain tumour segmentation, where the connectivity between voxels in the neighbourhood is essential for accurate segmentation as it is very likely that a tumour cell is adjacent to another tumour cell. Volumetric information and higher level features, such as whole objects detection, are also beneficial for some tasks, such as classification. In this review, several works were found that convert meshes and point clouds into voxels, e.g. \cite{Yang2017679, Kniaz20203, Nozawa2021}.

However, some proposals have been made to use point clouds as input to the network. Point cloud GAN (PC-GAN \citep{li2018point}) is a GAN architecture capable of processing point clouds without the need for voxelisation. The point cloud of an object ($\theta$) is a set of $n$ dimensional vectors $X=\{x_1, x_2, ..., x_n\}$ with $x_i\in \mathbb{R}^d$ where $d=3$ and $n\in \mathbb{Z}^+$. The corresponding point clouds of $M$ objects is then $X^{(M)}$. The generative model is defined as $p(X)$, which must be able to generate new sets $X$ and generate new points for a giving set, i.e., $x\sim p(x|X)$. Therefore, joint likelihood can be expressed by equation \ref{eq:PC-GAN_join_like}.

\begin{equation} \label{eq:PC-GAN_join_like}
\begin{split}
p(X,\theta)=p(\theta)\Pi_{i=1}^n p(x_i|\theta)
\end{split}
\end{equation}
where $p(\theta)$ is the object and $\Pi_{i=1}^n p(x_i|\theta)$ is the points for the object.

Existent generative models such as GANs only work with datasets that are a set of fixed dimensional instances, whereas point clouds are a set of sets. Using traditional GANs to learn a marginal distribution $p(x)$, where $X$ is point clouds, is unrealistic since the marginal distribution is uninformative for such algorithms.

\cite{li2018point} purpose the use of a generator that takes as input the noise vector $z$ and a descriptor encoding $\psi$ of the distribution of $\theta$, and defines the objective function of the GAN by equation \ref{eq:PC-GAN}.

\begin{equation} \label{eq:PC-GAN}
\begin{split}
\mathbb{E}_{\theta \sim p(\theta)}[min_{G_x,Q}D(\mathbb{P}||\mathbb{G})]
\end{split}
\end{equation}
where $\mathbb{P}$ is $p(X|\theta)$, $\mathbb{G}$ is $G_x(z,Q(X))$, where $Q(X)$ is an inference network to learn the informative description $\psi$ about the distribution $p(\theta |X)$.
However, the size of $X$ and the permutation of the points is variable, which increases the complexity of this problem. Furthermore, the PC-GAN also does not take into account the shape of the objects or the relationship between points in the same neighbourhood, which makes such a solution suboptimal compared to the use of voxel grids.

\cite{sulakhe2022crangan} is an example of a successful attempt to use point clouds as input for GANs. They have developed a solution for reconstructing skulls with GANs and point clouds instead of using 3D meshes and CAD software. In their experiments, for each surface (each 3D mesh of the ROI of the skull part to be reconstructed), m points are sampled, resulting in a point cloud of m points, where m=1024. Specifying a certain number of points per point cloud allows for easier use of conventional GAN approaches. The proposed CranGAN is an autoencoder based architecture conditioned by the defective skull $P_{in}$, where the goal is to train a generator: $P_{out}=G(P_{in})$. The encoder is based on the PointNet \citep{qi2017pointnet}, where the classifier head is replaced by a 256-dimensional embedding and the decoder consists of fully connected layers. The discriminator is adapted from the PC-GAN \citep{li2018point}, i.e. it consists of fully connected layers that classify the data as real or fake. The objective function is similar to equation \ref{eq:GAN}, where $z$ is replaced by $P_{in}$. In contrast to PC-GAN, the vanilla GAN objective as well as LSGAN and WGAN-GP were tested. The results show that using the vanilla GAN objective leads to better results compared to the other approaches. 

\cite{gutierrez2021discriminative} develops another framework that works directly with point clouds for anatomical shape analysis. They point out that using point clouds is more lightweight and simple than using meshes. Both the generator and discriminator are adopted from PointNet \citep{qi2017pointnet} to encode the point clouds. The generative network is a conditional encoder-decoder architecture. Their approach takes into account that a shape represented by a point cloud must be invariant to transformations such as scaling, translation and rotation. To this end, the data is preprocessed to be centred by its centre of mass and a network is trained to align the point cloud ($P_{raw}$) by rotation: $P_{raw}\rightarrow \theta$ that $P=T(\theta)P_{raw}$. The framework is then composed of: 1) a rotation network; 2) a conditional generator conditioned by the rotated point cloud; 3) a discriminator that distinguishes between the generated and the real data. With their approach, multiple structures can be used as input.
To test the framework, the ADNI dataset \citep{jack2008alzheimer} was used by converting the MRI scans into meshes and then uniformly sampling points to generate point clouds. The framework is capable of generating annotated point cloud data for various tasks such as classification, regression, among others.

\cite{ben2018multi} developed a GAN that works with meshes and generates human body and tooth meshes. The proposed architecture is based on \cite{karras2017progressive}, with the number of channels adapted to the mesh data, i.e. $k\times k\times 3|F|$ where $k$ is the grid size, $3$ the number of coordinates ($x$, $y$, $z$) and $F$ the face of each triangulation of the mesh, and the convolutions are replaced by periodic convolutions, producing spherical surfaces. With this model it is possible to automatically generate massively plausible random models.

In summary, the use of point clouds and meshes as direct input to deep learning algorithms is still an under-researched topic, leading researchers to convert such irregular data into a regular format, i.e. 3D voxel grids, and use the better-studied GANs, although this type of conversion adds unnecessary volume to the representation and makes it less memory-efficient \cite{qi2017pointnet}. 
It was found that when working with raw point clouds, the PointNet network is usually used as an encoder to capture the features of this data. A more specific network was expected for meshes, without using convolutions.  However, as can be seen in the work of \cite{ben2018multi}, the use of convolutions could be appropriate for the generation of meshes.
It is expected that more networks will emerge that can handle such data, as they contain a different type of information that is relevant to other areas such as autonomous driving.

\section{Generative Adversarial Networks}
\label{sec:GANs}
GANs were first introduced by \cite{Goodfellow2014}. They proposed a new framework in which two networks are trained to compete and overcome each other: the generator and the discriminator. The generator is trained to learn the real data distribution, i.e. it learns the distribution of the dataset and generates new synthetic data. The discriminator is trained to discriminate between real and synthetic data. The latter can be trained with one of two main objectives: calculate the probability of the data being real or fake, or calculate the realness or falseness of the given data \citep{Arjovsky2017}.

\begin{figure}[]
\centering
\includegraphics[width=0.8\columnwidth]{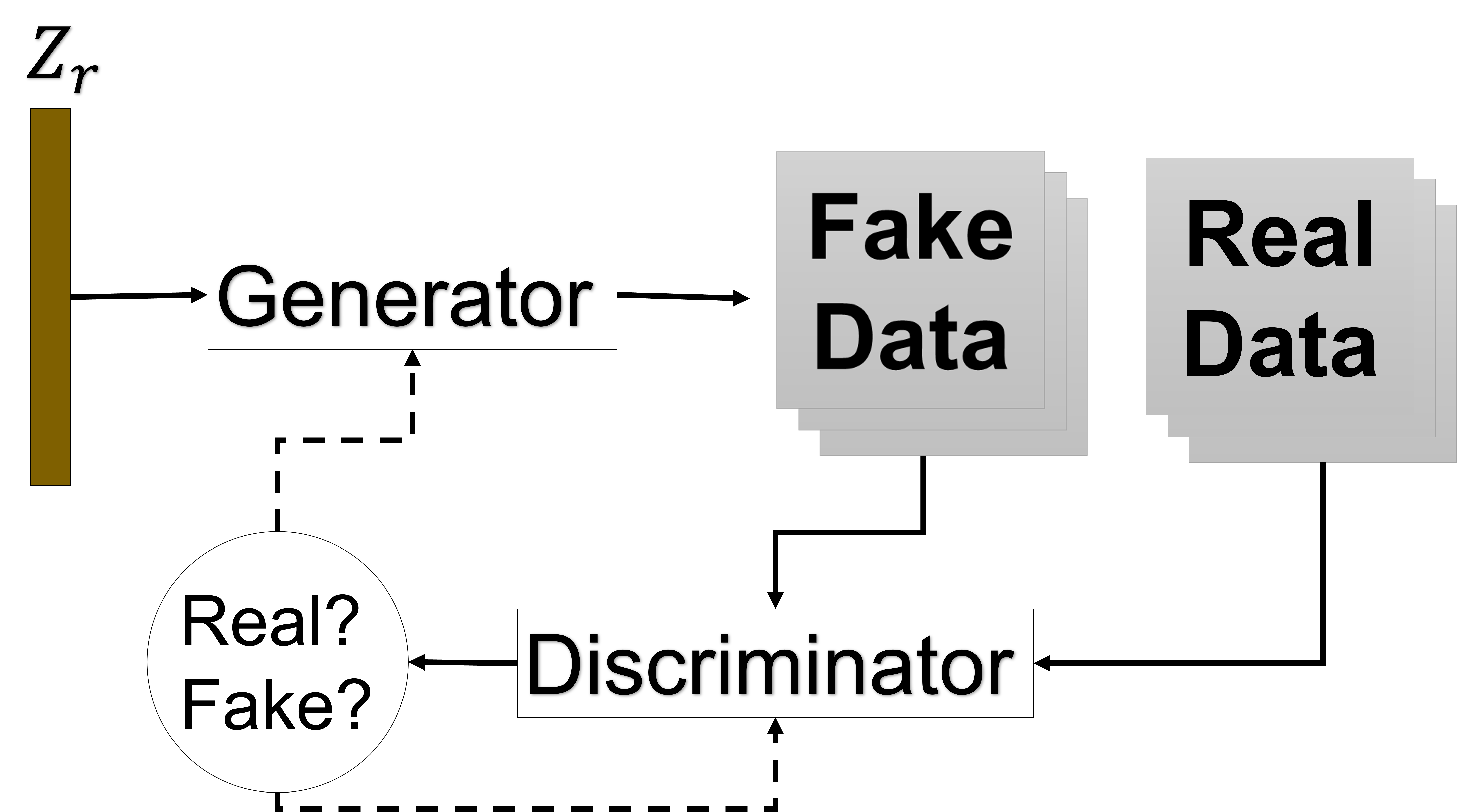}
\caption{Illustration of the vanilla generative adversarial network (GAN). The solid lines are the data transfer and the dashed lines are the feedback/losses.}
\label{GAN}
\end{figure}

Figure \ref{GAN} illustrates how a vanilla GAN works. The Generator receives a random vector ($Z_r$) as input and generates fake data. Then the Discriminator receives fake and real data and returns a probability. This value is the feedback/loss that is passed on to the generator and the discriminator itself. If the discriminator always outputs values close to $0.5$, this means that it is not able to distinguish between true and fake samples, so convergence has been archived.  

In the original work, random noise was used as input to the generator, but it can be extended to a variety of input types by using other GAN architectures, e.g. conditional GAN (cGAN) \citep{Mirza2014}. This allows the discriminator to receive auxiliary information, e.g. labels, in addition to synthetic and real data. 

GAN training, also called "minmax game", aims to satisfy the objective Function \ref{eq:GAN}, where $x$ denotes the real data, $p_z(z)$ denotes a prior input noise, $G(z)$ denotes the generated image, $D(G(z)))$ denotes the probability of the fake image to be true, and $D(x)$ the probability of a real image to be true. For the cGAN, the objective function is very similar with the original one, but $D(x) \rightarrow D(x\mid y)$, and $D(G(z)) \rightarrow D(G(z\mid y)$, where $y$ is the condition.

\begin{equation} \label{eq:GAN}
\begin{split}
min_{G}max_{D}V(D,G)= \mathbb{E}_{x\sim p_{data}(x)}[log(D(x))] \\ +\mathbb{E}_{z\sim p_{z}(z)}[log(1-D(G(z)))]
\end{split}
\end{equation}

However, this equation has the problem of saturation in minimising the loss of the generator $log(1-D(G(z)))$. To solve this problem, \cite{Goodfellow2014} propose to maximize $log(D(G(z)))$ instead. This technique is also known as a non-saturating GAN. For more detail about the vanilla GAN and cGAN, refer to \cite{Goodfellow2014, Mirza2014}.

\subsection{Main purposes of GANs}
\label{sec:main_purposes_of_GANs}
\cite{Goodfellow2014} used GANs to improve the realism of generated images beyond what was achieved with autoencoders, variational autoencoders and other generative models.
In the last few years, GANs have been further developed with works such as \cite{Karras2018, Karras2019}, which are capable of creating realistic human faces. The rapid progress of GANs has raised some concerns, such as the expansion and introduction of new threats and attacks, according to \cite{brundage2018malicious}. The report examines common artificial intelligence architectures and addresses several concerns about the security threats, for example phishing attacks or speech synthesis, which enables easier attacks on a larger scale.

Apart from these concerns, GANs enabled the development of great tools such as image-to-image translation \citep{isola2017image, CycleGAN2017}, text-to-image translation \citep{zhang2017stackgan, dash2017tac}, super-resolution \citep{Ledig2017}, 3D object generation \citep{Wang20172317}, semantic translation to image \citep{wang2018high}, removal of noise and image correction \citep{zhang2019image, tran2020gan}, disentanglement using GANs \citep{wu2021stylespace}, among many other applications.

GANs have also improved algorithms in the medical field, mainly through their ability to generate synthetic images to train other deep learning algorithms.  \cite{jeong2022systematic}  provide a systematic review of medical image classification and regression and demonstrates the results that can be achieved by using these architectures. \cite{yi2019generative} presents a more comprehensive overview of GANs in medical image analysis, highlighting the modalities and tasks. 

On the other hand, our systematic review examines the use of GANs to generate volumetric data, including medical and non-medical data but properly separated for clarity. These papers report on the use of GANs for denoising, nuclei counting, reconstruction, segmentation, classification, image translation or simply general applications.

\subsection{Advantages and disadvantages}
The main advantages of GANs are mentioned above in section \ref{sec:main_purposes_of_GANs}. They can be used to generate realistic data which might be used for data augmentation for other purposes, with a higher realism than other approaches. As demonstrated in \cite{ferreira2022generation}, the use of GANs for data augmentation can outperform the use of conventional data augmentation.

One of the main concerns with this generative technology is the ability to produce deep fakes \citep{shen2018deep, Ponomarev2019FoolingIC} and trick detection algorithms or even humans, which can lead to misunderstandings or fraud. As evidenced by \cite{brundage2018malicious}, this raises several cybersecurity concerns, as images or even fake videos can be created that are so realistic that they can fool anyone with a simple application, e.g. \href{https://deepfakesweb.com/}{Deep Fakes} (accessed [06-06-2022]). However, it can also be used to improve cybersecurity and combat deep fakes \citep{navidan2021generative, arora2020review}. Although this technology permits harmful effects, the benefits that can be derived from it are immense.

In addition to the concerns mentioned above, GANs also have challenges that are more technical \citep{chen2021challenges}:

\begin{itemize}
    \item \textbf{Mode collapse:} when the generator is not able to produce a large number of outputs, but always produces a small set of outputs or even the same one. This can happen when the discriminator is stuck in a local minimum and the generator learns that it is possible to produce the same set of outputs to fool the discriminator;
    \item \textbf{Non-convergence}: as the generator produces more and more realistic data and the discriminator cannot follow this evolution, the discriminator's feedback gradually becomes meaningless. The convergence point of the network can be characterised by the point at which the discriminator is no longer able to distinguish between real and fake data and only gives random guesses. When this point is passed, the generator is fed with poor feedback, causing the results to deteriorate;
    \item \textbf{Diminished gradient}: On the other hand, the feedback becomes meaningless to the generator if the discriminator performs too well. If the generator cannot improve as fast as the discriminator, or in the case of an optimal discriminator, this problem can occur;
    \item \textbf{Overfitting}: When the amount of data is very limited and no precautions are taken, the discriminator may overfit on the training data, i.e. the discriminator's output distribution for the real and the fake samples do not overlap, making the feedback meaningless;
    \item \textbf{Imperception}: no loss function or evaluation metric is able to mimic human judgement, which makes comparison between models very challenging without human intervention. The large-scale applications, e.g., denoising, reconstruction, synthetic data generation and segmentation, bring a lot of heterogeneity, which makes it harder to define how we can evaluate them.
\end{itemize}

These challenges are not limited to the generation of volumetric data, but are even more difficult to overcome due to the additional dimensionality. More dimensions mean more data complexity and more computational effort for processing, which makes the experiments more difficult and time-consuming. As a result of these challenges, the generation of volumetric data with GANs is still under-researched compared to 2D imaging. 

\subsection{Further reading}
As supplementary reading, we present the following list of papers:
\begin{itemize}
\item \cite{Goodfellow2014} — Generative adversarial nets;
\item \cite{Gui2020} — A Review on Generative Adversarial Networks: Algorithms, Theory, and Applications;
\item \cite{kazeminia2020gans} — GANs for medical image analysis;
\item \cite{lucic2018gans} Are GANs Created Equal? A Large-Scale Study.
\item \cite{CycleGAN2017} — Unpaired Image-to-Image Translation using Cycle-Consistent Adversarial Networks;   
\end{itemize}

These papers present how GANs work, the different ways they can be used, and the general advantages and disadvantages. They provide the reader with a deeper, but also broader, understanding of GANs, although reading them is not mandatory to understand the purpose of this review.

\subsection{Applications}

\subsubsection{\textbf{Image translation}}

\begin{figure}
    \centering
    \includegraphics[width=1\columnwidth]{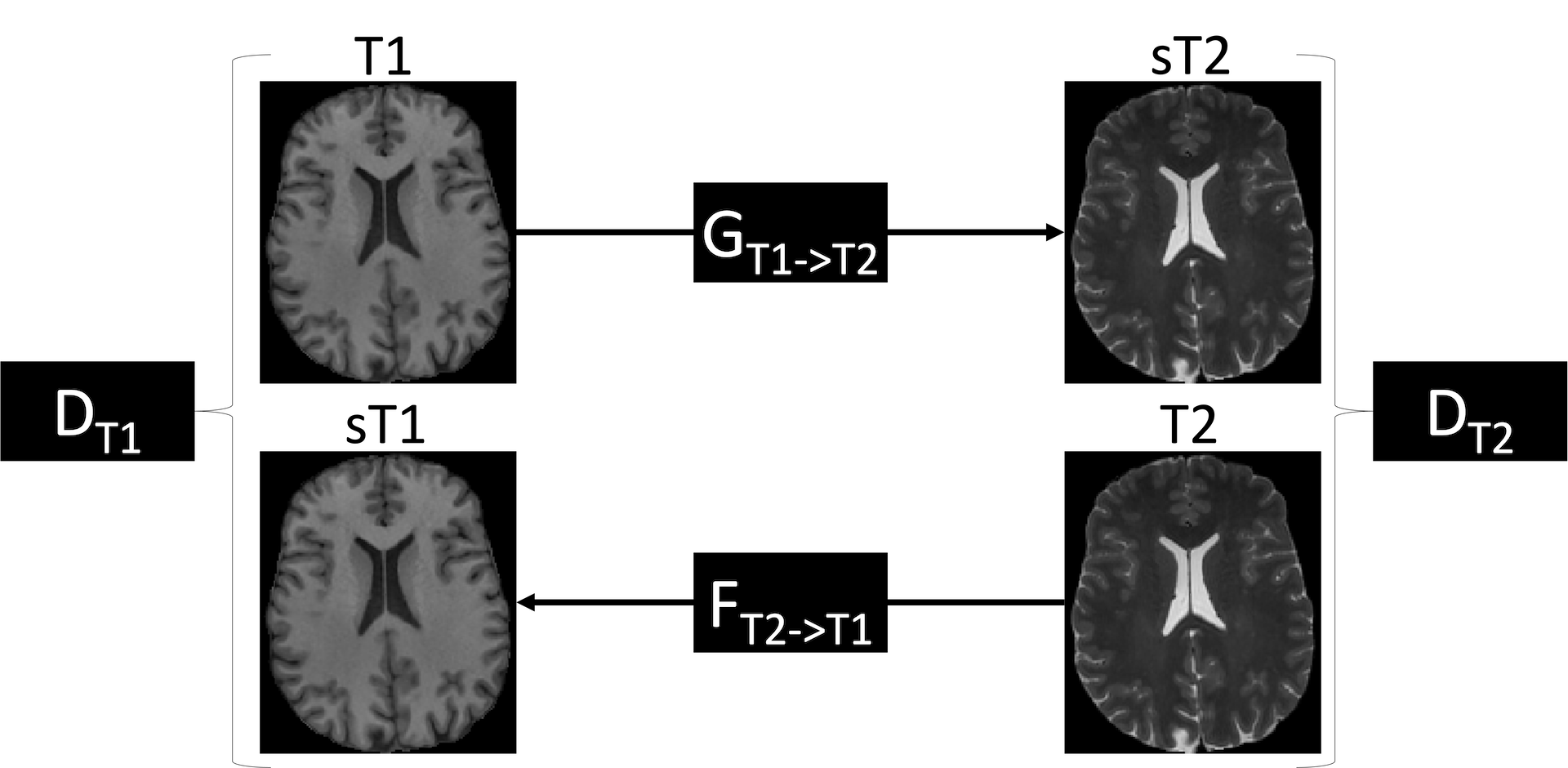}
    \caption{Basic illustration of the CycleGAN architecture.}
\label{Fig:CycleGAN}
\end{figure}

Image translation converts an input image into another synthetic version of that input image,  e.g. a photo taken in the morning into an evening photo. Only works dealing with the translation of medical images were found in this review. In the medical context, this is usually the translation between modalities, i.e., CBCT $\leftrightharpoons$ CT, CT $\leftrightharpoons$ PET, CT $\leftrightharpoons$ MRI, and MRI $\leftrightharpoons$ PET. 
When the dataset contains both images, a Pix2Pix-based cGAN architecture can be used. However, in cases where paired data is not available, a CycleGAN-based architecture is generally used which takes advantage of the cycle consistency loss (section \ref{subsub:Cycle_consistenc}). Taking as example Figure \ref{Fig:CycleGAN}, $sT2=G_{T1\rightarrow T2}(T1)$, $sT1=F_{T2\rightarrow T1}(T2)$. 

This application is particularly appealing in situations where the presence of paired data is important, e.g. in brain tumour segmentation, or in situations where the patient's exposure to a particular modality should be reduced, e.g. in creating synthetic CT scans from MRI scans so that no radiation is used. With the CycleGAN architecture, it is even possible to create paired data without having a previous paired dataset.

\subsubsection{\textbf{Reconstruction and Denoising}}
\begin{figure}
    \centering
    \includegraphics[width=0.7\columnwidth]{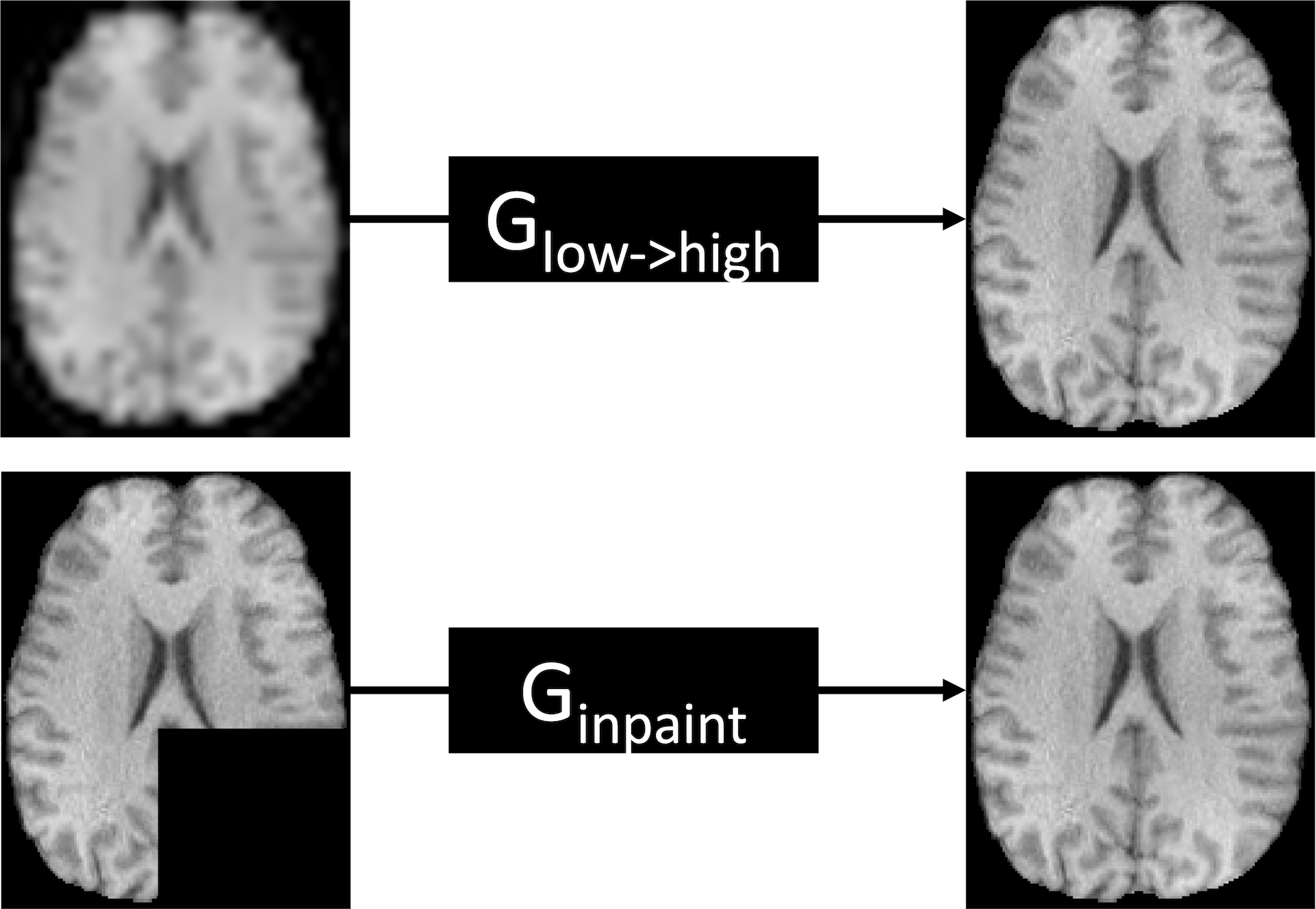}
    \caption{Basic illustrations of reconstruction architectures: First row - From low to high resolution; Second row - Inpainting.}
\label{Fig:Reconstruction}
\end{figure}

In computer vision, reconstruction is the process of capturing the shape and appearance of real objects in order to complete, i.e. reconstruct, incomplete objects. The basic pipelines of two reconstruction tasks (low-to-high and inpainting) are shown in Figure \ref{Fig:Reconstruction}, where the high resolution volume ($V_{high}$) is given by $V_{high} = G_{low\rightarrow high} (V_{low})$, and the complete volume ($V_{complete}$) is given by $V_{complete} = G_{inpaint}(V_{incomplete})$. The former is also known as "super-resolution", where a GAN is trained to increase the resolution of low resolution samples. Using GANs for this could be interesting for some tasks, however, unrealistic information could be added that could harm the downstream task, e.g. artefacts in super-resolution of low-dose CT scans.

In this review, most reconstruction tasks were from 2D to 3D, i.e. reconstruction of volumes from two-dimensional images. Other works seek to generate complete from corrupted 3D scans, e.g., \cite{Wang20172317}, who uses a generative network to complete damaged 3D objects (this task can also be called inpainting).
The remaining ones deal with the generation of high-resolution volumes from low-resolution, e.g., \cite{Halpert20192081}, who improve the resolution of 3D seismic images. Reconstruction can also be applied in medicine, e.g. \cite{Moghari2019} has developed a GAN to generate high-resolution CT from low-dose CT scans. This task can bring several advantages, such as shortening the acquisition time or reducing the radiation needed. Commonly, PSNR (section \ref{subsub:PSNR}) is used to assess the quality of these reconstructions, as a volume with low resolution can be considered a volume with noise, i.e. with a low PSNR. MS-SSIM, MSE and other voxel-wise metrics (section \ref{sec:Evaluation_Metrics}) can also be used to evaluate the reconstruction task.

Denoising is the process of removing noise (i.e. grainy appearance, artefacts and random information) from the volume to restore the true appearance of the volume. This process is very delicate because removing noise can result in removing important features and details of the volume. This task is very similar to reconstruction, but it is more specifically about removing artefacts from the volume rather than increasing resolution or inpainting. Formally, this task can be defined as $V_{denoised} = G_{denoising}(V_{noise})$. Normally, noise from a known distribution, such as a Gaussian distribution, is used to create pairs of noisy and denoised volumes, but in real-world scenarios the noise does not follow any particular distribution. To solve this problem, \cite{Li2020} has developed a GAN to learn the distribution of real-world fast optical coherence Doppler tomography (ODT) noise. Then, this noise is added to the noise-free ODT and a GAN-based denoiser is trained. Such approaches can lead to reducing the acquisition time of ODT scans without compromising the quality of the volumes.

\subsubsection{\textbf{Classification}}
\begin{figure}
    \centering
    \includegraphics[width=0.7\columnwidth]{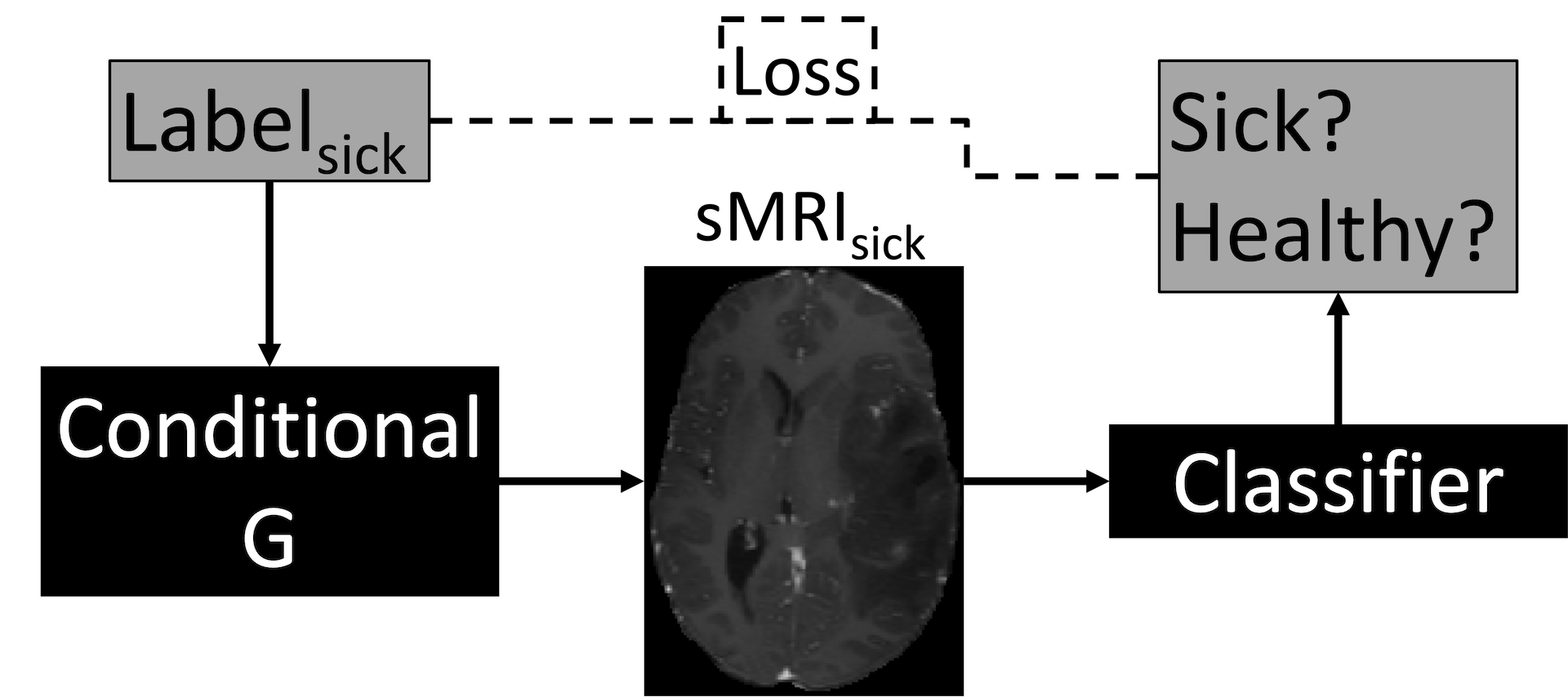}
    \caption{Basic illustrations of a classifier training using synthetic data.}
\label{Fig:Classifier}
\end{figure}

Classification involves training a model that predicts the correct label/class of the input data, i.e. instead of outputting a volume, the network only outputs the label. The vanilla GAN consists of a generator and a classification network, i.e. the discriminator. In some cases, the discriminator is trained not only to classify the input data as real or fake, but also to predict the class of the object, e.g. in \cite{Muzahid202120}. The classification task is not directly related to the generator, but can be used to improve it, e.g. a conditional generator ($cG$) must be able to produce synthetic data with features corresponding to the class used as a condition, as illustrated in Figure \ref{Fig:Classifier}. This is formally defined as 
$Prediction=Classifier(sMRI_{sick})$ where $sMRI_{sick}=cG(LabeL_{sick})$, and the loss can be calculated with e.g. CE, $Loss=L_{CE}(Prediction, LabeL_{sick})$. If this synthetic data is classified correctly, it means that the generator is producing good data. A classifier can be used to classify synthetic data directly from the conditional generator. This approach is very useful for situations with limited annotated data, as the synthetic data contains the appropriate labels that were used as input to the generator, e.g., \cite{Jung202079} trained a conditional generator to create synthetic brains with different stages of Alzheimer's disease. This task is also useful for evaluating the quality of the generated data and comparing models, e.g. when the goal of generating synthetic data is to improve a classification model. Accuracy, AUC, and CE (sections \ref{subsub:acc_sen_spe_pre_AFP_type_I_II}, \ref{subsub:AUC_ROC} and \ref{subsub:CE}) are often used to evaluate the performance of classifiers.

\subsubsection{\textbf{Segmentation and Nuclei counting}}
Segmentation is the classification of voxels to identify whole objects that belong to a class. The volumes are then divided into distinguishable and meaningful regions for object recognition. This task has been extensively researched by the computer vision community as it allows for the automation of various tasks such as counting the number of people in an image, detecting and measuring the volume of a tumour for treatment planning, organ segmentation and much more. The papers found in this review that use GANs for segmentation are all related to the medical field, but can also be used for non-medical tasks. For example, \cite{Zhang2021} uses two GAN architectures for semantic segmentation of 3D pelvis CT scans, using a generator ($G$) to produce unannotated synthetic scans. Then, the synthetic data and the annotated real data are used to train another GAN consisting of a segmentation network ($S$) and a discriminator to distinguish between real masks and masks generated by the segmentation network. With this approach, it is possible to use both labelled and unlabelled data. This is formally defined as $\hat{y}=S(\hat{x})$ with $\hat{x}=G(z)$ where $z$ is a random vector and $\hat{y}$ the segmentation. \cite{Ho2019} use a CycleGAN-based architecture to generate synthetic scans with the corresponding segmentation labels. The goal is similar to Figure \ref{Fig:CycleGAN}, but instead of translating between modalities, the translation is performed between the synthetic microscopy 3D volumes and the segmentation mask.

\cite{Han2019} is the only work that investigates GANs for nuclei counting. This task is very similar to \cite{Ho2019} segmentation explained earlier, but the real downstream task is not segmentation. \cite{Han2019} train a 3D GAN to produce synthetic distance maps, which are then used for nuclei counting by performing threshold and connected component analysis. 

\subsubsection{\textbf{General}}
In this overview, general task means generating synthetic data without applying it to a specific downstream task. This task aims to improve the visual aspect of the generated volumes. \cite{Rusak202011} used a GAN to improve the borders between tissues of synthetic brain MRI scans, \cite{Danu2019662} generated synthetic blood vessels indistinguishable from the real ones, and \cite{Liu20196164} generates synthetic Berea sandstone and Estaillades corbonate to increase the amount of data available for other tasks. Usually, the visual assessment is the chosen approach to evaluate such task, however, it is intriguing why so few works used the visual Turing test (\ref{subsub:visual}).

\section{Generating realistic synthetic 3D data: a review of works}
\label{sec:Generating_realistic_synthetic_3D_data:_a_review_of_works}

The use of GANs to generate synthetic data has increased significantly in recent years. These deep learning networks are becoming very popular, as they can generate more realistic and sharper synthetic images than other traditional generative approaches. GANs are implicit models as they do not use explicit density functions, in contrast to variational autoenconders, which are explicit models \citep{ mohamed2016learning}. Several review articles have been submitted on the use of GANs for different purposes, ranging from more general review articles, e.g. \cite{alqahtani2021applications}, to more specific articles, e.g. \cite{apostolopoulos2022applications}. 

\begin{figure}
  \centering
  \includegraphics[width=1\columnwidth]{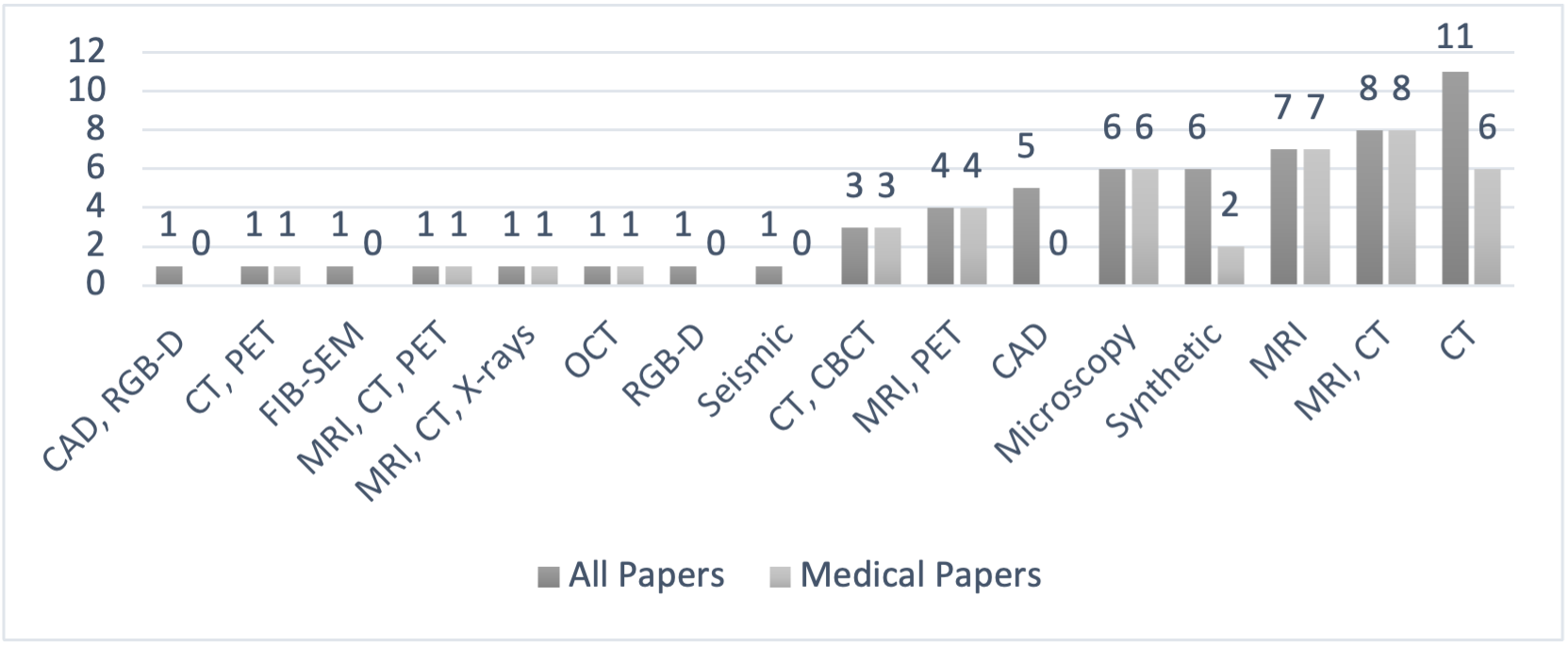}
  \caption{Number of volumetric data papers per modality.}
  \label{Fig:Modalities_All_Medical}
\end{figure}

This section provides statistical information on the papers considered relevant. The number of papers on volumetric imaging has increased in recent years, and from 2014 to 2016 no papers were published (Figure \ref{Fig:year_all_3d}). MRI and CT are the two most popular modalities (Figure \ref{Fig:Modalities_All_Medical}), with more than half of the papers considered applied in the medical field (Figure \ref{Fig:Medical_3D_non_human}).

The modalities used in the reviewed papers are: MRI \citep{mcrobbie2017mri},  CT \citep{scarfe2006clinical}, PET \citep{townsend2008positron}, optical coherence tomography (OCT) \citep{bezerra2009intracoronary}, microscopy\footnote{\href{https://www.umassmed.edu/cemf/whatisem/}{What is Electron Microscopy? (accessed [06-06-2022])}} \citep{fadero2018lite}, CAD \citep{shivegowda2022review}, red, green, blue depth sensors (RGB-D) \citep{zhou2021rgb}, seismic reflection data \citep{dumay1988multivariate}, focused ion beam scanning electron microscopy (FIB-SEM) \citep{fischer2020inkjet}, and kelvin probe force microscopy (KPFM) \citep{melitz2011kelvin}.

\begin{figure}
  \centering
  \includegraphics[width=0.8\linewidth]{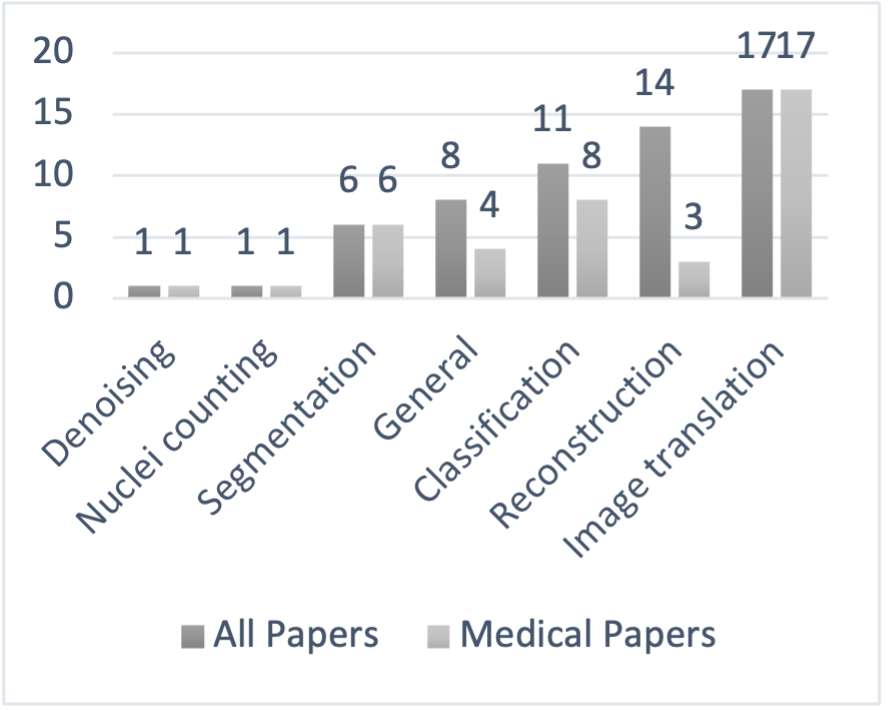}
  \caption{Number of volumetric data papers per non-medical application.}
  \label{Fig:Application_nonNmedical}
\end{figure}

\begin{figure}
  \centering
  \includegraphics[width=1\columnwidth]{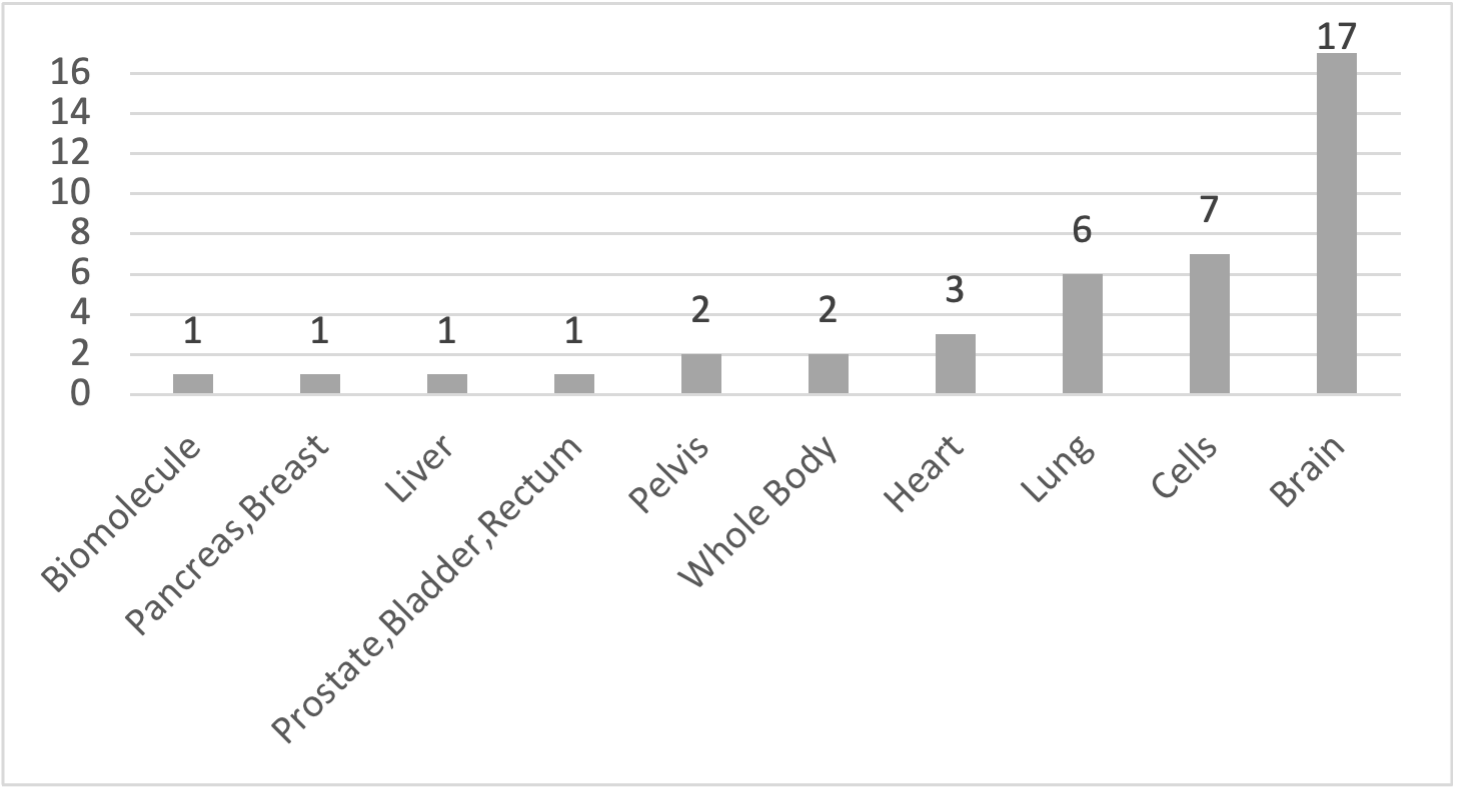}
  \caption{Number of medical papers per organ.
  \label{Fig:Organ}}
\end{figure}

Reconstruction is the main application of GANs in the non-medical context and image translation in the medical context, as can be seen in Figure \ref{Fig:Application_nonNmedical}.
It is worth noting that most papers in the field of medicine are related to humans (Figure \ref{Fig:Medical_3D_non_human}). The most commonly studied organ in the medical field is the brain, as shown in Figure \ref{Fig:Organ}. This popularity results from the increasing use of MRI to study the brain and the need for pipelines to process these images and automate processes. MRI scans of the brain are also easier to process/analyze compared to the rest of the body because there is less movement and variation, resulting in fewer artefacts and faster acquisitions.

In Figure \ref{Fig:Modalities_per_year} it is possible to find the number of publications per year in relation to the different modalities. CT and MRI are present almost every year, which corresponds to the reality of volumetric data acquisition. Especially in the medical field, MRI and CT are widely used. In the non-medical field, CT and X-rays are mainly used, but CAD is also intensively studied due to development of sensors such as RGB-D \citep{zollhofer2018state} and LiDAR \footnote{\href{https://oceanservice.noaa.gov/facts/lidar.html}{What is LiDAR? (accessed [06-06-2022])}} \citep{collis1970lidar}.

\begin{figure}
    \centering
    \includegraphics[width=1\columnwidth]{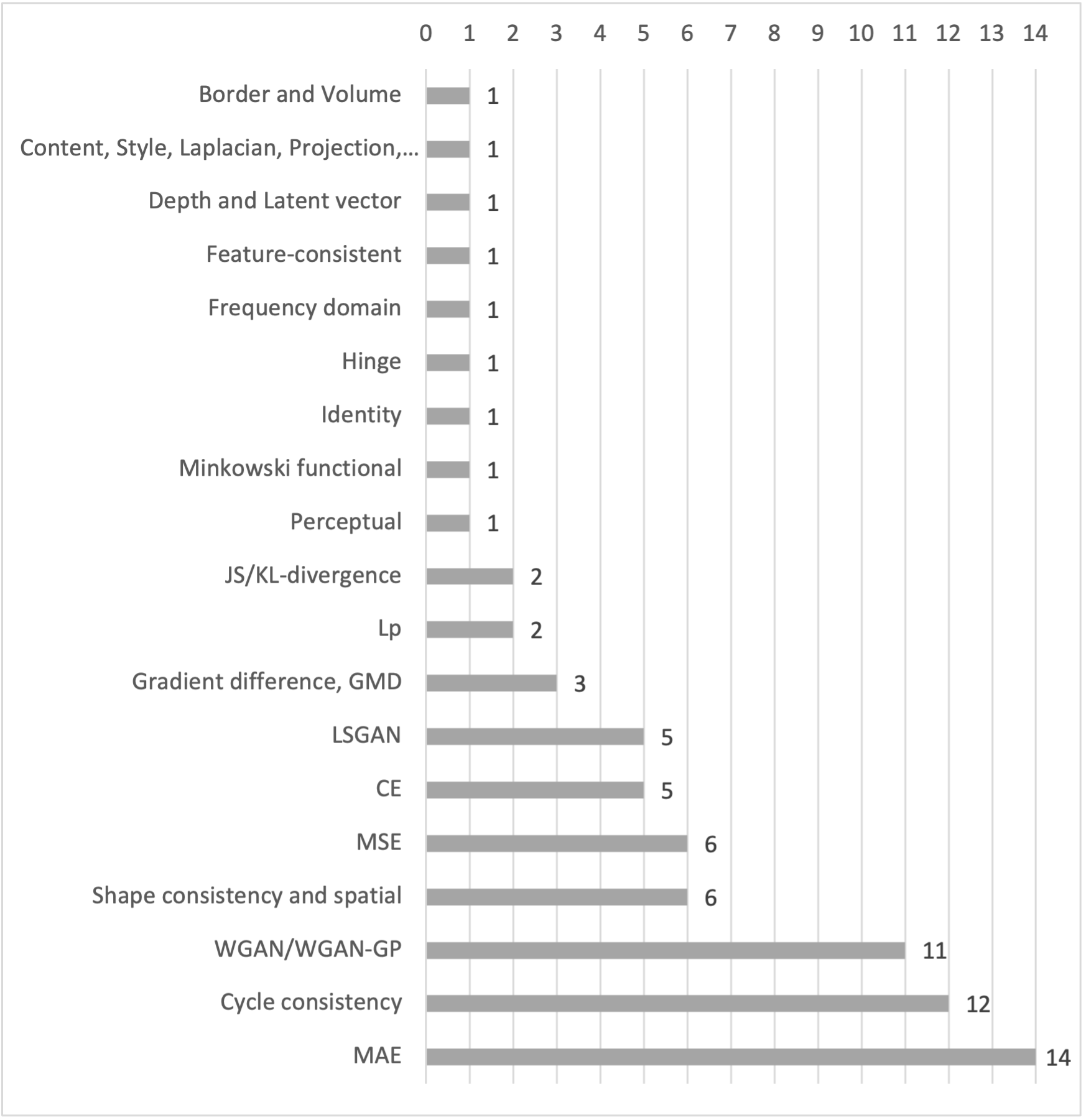}
    \caption{Number of papers per loss function/functions.}
\label{Fig:loss_function_3D}
\end{figure}

\begin{figure}
    \centering
    \includegraphics[width=1\columnwidth]{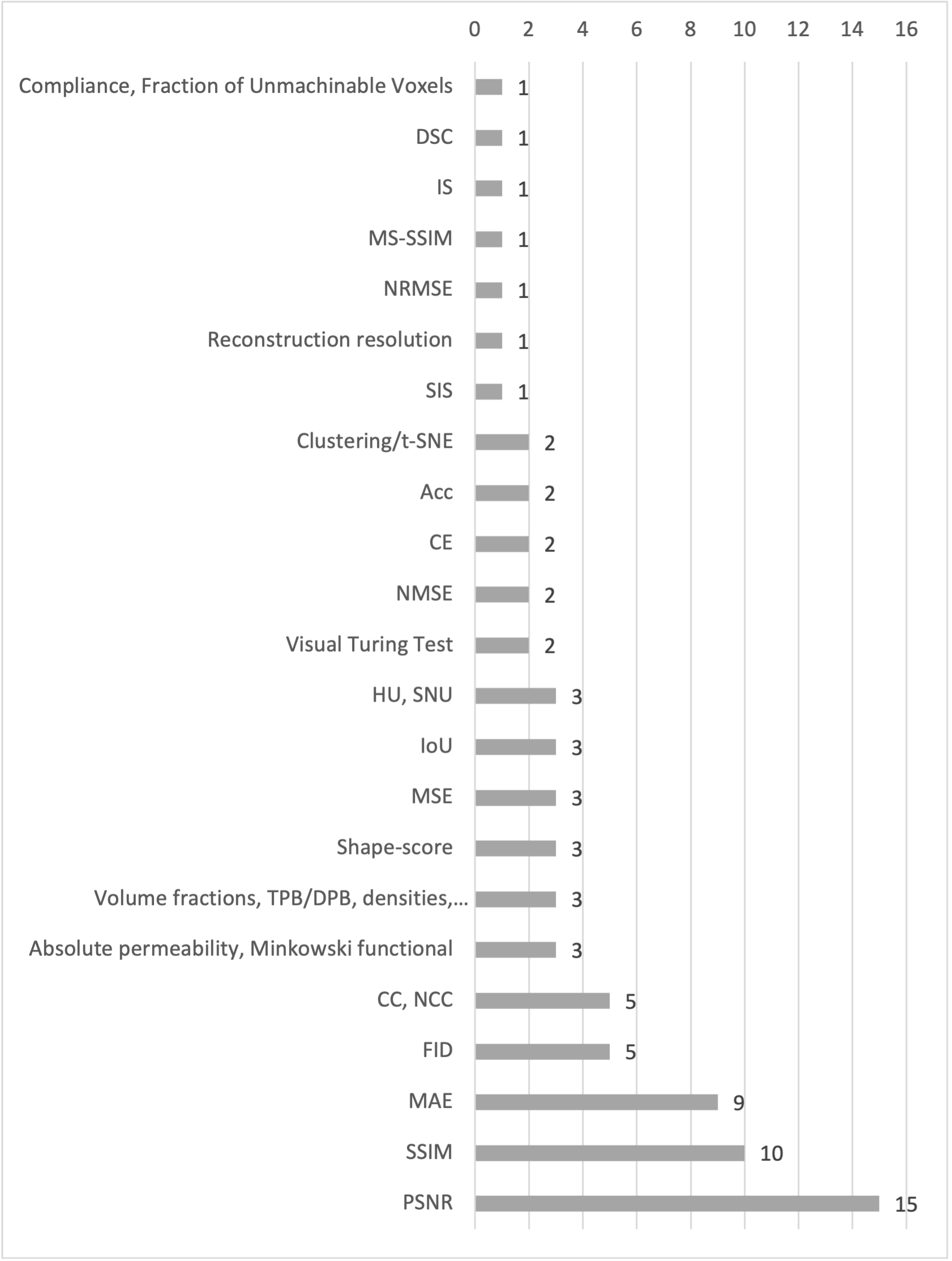}
    \caption{Number of papers per evaluation metric/metrics. The incomplete loss function description in the figure is "Volume fractions, triple phase boundary (TPB), double phase boundary (DPB), densities, relative surface area, relative diffusivity, surface area, two-point correlation function (TPCF)".}
\label{Fig:Evaluation_metrics_3D}
\end{figure}

Figures \ref{Fig:loss_function_3D} and \ref{Fig:Evaluation_metrics_3D} contain the number of papers that use a particular loss function or evaluation metric directly in the generation task (and not in a secondary task). Most of them are easy to explain and widely known, but some require some context to be understood, so it is necessary to read sections \ref{sec:Loss_Functions} and \ref{sec:Evaluation_Metrics}, as well as the Appendix \ref{app:loss_functions} and \ref{app:sec_Evaluation_Metrics} for better understanding.

Figure \ref{Fig:loss_function_3D} shows that mean absolute error (MAE) is the most commonly used loss function because it helps to stabilise the training of the generator and avoid mode collapse \citep{thanh2020catastrophic}, especially in the first steps of the training (cross entropy (CE) and mean square error (MSE) do the same). Cycle consistency is also one of the most commonly used loss functions due to its ability to perform image translations, and then WGAN/WGAN-GP (Wasserstein GAN / Wasserstein GAN with gradient penalty) for its ability to stabilise and prevent vanishing gradient \citep{Gulrajani2017}.

As can be seen in Figure \ref{Fig:Evaluation_metrics_3D}, the peak signal-to-noise ratio (PSNR) is the most common evaluation metric as it is able to evaluate reconstructions, which can be considered one of the main applications of volumetric GANs, although GANs are explored for numerous other purposes. Structural similarity index measure (SSIM) is the second most used metric because it can better assess the quality of the data, taking into account human perception. MAE is often used because it is very easy to calculate and helps stabilise GAN training by making the generator produce data that is realistic rather than just trying to trick the discriminator \citep{isola2017image, pathak2016context}. Usually, Fréchet inception distance (FID) is considered the most reliable metric. However, it is not easy to adapt to volumetric data and even the existing adaptations are not well accepted by researchers.  Accuracy is not really used to assess the quality of the images in terms of perception, but to assess the quality of the data to perform data augmentation, and to assess the improvements of other tasks when synthetic data are used, e.g. segmentation and classification tasks. For that reason, dice similarity coefficient (DSC), sensitivity and F1-score are also commonly used.

\begin{figure}
    \centering
    \includegraphics[width=1\columnwidth]{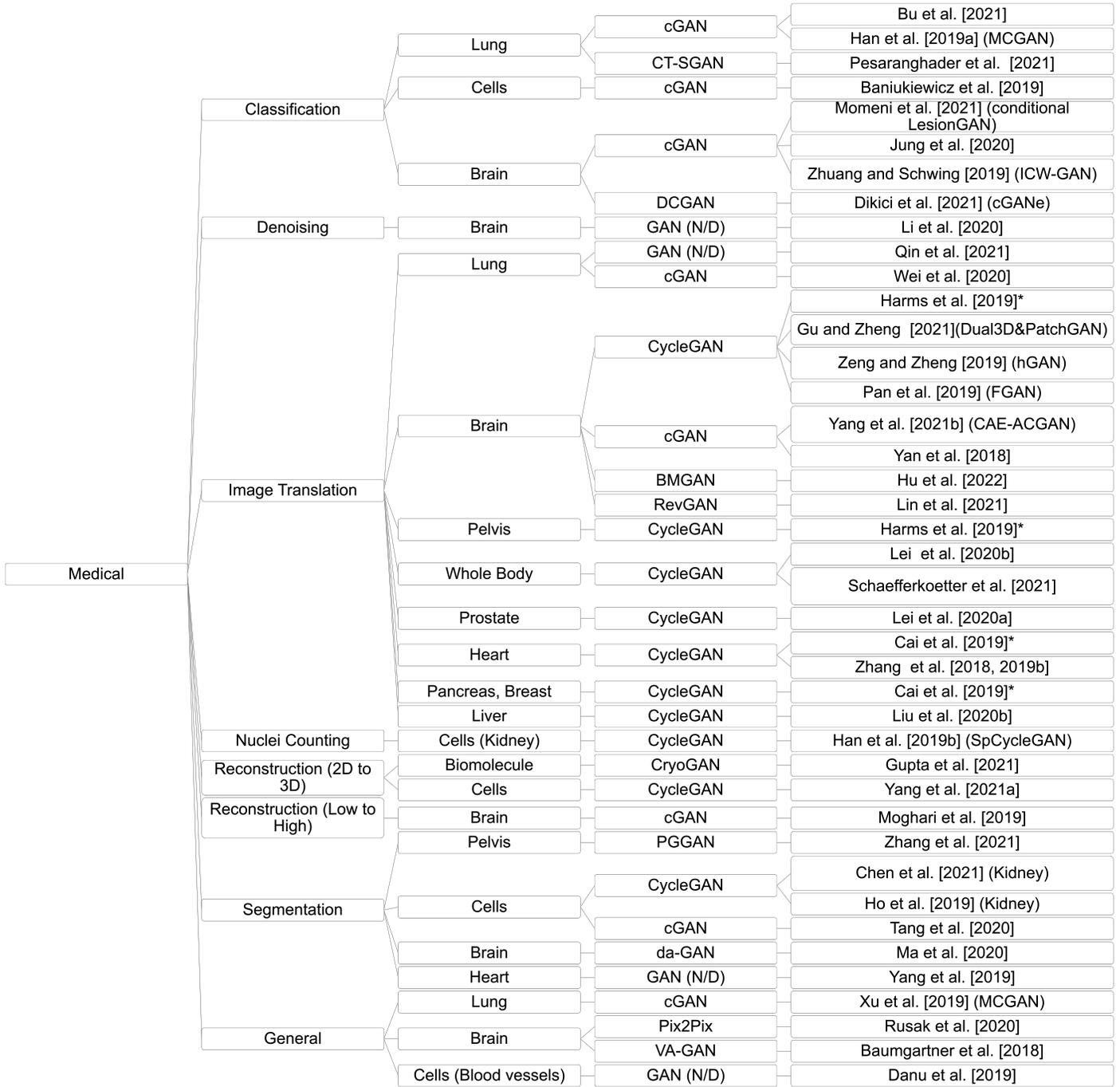} 
    \caption{Proposed taxonomy of medical volumetric GANs.  *These works performed tests in more than one structure.}
\label{Tax:Medical}
\end{figure}

\begin{figure}
    \centering
    \includegraphics[width=1\columnwidth]{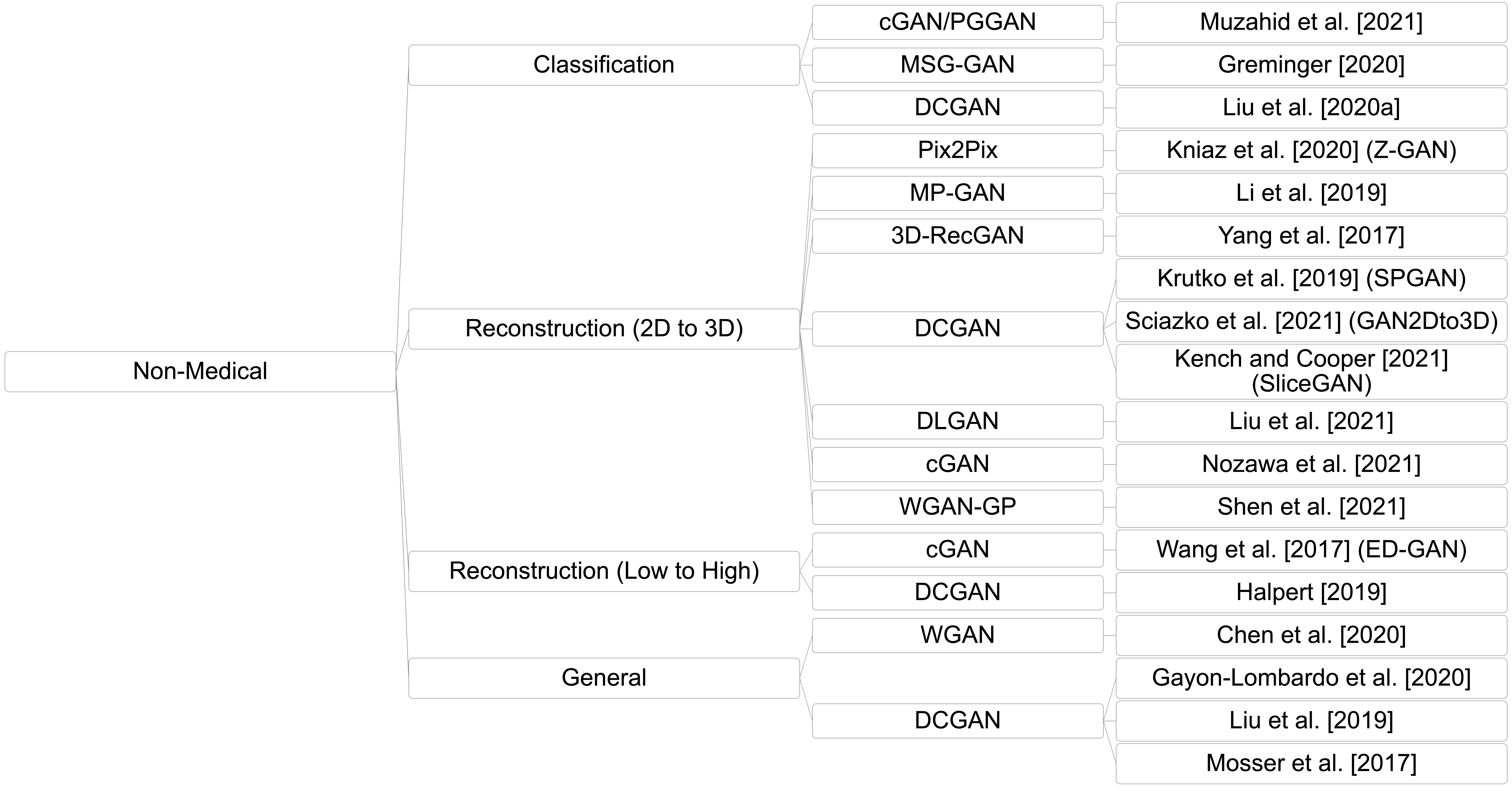}
    \caption{Proposed taxonomy of non-medical volumetric GANs.}
\label{Tax:Non_medical}
\end{figure}

The proposed taxonomy for medical and non-medical GANs is shown in Figures \ref{Tax:Medical} and \ref{Tax:Non_medical} which can guide researchers to relevant publications they are interested in. The taxonomy has been divided into two different figures due to its size. First, a distinction is made between medical and non-medical applications, then by application, architecture, and the respective papers. For example: A user wants to perform a modality translation of volumetric brain images, following "Medical" $\rightarrow$ "Image Translation" $\rightarrow$ "Brain" will find the architectures used and the corresponding papers;
Another user wants to reconstruct non-medical 3D volumes from 2D images, following "Non-Medical" $\rightarrow$ "Reconstruction (2D to 3D)" will find all available architectures and papers for 2D to 3D reconstruction.

Equivalent to \cite{fragemann2022review}, we group the works by task and body part. We have explicitly decided against a technical grouping at the first level of the hierarchical level, as the focus of the review is on the data aspect. So, from the readers' point of view, with a certain type of data in mind, they can find appropriate techniques to utilize. Furthermore, our taxonomy shows that bins/leafs are still under-researched and need more attention from the research community.

\begin{figure}
    \centering
    \includegraphics[width=0.9\columnwidth]{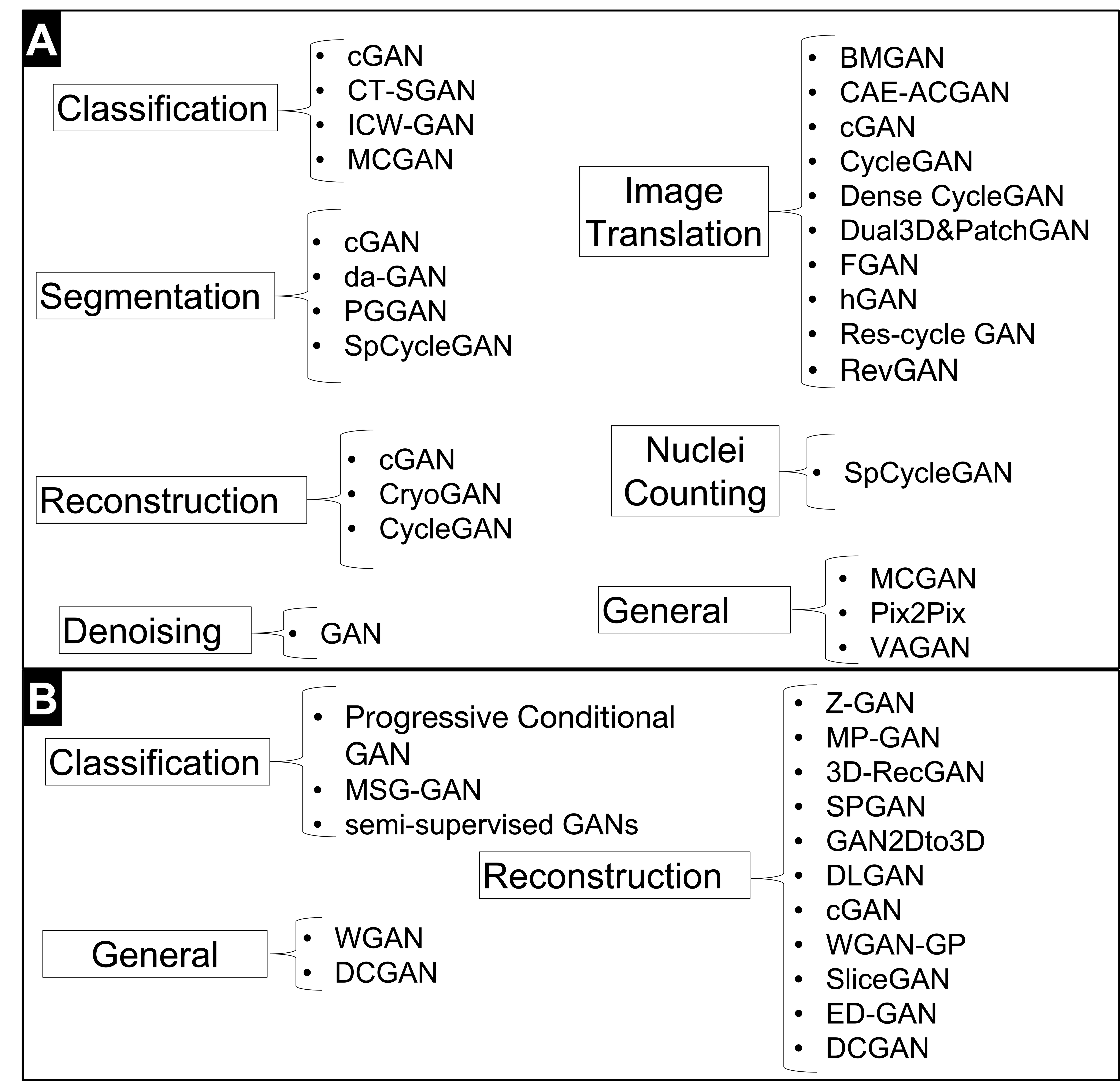}
    \caption{Architectures used in every A) medical application B) non-medical application.}
\label{Fig:ArchToApp}
\end{figure}

Figure \ref{Fig:ArchToApp} shows each different architecture used in each application for medical and non-medical purposes, showing that cGAN-based and CycleGAN-based architectures are used for several different applications.

\onecolumn

\twocolumn

\subsection{Loss Functions}
\label{sec:Loss_Functions}

This section provides a brief explanation of each loss function used in the reviewed papers, along with any references needed and each paper in which each metric was used. It is important to note that some loss functions were developed for a specific problem and only make sense in that context, or they are an adaptation of existing metrics, such as shape consistency and identity. 

In Tables \ref{tab:table2}, \ref{tab:table3} and \ref{tab:table4}, the loss functions between the parentheses were used in the downstream/secondary task and not directly in the generation process.

By definition, the adversarial loss is mandatory when using GANs, as this is the breakthrough for these architectures, so this loss function is always used, although with some variations, e.g., JS, WGAN, WGAN-GP, LSGAN, Hinge. When the authors do not mention which strategy was used, the vanilla GAN loss function is usually applied, i.e. JS, so that "Adv" appears in the tables when the vanilla GAN is used, or it is not mentioned otherwise.

If a cycle GAN-based architecture is chosen, cycle consistency is a good loss function (as are feature consistency, identity, shape consistency and spatial consistency, if possible), as it allows image translation with more stability and realism. 

The use of a pixel-wise loss function (e.g. CE, MSE, MAE, L\textsubscript{p}) in the generator is also important to stabilise it, especially in the first steps of the training. The WGAN-GP loss function is also a good choice to stabilise training and avoid mode collapse, as explained below.

The loss function in the frequency domain seems to be very promising to avoid blurring, however, it has only been used in one paper \citep{Ma2020}, but the ablation studies show that the use of this loss function has improved their results. 

The loss functions are grouped in 4 groups: "Adversarial loss", "Loss functions to explore the intermediate layers", "Pixel/Voxel-wise loss functions", and "Other loss functions".
In sections \ref{subsub:sum_adv_loss_functions}, \ref{subsub:summary_intermediate_layers},
\ref{subsub:Summary_of_the_pixel/voxel-wise_loss_functions}, and \ref{subsub:Summary_of_other_loss_functions} a summary and insight about each group is given. All the metrics are explained in detail in Appendices \ref{app:sub_Adversarial_loss}, \ref{app:sub_Loss_functions_to_explore_the_intermediate_layers}, \ref{app:sub_Pixel/Voxel-wise_loss_functions}, and \ref{app:sub_Other_loss_functions} respectively.
For readers interested in various loss functions, we refer to an excellent review of them \citep{taha2015metrics}.

\subsubsection{Summary of the adversarial loss functions}
\label{subsub:sum_adv_loss_functions}
\cite{lucic2018gans} conducted a study between different GAN objective functions, including the vanilla GAN, LSGAN, WGAN and WGAN-GP (functions used to generate volumetric data, as seen in this review). After experimenting with different datasets, the authors concluded that no method consistently outperformed the non-saturating version of the vanilla GAN (section \ref{sec:GANs}). The authors point out the limitations of the conducted study and mention that more complex datasets with higher resolution might require more layers in the neural network architectures. This is especially true for volumetric data, as a volume with a resolution of only $128$×$128$×$128$ has more voxels than an image of $1024$×$1024$ ($2097152$voxels $>$ $1048576$ pixels), and the presence of contextual and spatial information in volumes which adds more complexity to volumetric datasets even if they contain fewer samples. A pattern can be discerned: WGAN and WGAN-GP seem to perform better on more complex datasets, inferring that such cost functions provide better and more stable training. However, the authors recommend investing more time in optimising the hyperparameters than in the cost functions. None of the functions stand out from the others in terms of stability and generation quality, as this depends heavily on the dataset, its complexity, the complexity of the architecture and the optimisation method. \cite{sulakhe2022crangan} tested the use of Vanilla GAN, LSGAN and WGAN-GP and achieved better results with the first function, which also shows that no function is generally better than the other, it depends on the task at hand. Therefore, tuning the hyperparameters in each of the cost functions used can lead to better performance and thus a better return on investment. As can be seen in this overview, most works stick to the use of the vanilla GAN or the WGAN/WGAN-GP, which should be the cost functions to experiment with first.

Some recommendations can be extracted from this and other papers published until now:
\begin{itemize}
    \item Starting with the non-saturating version of the vanilla GAN cost function. If it still collapses after the following recommendations, experiment with the WGAN-GP.
    \item Using the Adam optimiser (and it is recommended to use this optimiser except for WGAN which should use RMSProp), $\beta _1$ = 0.5.
    \item Using a learning rate of 0.0002 or less can result in more stable training. Tuning the learning rate is very important if the training collapses.
    \item If the discriminator or the generator is trained more often than the other, mode collapse and training instability can be reduced. This can be deduced from the loss plots of the generator and discriminator. If one of the two clearly dominates, such an approach should be tested.
    \item Choosing a lower learning rate for the generator than for the discriminator can also contribute to stability by forcing the generator to make minor changes to fool the discriminator, rather than sudden changes. 
    \item For the WGAN-GP, the weighting of the GP is also important. 10 is recommended, but other values can also provide better results, e.g., \cite{Gupta2021759} set 1. 
    \item Do not use batch normalisation in the discriminator for WGAN-GP training. Instance normalisation should be used for every approach.
     \item The use of spectral normalisation is also appropriate to stabilise the training of the discriminator \citep{ferreira2022generation}.
\end{itemize}

In environments with limited computing power, starting with a subset of the original dataset can speed up hyperparameter tuning and detect bugs:
\begin{itemize}
    \item First make sure that the generator is able to learn by freezing the discriminator and vice versa. 
    \item Second, test with only one sample to confirm that both networks are capable of overfitting. If everything works, this means that the networks should also be suitable for larger datasets. 
    \item Third, use a larger subset and review the loss plots to check for training instability and mode collapse (looking at some generated samples can also help identify mode collapse). If one of the networks outperforms the other, the above tactics should be used.
    \item Finally, test with the entire dataset.
\end{itemize}

For image translation, e.g. between MRI and CT, CycleGAN with cycle consistency loss should be used, whenever possible with shape consistency, spatial consistency, identity consistency or feature consistency (which are explained below).

In our personal experiments with the BraTS2021 dataset to generate synthetic tumours, using WGAN-GP led to more instability and worse results than using the non-saturating vanilla GAN loss. We also found that training the generator twice as much as the discriminator per iteration led to more stable training and thus better results.

\subsubsection{Summary of the loss functions to explore the intermediate layers}
\label{subsub:summary_intermediate_layers}

All of this loss functions (except content and style) assume that a pre-trained model is capable of extract the features that correctly describe the ground truth. In a certain level, these approaches can be compared with knowledge distillation, but here the "teacher" (the pre-trained model) and the "student" (the model to train) have distinct goals and the "student" might not be smaller than the "teacher". However, as discussed in Appendix \ref{subsub:Perceptual}, this assumption may be wrong and the pre-trained model may not be trained correctly (e.g. biased by the data used) or the dataset used for training and the new dataset may be very different, leading to an incorrect extraction of high and low level features.

In any case, if these functions are intended to be used, the latent vector or the feature-consistent  should be chosen, avoiding the perceptual loss. The feature-consistent loss is particularly interesting, as it does not require ground truth to work well with the CycleGAN architecture. If ground truth is available, it is recommended to use the identity loss (Appendix \ref{subsub:Identity}) as it is more intuitive to use, less computationally intensive and not potentially biased.

The content and style loss functions are a mixture of the feature and vanilla adversarial loss, where importance is also given to the encoded vector of each layer of the discriminator to compare high and low level features, rather than just comparing one value of reality. This idea is interesting, especially if one of the features is to be emphasised more. In other cases, however, it could be impractical.

\subsubsection{Summary of the pixel/voxel-wise loss functions}
\label{subsub:Summary_of_the_pixel/voxel-wise_loss_functions}
Adding a pixel-wise metric, such as CE, BCE, MAE, MSE or $L_p$, makes the training more stable, especially at the beginning, and helps the generator learn more quickly what the expected values of each voxel are, rather than just trying to trick the discriminator. Using the adversarial loss alone could lead to unrealistic results, as the generator would only have the task of fooling the discriminator. However, using a voxel-wise metric alone would produce blurred averaged images, which is why using the adversarial loss is important for sharper results.

BCE is suitable for classification problems where only two options are possible, e.g. real and fake, CE can be used when more than two classes need to be compared, e.g. when classifying normal control (NC), mild cognitive impairment (MCI), Alzheimer’s disease (AD).

Projection, orientation and depth losses ensure good 3D generation from 2D data, i.e. these metrics ensure that the visible part used as input is perfectly reconstructed, but they are not able to control the reconstruction of the non-visible part. For the non-visible part, several outputs are possible, but using Laplacian loss or MAE/MSE to compare the volumetric ground truth with the reconstruction could lead to more stable training and better results.

It was expected that more work would be found in the frequency domain, as several features of an image, e.g. texture, edges, among others, can be captured by frequency. However, it is argued that such translation to the frequency domain is unnecessary as 2D and 3D convolutions are capable of capturing the same features.

For shape consistency and spatial losses, it is assumed that the trained segmentation networks are able to correctly segment both modalities. However, as mentioned in the section \ref{subsub:summary_intermediate_layers}, this assumption may lead to incorrect feedback and thus to poorer training of the GAN. 

GMD is the loss function that focuses more on the edges. Therefore, if expressive and non-blurred edges are important for the realism of the generated volumes, this loss function is recommended.

\subsubsection{Summary of other loss functions}
\label{subsub:Summary_of_other_loss_functions}
Many other loss functions can be used as long as they are able to capture the essential features for the realism of the data. Boundary and volume loss functions are used by \cite{Momeni2021} to ensure that the generated cerebral microbleed has the same volume as the input and that the generated portions can merge with the surrounding tissue.
The Minkowski functional is used to measure the distance between geometric features of volumes, i.e. the volume, surface area, average width and Euler's number. This has been used to create synthetic porous media, but can also be used for other structures, such as biomedical images \citep{depeursinge2014three}.

\subsection{Evaluation Metrics}
\label{sec:Evaluation_Metrics}

This section presents a brief explanation of each evaluation metric used in the papers studied, along with the required references and each paper in which the particular metric was used. It is important to note that some metrics are only used for specific purposes and are not well-known, or are sometimes just an adaptation of existing metrics for the problem in question but with a different name, such as fraction of unmachinable voxels and shape-score. 

The evaluation metrics are divided into "Generation Task" and "Downstream Task", as some of them are not used to directly evaluate the quality of the generated data, i.e. the metrics in the "Generation Task" are used to directly evaluate the quality of the synthetic data, while the metrics in the "Downstream Task" are used to evaluate a specific task that uses synthetic data. In Tables \ref{tab:table2}, \ref{tab:table3} and \ref{tab:table4}, the metrics in parentheses are those used in the "Downstream Task". The metrics for assessing the downstream task are not discussed in detail, as these metrics are not used to directly assess the quality of the synthetic data and are highly dependent on each specific task. For more information on each individual evaluation metric for the downstream task, see Appendix \ref{app:sec_Evaluation_Metrics_Downstream_Task}.

It was expected that the visual Turing test would be used more often, as human specialists are the most reliable method to determine the realism of the generated data. However, this is comprehensible as it is very time-consuming and requires specialists. FID comes closest to human judgement, but is mainly used for the evaluation of 2D data, as the adaptations to 3D lack in volumetric context. When ground truth is available, e.g. in a denoising task, metrics such as PSNR, SSIM and MS-SSIM are the most appropriate to use.

\subsubsection{Summary of the evaluation metrics for the generation task}
\label{sec:Evaluation_Metrics_Generation_Task}

The generation task evaluation metrics are divided by voxel-wise metrics (Appendices \ref{subsub:CC_NCC_PCC}, \ref{subsub:CE_eval}, \ref{subsub:MAE_MAPE}, \ref{subsub:MSE_NMSE_NRMSE}, \ref{subsub:IoU}, \ref{subsub:PSNR}, and \ref{subsub:SSIM_MS-SSIM}), visualisation (Appendices \ref{subsub:visual} and \ref{subsub:t-SNE}), middle layers (Appendices \ref{subsub:SIS_S-score} and \ref{subsub:FID_IS}), and problem specific metrics (Appendices \ref{subsub:abs_perm_mink}, \ref{subsub:comp_fraction_un_voxels}, \ref{subsub:HU_SNU}, \ref{subsub:recons_reso}, and \ref{subsub:TPB/DPB}).

For GANs evaluation, every metric that is capable of measuring the distance between the distribution of real world data and the distribution of the generator model can serve as quantifier of performance for generative models.

PSNR is preferred over MSE because it more accurately assesses the visual quality of the data produced. MSE is insensitive to blur, giving good results even with blurred scans. However, PSNR has been shown to be unreliable with human assessment. Slight changes in brightness and contrast do not alter the visual perception of such images, but drastically change the PSNR. MS-SSIM and SSIM are preferred over PSNR and MSE for quality assessment, as they are able to detect structural changes, e.g., distortions, but are almost insensitive to changes in hue (e.g., colour change). Therefore, if the colour is important for the realism of the generated data, both metrics (PSNR and MS-SSIM/SSIM) should be used for the evaluation. However, this will rarely be the case for volumetric data. 

Metrics such as CE, MSE, NCC to compare voxel values are better used for training the generator or classifier, not so much for comparing models, as this metric is too general and not able to capture small details that are important for the realism of the generated data. For each specific case, there are the best metrics, e.g. the Minkowski functional for synthetic generation of porous media, reconstruction resolution, fraction of unmachinable voxels, and other attributes that are specific to the object under study and should be retained in the synthetically generated volumes. These attributes can be compared using graphs or using metrics, e.g, MSE or MAE. When comparing models trained with different datasets, or when the datasets are too heterogeneous, MAPE between absolute values is also highly recommended, although MAE is more commonly used.

In summary, the best approach for comparing models and accessing the realism of the generated data strongly depends on the task, as different tasks have different objects with different characteristics that should be analysed in the evaluation or even in the training phase. However, the use of MS-SSIM, PSNR and clustering t-SNE is strongly recommended for an objective comparison and the visual Turing test for a subjective comparison. The use of approaches that use pre-trained models should be avoided unless the pre-trained model has been trained on the dataset used for GAN training, but even then it is difficult to ensure that the metric is accurate, as explained in Appendices \ref{subsub:SIS_S-score} and \ref{subsub:FID_IS}. Therefore, SIS and S-score are preferable to IS and FID.

Naturally, using synthetic data for the downstream task is the best way to assess whether the generated data is capable of improving the task, but it does not guarantee realism.

\subsection{Architectures}
\label{sub:architectures}
\subsubsection{Based on Vanilla GAN}
\label{sub:Vanilla_GAN_based}

\textbf{HingeGAN / LSGAN / WGAN / WGAN-GP:} These are objective functions rather than architectures. The HingeGAN uses the hinge loss to measure the distance between the discriminator's output and the label, i.e. true or false (Appendix \ref{subsub:Hinge}). The vanilla GAN tries to solve the equation \ref{eq:GAN} using JS divergence (Appendix \ref{subsub:JSD_KLD}), for this it uses the BCE (Appendix \ref{subsub:CE}), but the LSGAN replaces it with least squares loss (Appendix \ref {subsub:LSGAN}).
WGAN uses Wasserstein distance instead of JS divergence, also called earth mover’s distance (Appendix \ref{subsub:WGAN}). In the WGAN architecture, the discriminator is called critic because the output is not a probability of reality, but how real it is, i.e. it is such as the regular discriminator, but without the sigmoid function, so it outputs scalar values.
The WGAN-GP replaces the weight clipping used by the WGAN with gradient penalty, as explained in Appendix \ref{subsub:WGAN}. Several researchers prefer using the WGAN or WGAN-GP loss function over the vanilla GAN function, so this is discriminated in Tables \ref{tab:table2}, \ref{tab:table3} and \ref{tab:table4}.
As explained in section \ref{subsub:sum_adv_loss_functions}, no objective function is superior to the other in every case, but depends mostly on a case-by-case basis. It should be noted, however, that some papers that have used WGAN-GP have also tried the vanilla GAN and obtained better results with the former. However, others mention that no difference in results was found when using one or the other. Therefore, it is recommended to read section \ref{subsub:sum_adv_loss_functions} for further recommendations on the best approaches to generate synthetic volumetric data with GANs.

\textbf{PGGAN (Progressive Growing GANs):} This network attempts to stabilise the training and increase the realism of the generated data by gradually increasing the resolution of the output volumes. In a first step, the generator produces 4×4×4 volumes from random noise and the discriminator is required to distinguish between the synthetic data and the downscaled real data. If the results are satisfactory with this resolution, the networks are increased to generate and discriminate 8×8×8 volumes. This process is repeated until the desired resolution is achieved, which in the case of \cite{Zhang2021} is 64×128×128.

\textbf{CryoGAN:} This architecture is very similar to the vanilla GAN, but the generator is a cryo-EM physics simulator instead of a regular network \citep{Gupta2021759}, and the random vector is replaced by a 3D density map from the cryo-EM physics simulator. This approach can be adapted for different areas where a simulator can replace a regular generator.

\textbf{MSG-GAN (Multi-Scale Gradient GAN):}
This architecture can be understood as a vanilla GAN with skip connections between the generator and the discriminator, with a convolution layer between these skip connections. The generator produces multiple scaled samples that are fed to the discriminator, i.e. the full resolution volume produced by the last block of the generator is the input of the first block of the discriminator, then this output is concatenated with the output of the penultimate block of the generator, and so on. \cite{greminger2020generative} claims that this architecture leads to stable training.

\textbf{semi-supervised GANs:} This architecture is based on the semi-supervised learning technique, which uses both labelled and unlabelled data to train the network \citep{Liu2020O47}. Here the discriminator is also a classifier and the loss function of the discriminator also contains the supervised component, i.e., if the discriminator classifies the input data correctly.

\textbf{DCGAN (deep convolutional generative adversarial networks):} It is based on vanilla GAN, but consists of convolutional layers and transposition convolutional layers instead of fully connected layers. For the generation of images and volumes, GANs consisting of convolutional layers are usually used, but fully connected layers may also be used, as in the vanilla GAN.

\subsubsection{3D from 2D}
\label{subsub:3Dfrom2D}
This section reviews some architectures that are capable of generating volumes from 2D data. MP-GAN, GAN2Dto3D and SliceGAN are able to generate volumetric data from a dataset consisting only of 2D images by using a 3D generator and a 2D discriminator. Z-GAN and SPGAN are able to generate volumes from 2D images, but their training dataset also contains the corresponding 3D data. For this task, a generator consisting of a 2D encoder and a 3D decoder is used.

\textbf{MP-GAN (Multi-Projection GAN):} This architecture aims to solve the problem of generating volumes when only 2D images are available \citep{Li20195530}. The generator is given a random vector and produces volumes that are projected into 2D images with multiple viewpoints. Multiple discriminators are used to evaluate the realism of such projections. The number of discriminators is equal to the number of viewpoints, so each discriminator only needs to learn the distribution of the particular view. Therefore, the discriminator is a 2D network and the generator is a 3D network.

\textbf{GAN2Dto3D:} This architecture is very similar to the MP-GAN architecture, but instead of projecting the output of the generator into multiple views, the generated volume is sliced into three coordinates, i.e. X-Y, X-Z and Y-Z, and only one discriminator is used. It is used by \cite{Sciazko20211363} to generate synthetic 3D microstructures with only 2D real data available for training.

\textbf{SliceGAN:} This architecture is used for the generation of a 3D volume from a simple 2D image. \cite{Kench2021299} presents this architecture for generating 3D microstructures from 2D slices of isotropic material. This architecture is very similar to GAN2Dto3D in that the generator creates a volume from a random vector, decomposes the volume in three axes and feeds the discriminator to compare with slices from large 2D isotropic images.  \cite{Kench2021299} adds that this approach is only possible for isotropic, nondirectional material. Anisotropic materials require more than one view of 2D slices, similar to MP-GAN.

\textbf{Z-GAN:}
It is employed for image-to-voxel translation using a conditional generator consisting of an encoder that encodes the 2D image input and a decoder that generates the volume in a UNet-based format \citep{Kniaz20203}. Skip connections between the 2D and 3D convolutions are made by "copy inflating" the 2D sensors to match the dimensions of the 3D sensors. The discriminator tries to distinguish between synthetic and real volumes by using a 3D PatchGAN. Therefore, Z-GAN is based on the Pix2Pix architecture, with the decoder and discriminator adapted for volumetric data.

\textbf{SPGAN (Slice to Pores GAN):}
This architecture is used in \cite{Krutko2019} to generate 3D volumes of porous media from 2D images of thin sections. For this, the 2D image is encoded and concatenated with a random noise vector, which is decoded by a 3D decoder that generates the volume, i.e. the conditional generator. The discriminator receives both synthetic and real volumes. This architecture is identical to the Z-GAN, but no skip connections are used between the encoder and the decoder, and noise is added to the encoded 2D slice before feeding the decoder.

\subsubsection{Based on conditional GAN}
\label{subsub:conditional_GAN_based}
The cGAN allows the generation of synthetic data with specific user-defined features, e.g. the generation of different classes, defining the position and size of the synthetic tumour, among others. The vanilla cGAN is formally defined by equation \ref{eq:cGAN}

\begin{equation} \label{eq:cGAN}
\begin{split}
min_{G}max_{D}V(D,G)= \mathbb{E}_{x\sim p_{data}(x)}[log(D(x|y))] \\ +\mathbb{E}_{z\sim p_{z}(z)}[log(1-D(G(z|y)))]
\end{split}
\end{equation}
where $y$ is the condition. Typically, both the generator and the discriminator have access to the condition, but some approaches use the condition only in the generator, as can be seen in the subsequent architectures. However, the use of the condition in the discriminator is also important because it helps the discriminator to recognise distinguishing features between different conditions, which leads to better feedback and thus to realistic results.
When the generator input is a volume and not a random vector, the generator is usually based on an autoencoder/U-Net.

\textbf{ICW-GAN (Improved Conditional Wasserstein GAN):}
This architecture is a conditional version of the WGAN-GP. The generator and discriminator are conditioned by the labels, i.e. the labels are concatenated with the input of the generator and the penultimate layer of the discriminator \citep{Zhuang2019}.

\textbf{MCGAN (Multi-Conditional GAN):}
This architecture uses multiple conditions to guide the generator to produce synthetic data with specific features. \cite{Han2019729, Xu201962} generate CT lung tumours using the background volume as input to an autoencoder-based generator.  Conditions can be concatenated with the input background volume \cite{Han2019729} or in the middle of the generator network \cite{Xu201962}. These conditions are also passed to the discriminator to make easier the distinction between real and fake samples, and to force the generator to produce samples with these specific characteristics.  \cite{Xu201962} also add an encoder to predict the characteristics of the synthetically generated volumes to ensure that the generator produces the volumes with the correct features.

\textbf{BMGAN (Bidirectional Mapping GAN):}
BMGAN consists of a conditional generator, a discriminator and an encoder. \cite{Hu2022145} uses this network to generate PET from MRI scans and encode the PET scans back into the latent space. This approach is particularly interesting because using an MRI scan with the random latent vector gives the network more variability and robustness. This architecture is similar to the $\alpha$-GAN proposed by \cite{rosca2017variational}, but they treat the distribution as implicit, whereas BMGAN assumes a standard normal distribution. BMGAN is a cGAN, i.e. it receives as input not only a random vector ($z$) but also the MRI scan, and uses KL divergence (Appendix \ref{subsub:JSD_KLD}) to bring the coded vector distribution close to the random distribution. \cite{kwon2019generation} also uses the stability of the $\alpha$-GAN architecture, but replaces the KL divergence with the Wasserstein distance, achieving better stability in a 3D space. This architecture prevents mode collapse, which is advantageous for training with volumetric data, but also requires more computing power, since four networks are trained instead of only two \cite{ferreira2022generation}.

\textbf{CAE-ACGAN (Conditional Auto-Encoding, Auxiliary Classifier GAN):}
This consists of a variational autoencoder (the generator), a discriminator and a classifier. It is identical to the BMGAN architecture, but instead of an encoder that encodes the synthetic data back into an encoded vector, CAE-ACGAN has a classifier that classifies the synthetic data into the appropriate class. The generator is tuned based on whether the classifier is able to classify the synthetic data correctly or not. \cite{Yang2021415} uses this architecture to perform image translation from CT to multicontrast MRI instead of using a CycleGAN architecture. A comparison is made between their approach, WGAN-GP and Pix2Pix, showing that CAE-ACGAN gives better results, but CycleGAN should also have been tested.

\textbf{Progressive Conditional GAN:}
This architecture consists of a conditional generator that is conditioned by a one-hot coder attached to the random noise vector. The discriminator is a classifier that distinguishes between real and fake data, and outputs a prediction of the class, which is also part of the feedback to the generator. The volumes are refined by a refiner, which consists of four residual blocks after they have been generated by the generator. As with PGGAN, the number of layers of the discriminator and generator increases, increasing the resolution of the volumes. The main difference between PGGAN and this architecture is the conditional input and that the discriminator is also a classifier \citep{Muzahid202120}.

\textbf{VA-GAN (Visual Attribution GAN):}
The VA-GAN presented by \cite{Baumgartner20188309} consists of a mapping function ($M(x)$, UNet-based) that creates a mapping between the differences of two volumes from two different classes (0 and 1), e.g. MRI brain scans from healthy patients and from patients with Alzheimer's disease.  Therefore, instead of using a generator to generate class (1) volumes, this architecture trains an $M(\cdot)$ to generate a map that is concatenated with the input $x$ (of class 1) to produce the synthetic class 0 image, i.e., $y=M(x)+x$. The discriminator must distinguish between the real and the fake class 0 volume.

\textbf{3D-RecGAN:}
This architecture is used by \cite{Yang2017679} to reconstruct 3D volumes from 2.5D partial views of the object.  It consists of a conditional generator (UNet-based) and a conditional discriminator, both conditioned by the 2.5D input view. The 2.5D view is a partial voxel grid, i.e. an incomplete volume that is reconstructed by the generator. The ground truth volumes are needed so that a voxel-wise comparison can be performed.

\textbf{DLGAN (Depth-preserving Latent GAN):}
This architecture, used by \cite{Liu20212843}, is identical to 3D-RecGAN in that it uses a conditional generator (UNet-based) and a conditional discriminator to reconstruct 2.5D voxel grids. DLGAN has an additional autoencoder that learns how to reconstruct the 3D ground truth volume, which helps the generator learn how to represent the volumes in latent space (Appendix \ref{subsub:latent}). Since the final volume consists of binary values, a classification network (consisting of fully connected layers) is used to binarise the output of the generator instead of using a fixed threshold to define which values are 1 and which are 0.

\textbf{ED-GAN (Encoder-Decoder GAN):}
\cite{Wang20172317} use this architecture for the inpainting task. The ED-GAN architecture is similar to the 3D RecGAN architecture in that it consists of an encoder-decoder generator that receives a corrupted 3D volume as input. The main differences between these two architectures are that the ED-GAN discriminator is not conditional, i.e. it only receives the real and reconstructed volumes and not the damaged 3D volume, and that ED-GAN has a recurrent long-term convolutional network (LRCN) to increase the resolution of the generator's output. The LRCN consists of a 3D CNN encoder, a Long Short Term Memory Network (LSTM) and a 2D Full CNN decoder. The 3D CNN encoder receives sparse volumes (more than one slice per input) of the reconstructed volume, the LSTM captures the relationship between the slices, and 2D Full-CNN reconstructs the slices, which are then concatenated back to the volume. By using the LRCN, it is possible to create volumes with higher resolution even if GPU memory is limited.

\textbf{Pix2Pix:}
This architecture is used for image-to-image translations. It consists of a conditional generator (UNet-based) and a PatchGAN discriminator. PatchGAN discriminator uses patches and discriminates
each patch instead of the whole volume at once. The main weakness of Pix2Pix-based architectures is the need for paired data to perform image translation, whereas CycleGAN does not. Pix2Pix calculates the difference between the paired images (using a voxel-wise metric), while CycleGAN calculates the cycle consistency loss (Appendix \ref{subsub:Cycle_consistenc}). Several architectures are Pix2Pix-based, such as Z-GAN.

\subsubsection{CycleGAN based}
The CycleGAN is a specific architecture based on the cGAN, which is often used for image-to-image translation tasks. It consists of four networks, i.e. two generators and two discriminators. It is very similar to Pix2Pix as it learns to map images from one modality to another and vice versa, but it does not require a paired dataset. Figure \ref{Fig:CycleGAN} shows a basic illustration of a CycleGAN applied to modality translation of T1 and T2 from MRI brain scans.
The CycleGAN is formally defined by the vanilla GAN loss function (equation \ref{eq:GAN}, for each pair generator/discriminator) and the cycle consistency loss (equation \ref{eq:cycle_GAN}). It is mainly used for image translation, but has also been applied for segmentation (SpCycleGAN) and reconstruction (\cite{Yang2021130}).

\textbf{SpCycleGAN (spatially constrained CycleGAN):}
This architecture is essentially based on the CycleGAN, but a spatial constraint is added to control the intermediate step \citep{Fu2018}, as explained in Appendix \ref{subsub:shape_spatial}.

\textbf{Dense CycleGAN and Res-cycle GAN (residual block cycle-consistent GAN):}
The architectures of the dense CycleGAN and the Res-cycle GAN are very similar to the vanilla CycleGAN, but in the first dense blocks are added in the middle layers for better contextual and structural feature extraction, i.e. high and low level features, and in the second residual blocks are used for the same purpose. These GANs are described by \citep{Liu2020} and \citep{Harms2020}, respectively. Looking at the two schematic flowcharts, the dense block is identical to the residual block, but in a dense block all the outputs of the previous layers are concatenated in the next layers, and the residual block concatenates the summation of the features, i.e. the dense block has a denser connection between the layers \citep{huang2017densely}.

\textbf{Dual3D\&PatchGAN:}
It is similar to the CycleGAN architecture and can also be used for image translation without the need for paired datasets \citep{Gu2021}. The main difference between Dual3D\&PatchGAN and CycleGAN is that in the first case both generators are updated simultaneously with the same loss value, so that the weights are shared by both networks. PatchGAN discriminator is used for memory efficiency.

\textbf{FGAN (Feature-consistent GAN):}
It is based on the CycleGAN architecture, but a disease-specific neural network (DSNN) is used to extract features from the real and synthetic data, and the difference between these features is calculated. This is called the feature-consistent component (Appendix \ref{subsub:Feature-consistent}). The GAN training is thus composed of a CycleGAN and a DSNN \citep{Pan2019137}.

\textbf{hGAN (hybrid GAN):}
This architecture is composed by a 3D generator and a 2D discriminator. The generator follows an autoencoder structure. This network is weakly supervised by the scans generated by a 2D-CycleGAN, i.e. the volumetric output of the generator is decomposed into 2D images and the MAE between them and the images generated by a trained 2D-CycleGAN is calculated. Using a 3D generator allows for better modelling of spatial information and solves the discontinuity between layers that occurs when layers are generated individually, and using a 2D discriminator is well suited for situations \textbf{where limited data is available} \citep{Zeng2019759}.

\textbf{RevGAN (Reversible GAN):}
RevGAN is based on the CycleGAN architecture, but uses only one reversible generator and two discriminators to perform the image conversion instead of two generators and two discriminators \citep{Lin2021}. This is only possible thanks to the reversible core of the generator, which consists of several reversible blocks \citep{gomez2017reversible}. This architecture is more memory-efficient than the vanilla CycleGAN, since three networks are trained instead of four.

\subsubsection{Double discriminator}
Although this approach has not been extensively researched, it can provide strong results for a variety of tasks. The use of two discriminators can be advantageous when it is crucial to focus on specific parts of the generated volume, but also on the whole volume. For example, when generating MRI scans of the brain with the aim of using them to segment brain tumours, it might be useful to use a specific discriminator to distinguish only that part of the brain. Such an approach is therefore recommended if a specific feature is essential for the realism of the generated data or for the downstream task. 

\textbf{CT-SGAN (CT Synthesis GAN):}
It is composed of 4 subnetworks: Bidirectional Long/Short-Term Memory (BiLSTM) network; a slice generator ($G_{slice}$); a slice discriminator ($D_{slice}$) and a slab discriminator ($D_{slab}$). The $G_{slice}$ is able to generate a volume by creating multiple slices controlled by the BiLSTM network. The $D_{slab}$ is able to learn the anatomical distribution across the slices of the volumetric CT scans and provide volumetric feedback (using 3D convolutions), while the $D_{slice}$ only provides feedback about each slice \citep{Pesaranghader202167}.

\textbf{da-GAN (difficulty-aware attention GAN):}
It consists of two discriminators with difficulty-aware attention mechanism. The first discriminator classifies the entire volume as real or fake, but the second discriminator is specialised in a particular part of the volume that is crucial. This architecture is presented in \cite{Ma2020} for generating 2D data, but the concept can also be applied to volumetric data. \cite{Ma2020} uses a global discriminator for the whole brain and a local discriminator to enhance the hippocampus.

\subsection{3D volumetric data generation}
\label{sec:3D_data_generation}

In this section, it is possible to find all volumetric papers in Tables \ref{tab:table2}, \ref{tab:table3} and \ref{tab:table4}. Table \ref{tab:table2} contain all papers where CT or MRI scans are used, sorted by modality and then by year. If a dataset has no link, this means that it is not available or has not been found. The footnotes of the studies are the links to the available code/framework. 

\textbf{List 1: List of abbreviations of Table \ref{tab:table2}:}
\begin{itemize}
\small
\setlength\itemsep{-0.5em}
\item AFP — Average False Positives;
\item AIBL — Australian Imaging Biomarkers and Lifestyle;
\item C3D — Convolutional 3D;
\item CBF — Cerebral Blood Flows;
\item cGANe — constrained GAN ensembles;
\item CPM — Competition Performance Metric;
\item CTP — CT Perfusion;
\item CT-SGAN — CT Synthesis GAN;
\item da-GAN — difficulty-aware attention GAN;
\item IBC — Individual Brain Charting;
\item ICW-GAN — Improved Conditional Wasserstein GAN;
\item IS — Inception Score;
\item JS — Jensen-Shannon loss;
\item Li-ion battery — Lithium-ion battery;
\item MCGAN — Multi-Conditional GAN;
\item micro-CT — micro Computed-Tomography;
\item LUNA — LUng Nodule Analysis;
\item NLST — National Lung Screening Trial;
\item Noise SD — background noise levels;
\item OCT — Optical Coherence Tomography;
\item ODT — Optical coherence Doppler Tomography;
\item ROC — Receiver Operating Characteristic curve;
\item SIS — Semantic Interpretability Score;
\item SOFC — Solid Oxide Fuel Cell;
\item SPGAN — Slice to Pores GAN;
\item SSA — Specific Surface Area;
\item TPCF — Two-Point Correlation Function;
\item t-SNE — t-distributed Stochastic Neighbor Embedding;
\item VA-GAN — Visual Attribution GAN;
\item XCT — X-ray Computed Tomography;
\end{itemize}

Table \ref{tab:table3} presents a summary of all multimodal papers where CT, MRI, PET and/or CBCT were used. The papers \cite{Zhang2019183, Zhang20189242, Cai2019174} are interconnected and should therefore be read together for better understanding.

\textbf{List 2: List of abbreviations of Table \ref{tab:table3}:}
\begin{itemize}
\small
\setlength\itemsep{-0.5em}
\item AC — Attenuation Correction;
\item BMGAN — Bidirectional Mapping GAN;
\item CAE-ACGAN — Conditional Auto-Encoding, Auxiliary Classifier GAN;
\item CMD — Center of Mass Distance; 
\item GMD — Gradient Magnitude Distance;
\item FGAN — Feature-consistent GAN;
\item hGAN — hybrid GAN (3D Generator and 2D Discriminator);
\item MSD — Mean Surface Distance;
\item MS-SSIM — Multi-Scale Structural Similarity Index Measure; 
\item PCC — Pearson Correlation Coefficient;
\item PVD — Percentage Volume Difference;
\item res-cycle GAN — residual block cycle-consistent GAN;
\item RevGAN — Reversible GAN;
\item RMSD — Residual Mean Square Distance; 
\item SNU — Spatial Non-Uniformity;
\item S-score — Shape-score;
\end{itemize}
Table \ref{tab:table4} contains the papers on the other modalities, such as CAD, RGB-D, FIB-SEM, microscopy, seismic and synthetic. The papers \cite{Ho2019, Han2019, Chen2021961} have some authors in common and complement each other. Note that \cite{Muzahid202120} give the same name to their GAN as the point cloud GAN, i.e. PC-GAN, but these networks are different, as voxels and not point clouds are used.

\textbf{List 3: List of abbreviations of Table \ref{tab:table4}:}
\begin{itemize}
\small
\setlength\itemsep{-0.5em}
\item CC — Correlation Coefficient;
\item cryo-EM — cryo-Electron Microscopy;
\item diSPIM — dual inverted Selective Plane Illumination Microscope;
\item DLGAN — Depth-preserving Latent GAN;
\item DPB — Double Phase Boundary;
\item FIB — Focused Ion Beam;
\item KPFM — Kelvin Probe Force Microscopy;
\item LSFM — Light Sheet fluorescence Microscopy;
\item MAPE — Mean Absolute Percentage Error;
\item MP-GAN — Multi-Projection GAN;
\item MSG-GAN — Multi-Scale Gradient GAN;
\item NRMSE — Normalized Root Mean Square Error;
\item PSF — Point Spread Functions;
\item RecGAN — Reconstruction GAN;
\item SpCycleGAN — Spatially Constrained CycleGAN;
\item TPB — Triple Phase Boundary;
\item Voxel/ PC — Voxelized point clouds;
\end{itemize}

\onecolumn
\begin{table*}[!t]
\begin{small}

\begin{center}
\caption{Compact overview of reviewed studies with modalities CT and MRI, sorted by modality and then by year of publication, indicating the modality, and the datasets used with the respective localisation, if available. It also shows the network used, the loss functions, the evaluation metrics, the resolution of the synthetic data generated, and whether or not it is a medical application. The abbreviations used in this table can be found in List 1 or in section 1.1.3 (Acronyms and Abbreviations). "Modal." stands for Modality}
\label{tab:table2}
\begin{tabular}{|p{2.2cm}|p{1cm}|p{0.7cm}|p{2.5cm}|p{2cm}|p{2cm}|p{3cm}|p{2cm}|}
\hline
\textbf{Study} & \textbf{Modal.} & \textbf{Med} & \textbf{Dataset} & \textbf{Network (G/D)} & \textbf{Loss function} & \textbf{Metrics} & \textbf{Comments} \\ \hline

\cite{Bu2021670} & CT & Yes & LUNA16 $^a$ & cGAN (3D-UNet / 3D CNN) & Adv & VTT, (Sen, CPM) & $32$×$32$×$32$ \\ \hline

\cite{Pesaranghader202167} & CT & Yes & NLST $^b$, LIDC $^c$ & CT-SGAN & JS, WGAN-GP & Visual, FID,  IS, (Acc) & $\textgreater{}= 224$×$224$×$224$ (stacks $224$×$224$×$3$) \\ \hline

\cite{Zhang2021} & CT & Yes & 120 intact prostate cancer patients & PGGAN & Adv, CE & HU (DSC, Average HD, Average Surface HD) & $64$×$128$×$128$ \\ \hline

\cite{Chen2020} & CT & No & Berea Sandstone ct scan & WGAN with two discriminators & WGAN, Minkowski functional & Minkowski functional & $64$×$64$×$64$ \\ \hline

\cite{Gayon-Lombardo2020}$^d$ & CT (XCT) & No & Li-ion battery cathode$^e$, SOFC anode$^f$ & DCGAN & Adv & Phase volume fraction, SSA, TPB, TPCF, Relative diffusivity & $64$×$64$×$64$ \\ \hline

\cite{Liu20196164} & CT & No & Berea sandstone$^g$, Estaillades carbonate$^h$ & DCGAN & Adv & Visual & $64$×$64$×$64$, $128$×$128$×$128$ \\ \hline

\cite{Krutko2019} & CT & No & Sandstone$^i$ & SPGAN (3D DCGAN based) & Adv & Minkowski functional, Absolute permeability & $64$×$64$×$64$ \\ \hline

\cite{Moghari2019} & CT (CTP) & Yes & 18 acute stroke patients & cGAN & Adv & PSNR, NMSE, SSIM & $64$×$64$×$64$ \\ \hline

\cite{Han2019729} & CT & Yes &LIDC$^c$ & MCGAN (U-Net/Pix2Pix) & LSGAN, WGAN-GP & VTT, t-SNE, (CPM) & $32$×$32$×$32$ \\ \hline

\cite{Xu201962} & CT & Yes & LIDC$^c$  & MCGAN & LSGAN, MAE, MSE & Visual & $64$×$64$×$64$ \\ \hline

\cite{Mosser2017}$^j$ & Micro-CT & No & Spherical Bead pack, Berea sandstone, oolitic, Ketton limestone$^k$  & DCGAN & Adv & Minkowski functional & $64$×$64$×$64$ \\ \hline

\cite{Dikici2021} & MRI & Yes & 217 post-gadolinium T1-weighted  & cGANe (DCGAN based) & Adv & FID, (AFP) & $16$×$16$×$16$ \\ \hline

\cite{Momeni2021} & MRI & Yes & AIBL$^l$, MICCAI Valdo 2021$^m$  & conditional LesionGAN (cGAN based) & Adv, Volume, Border & (AUC, Sen, ROC) & $11$×$11$×$11$ \\ \hline

\cite{Rusak202011} & MRI & Yes & ADNI$^n$  & GAN (Pix2Pix based) & Adv, MAE  & PSNR, SSIM, MAE, MSE, DSC & $181$×$218$×$181$ \\ \hline

\end{tabular}
\end{center}
\end{small}
\end{table*}

\setcounter{table}{0} 

\begin{table*}[!t]
\begin{small}

\begin{center}
\caption{Continuation of the previous Table \ref{tab:table2} - Compact overview of reviewed studies with modalities CT and MRI, sorted by modality and then by year of publication, indicating the modality, and the datasets used with the respective localisation, if available. It also shows the network used, the loss functions, the evaluation metrics, the resolution of the synthetic data generated, and whether or not it is a medical application. The abbreviations used in this table can be found in List 1 or in section 1.1.3 (Acronyms and Abbreviations). "Modal." stands for Modality}

\begin{tabular}{|p{2cm}|p{1cm}|p{0.7cm}|p{1.8cm}|p{2.5cm}|p{2.5cm}|p{2.5cm}|p{1.8cm}|}\hline

\textbf{Study} & \textbf{Modal.} & \textbf{Med} & \textbf{Dataset} & \textbf{Network (G/D)} & \textbf{Loss function} & \textbf{Metrics} & \textbf{Comments} \\ \hline

\cite{Jung202079} & MRI & Yes & ADNI$^n$  & CycleGAN (cGAN based) & WGAN-GP, CE, Cycle consistency & FID, (Acc) & $192$×$192$× $(3,6,8)$ stacks $192$×$192$×$1$   \\ \hline

\cite{Ma2020} & MRI & Yes & Kulaga-Yoskovitz $^o$  & da-GAN  & Adv, Gradient Difference, Frequency domain, MAE & Visual, PSNR, SSIM, SIS, (DSC) & $384$×$512$×$384$ (stacking $512$ imgs of $384$×$384$) \\ \hline

\cite{Zhuang2019} & MRI & Yes & OpenfMRI$^p$, IBC$^q$  & ICW-GAN & WGAN-GP & (Acc, Macro F1, Pre, Sen) & $53$×$63$×$46$ \\ \hline

\cite{Baumgartner20188309}$^r$ & MRI & Yes & ADNI$^n$  &  VA-GAN (WGAN based 3D U-Net / C3D) & WGAN-GP, MAE & Visual, NCC & $128$×$160$×$112$ \\ \hline

\cite{Li2020} & OCT & Yes & 3D ODT of mouse CBF  & GAN (N/D) & WGAN & (PSNR, DSC, Noise SD) & $64$×$64$, $64$×$64$×$64$ \\ \hline

\multicolumn{8}{l}{
$^a$\href{https://luna16.grand-challenge.org/}{LUNA16}; 
$^{b}$ \href{https://www.cancer.gov/types/lung/research/nlst}{NLST};
$^{c}$ \href{https://pubmed.ncbi.nlm.nih.gov/21452728/}{LIDC};
$^{d}$ \href{https://github.com/agayonlombardo/pores4thought}{Pores for thought};
$^{e}$ \href{https://iopscience.iop.org/article/10.1149/2.0981813jes}{Li-ion battery cathode};
$^{f}$ \href{https://www.osti.gov/biblio/10182383-two-point-correlation-functions-characterize-microgeometry-estimate-permeabilities-synthetic-natural-sandstones}{SOFC anode};
$^{g}$ \href{https://pubmed.ncbi.nlm.nih.gov/19905212/}{Berea sandstone};
} \\

\multicolumn{8}{l}{
$^{h}$ \href{https://www.imperial.ac.uk/earth-science/research/research-groups/pore-scale-modelling/micro-ct-images-and-networks/}{Estaillades carbonate (wrong in the paper)};
$^{j}$ \href{https://github.com/LukasMosser/PorousMediaGan}{PorousMediaGAN};
$^{k}$ \href{https://www.imperial.ac.uk/earth-science/research/research-groups/pore-scale-modelling/micro-ct-images-and-networks/}{Micro-CT Images and Networks};
$^{l}$ \href{https://doi.org/10.25919/aegy-ny12}{AIBL};
} \\

\multicolumn{8}{l}{
$^{m}$ \href{https://valdo.grand-challenge.org/}{MICCAI Valdo 2021};}\\

\multicolumn{8}{l}{$^{i}$ Low-permeable sandstone from a hydrocarbon reservoir located in Russia rotationally scanned X-ray L240 GE system;
} \\

\multicolumn{8}{l}{
$^{n}$ \href{https://adni.loni.usc.edu/}{ADNI};
$^{o}$ \href{http://www.nitrc.org/projects/mni-hisub25}{Kulaga-Yoskovitz};
$^{p}$ \href{https://neurovault.org/collections/1952/}{OpenfMRI};
$^{q}$ \href{https://project.inria.fr/IBC/}{IBC};
$^{r}$ \href{https://github.com/baumgach/vagan-code}{VA-GAN}
}

\end{tabular}
\end{center}
\end{small}
\end{table*}

\begin{table*}[!t]
\begin{small}

\begin{center}
\caption{Compact overview of reviewed studies with multimodalities CT, MRI, PET and/or CBCT, sorted by modality and then by year of publication, indicating the modality, and the datasets used with the respective localisation, if available. It also shows the network used, the loss functions, the evaluation metrics, the resolution of the synthetic data generated. All studies are medical applications. The abbreviations used in this table can be found in List 3 or in section 1.1 (Acronyms and Abbreviations). "Modal." stands for Modality}
\label{tab:table3}
\begin{tabular}{|p{2.4cm}|p{1cm}|p{2.5cm}|p{2.5cm}|p{2.5cm}|p{2.8cm}|p{1.8cm}|}
\hline
\textbf{Study} & \textbf{Modal.}  & \textbf{Dataset} & \textbf{Network (G/D)} & \textbf{Loss function} & \textbf{Metrics} & \textbf{Comments} \\ \hline

\cite{Qin2021S98} & CT, CBCT  & 27 advanced lung cancer & GAN (Residual-UNet/CNN) & Adv & MAE & N/D \\ \hline

\cite{Wei2020} & CT, CBCT  & 4D-CT and raw CBCT   projections & cGAN & Adv, MAE & Acc, (Tumour localisation error) & $128$×$128$×$32$ \\ \hline

\cite{Harms2020} & CT, CBCT  & 24 brain and 20 pelvic CT/CBCT  & res-cycle GAN & Adv, Cycle consistency, L1.5, GMD & MAE, PSNR, NCC, SNU & $96$×$96$×$5$ (stacked to produce 3D) \\ \hline

\cite{Lei2020b} & CT, PET  & 16 CT/PET & CycleGAN & Adv, MSE & Visual, MAE, NMSE, NCC, PSNR & $64$×$64$×$64$ \\ \hline

\cite{Gu2021} & MRI, CT  & BrainWeb$^a$ & Dual3D\& PatchGAN (3DGAN) & Adv & SSIM, PSNR & N/D \\ \hline

\cite{Yang2021415} & MRI, CT  & 9 healthy MRI$^b$ & CAE-ACGAN & Adv, MAE & Visual, PSNR, SSIM, MAE & $128$×$128$×$32$ \\ \hline

\cite{Liu2020} & MRI, CT  & 21 cancer hepatocellular  & dense CycleGAN & Adv, L1.5, Gradient Difference & MAE, PSNR, NCC, HU & $64$×$64$×$64$ \\ \hline

\cite{Lei2020a} & MRI, CT  & 45 prostate cancer MRI/CT  & CycleGAN & Adv, Cycle consistency & (DSC, Sen, Spe, HD, MSD, RMSD, CMD, PVD) & $64$×$64$×$64$ \\ \hline

\cite{Yang2019181} & MRI, CT  & N/D & N/D (3D-UNet/3D CNN) & Adv, MSE & PSNR, MSE & N/D \\ \hline

\cite{Zeng2019759} & MRI, CT  &  50 subjects MRI/CT & hGAN (CycleGAN based) & Adv, MAE & Visual, MAE, PSNR & $256$×$288$×$32$, $256$×$288$×$12$ \\ \hline

\cite{Zhang2019183} & MRI, CT  & 4496 cardiovascular MRI/CT  & CycleGAN & Adv, Cycle and Shape consistency & Visual, S-score, (DSC) & $112$×$112$×$86$ \\ \hline

\cite{Zhang20189242} & MRI, CT  & 4354 contrasted cardiac CT scans & CycleGAN & Adv, Cycle and Shape consistency & Visual, S-score, (DSC) & $86$×$112$×$112$ \\ \hline

\cite{Schaefferkoetter20213817} & MRI, CT, PET  & 60 patients & CycleGAN (Residual-UNet/ patchGAN) & MAE, Cycle consistency, Identity, MSE & (MSE, PCC, AC) & $96$×$96$×$96$ \\ \hline

\cite{Cai2019174} & MRI, CT, X-rays  & Several datasets$^c$ & CycleGAN (cGAN/ PatchGAN) & LSGAN, Cycle and Shape consistency & Visual, S-score, (DSC) & $80$×$128$×$128$ \\ \hline

\end{tabular}
\end{center}
\end{small}
\end{table*}

\setcounter{table}{1} 

\begin{table*}[!t]
\begin{small}

\begin{center}
\caption{Continuation of the previous Table \ref{tab:table3}. Compact overview of reviewed studies with multimodalities CT, MRI, PET and/or CBCT, sorted by modality and then by year of publication, indicating the modality, and the datasets used with the respective localisation, if available. It also shows the network used, the loss functions, the evaluation metrics, the resolution of the synthetic data generated. All studies are medical applications. The abbreviations used in this table can be found in List 3 or in section 1.1 (Acronyms and Abbreviations). "Modal." stands for Modality}
\label{tab:table3}
\begin{tabular}{|p{2cm}|p{1cm}|p{2.5cm}|p{2.5cm}|p{2.5cm}|p{2.8cm}|p{1.8cm}|}
\hline
\textbf{Study} & \textbf{Modal.}  & \textbf{Dataset} & \textbf{Network (G/D)} & \textbf{Loss function} & \textbf{Metrics} & \textbf{Comments} \\ \hline

\cite{Hu2022145} & MRI, PET  & ADNI$^d$ & BMGAN(Dense-UNet/Patch-Level) & LSGAN, KL-divergence constraint, MAE, Perceptual (VGG-16 $^e$) & MAE, PSNR, MS-SSIM, FID & $128$×$128$×$128$ \\ \hline

\cite{Lin2021}$^f$ & MRI, PET  & ADNI$^d$ & RevGAN & Adv, Cycle consistency & SSIM, PSNR & $96$×$96$×$48$ \\ \hline

\cite{Pan2019137} & MRI, PET  & ADNI$^d$& FGAN & Adv, Feature-Consistent & MAE, PSNR, SSIM, (AUC) & N/D \\ \hline

\cite{Yan2018} & MRI, PET  & ADNI$^d$ & cGAN & Adv, MAE & Visual, SSIM, (Acc, AUC) & $160$×$160$×$96$ \\ \hline

\multicolumn{7}{l}{
$^a$\href{https://brainweb.bic.mni.mcgill.ca}{BrainWeb}; 
$^{b}$ \href{https://link.springer.com/article/10.1007/s11036-020-01678-1\#data-availability}{Data and code available upon request};
$^{d}$ \href{https://adni.loni.usc.edu/}{ADNI};
$^{e}$ \cite{Hu2022145};
$^{f}$ \href{https://www.frontiersin.org/articles/10.3389/fnins.2021.646013/full\#h9}{Code available upon request};
} \\

\multicolumn{7}{l}{$^{c}$  \href{https://www.ncbi.nlm.nih.gov/pmc/articles/PMC3824915/}{4354/142 contrasted cardiac CT/MRI},  \href{https://wiki.cancerimagingarchive.net/display/Public/Pancreas-CT}{82/78 pancreatic abdomen CT/MRI scans}, mammography X-rays (\href{https://studylib.net/doc/8826147/bcdr--a-breast-cancer-digital-repository}{BCDR},  \href{https://pubmed.ncbi.nlm.nih.gov/22078258/}{INbreast})}

\end{tabular}
\end{center}
\end{small}
\end{table*}

\begin{table*}[!t]
\begin{small}

\begin{center}
\caption{Compact overview of reviewed studies with other modalities, sorted by modality and then by year of publication, indicating the modality, and the datasets used with the respective localisation, if available. It also shows the network used, the loss functions, the evaluation metrics, the resolution of the synthetic data generated, and whether or not it is a medical application. The abbreviations used in this table can be found in List 4 or in section 1.1 (Acronyms and Abbreviations)}
\label{tab:table4}

\begin{tabular}{|p{1.6cm}|p{1.7cm}|p{0.7cm}|p{3.2cm}|p{2cm}|p{2cm}|p{2.2cm}|p{2cm}|}
\hline

\textbf{Study} & \textbf{Modality} & \textbf{Med} & \textbf{Dataset} & \textbf{Network (G/D)} & \textbf{Loss function} & \textbf{Metrics} & \textbf{Comments} \\ \hline

\cite{Muzahid202120} & CAD & No & ModelNet10/40$^a$  & Progressive Conditional GAN (PGGAN/cGAN based) & Adv & Visual, Clustering, (Acc) & $32$×$32$×$32$ \\ \hline

\cite{greminger2020generative} & CAD & No & Synthetic data generated by CadQuery$^b$  & MSG-GAN & Hinge & Compliance, Fraction of Unmachinable Voxels & $32$×$32$×$32$ \\ \hline

\cite{Kniaz20203}$^c$ & CAD & No & SyntheticVoxels$^d$, VoxelCity$^e$ & Z-GAN (pix2pix based) & Adv, MAE & IoU & $128$×$128$×$128$, Voxel/ PC \\ \hline

\cite{Li20195530} & CAD & No & ShapeNet$^f$, Pix3D$^g$, Stanford car$^h$, CUB-Birds-200-2011$^i$  & MP-GAN & Adv & FID, Acc & $64$×$64$×$64$ \\ \hline

\cite{Yang2017679}$^j$ & CAD & No & ModelNet40$^a$ & 3D-RecGAN & Modified CE, WGAN-GP & IoU, CE & $64$×$64$×$64$, Voxel/ PC \\ \hline

\cite{Wang20172317}$^k$ & CAD, RGB-D & No & Real-world scans$^l$, ModelNet10/40$^a$  &  ED-GAN (cGAN based) & Adv, CE, (MAE) & (CE, Acc) & $32$×$32$×$32$, $128$×$128$×$128$ \\ \hline

\cite{Sciazko20211363} & FIB-SEM & No & Ni70GDC30, Ni30GDC70, Ni70GDC30(200MPa), Ni30GDC70(200MPa) & GAN2Dto3D & Adv & Average volume fractions, Volume fraction variations, TPB/DPB densities & $64$×$64$×$64$ \\ \hline

\cite{Gupta2021759}$^m$ & Microscopy (cryo-EM) & Yes & EMPIAR-10061$^n$, EMPIAR-10028$^o$  & CryoGAN (cryo-EM physics simulator) & WGAN-GP & Reconstruction resolution ({\AA}) & $180$×$180$×$180$ \\ \hline

\cite{Chen2021961} & Microscopy & Yes & Human kidney  & SpCycleGAN & Adv, Cycle consistency, Spatial & (DSC, Type-I/Type-II error, Acc, IoU, Pre, Sen, F1) & $128$×$128($×$128)$ Pos-processing \\ \hline

\cite{Tang20201775} & Microscopy & Yes & Janelia-Fly/Tokyo-Fly$^p$  & cGAN & Adv, MAE & PSNR, SSIM & $128$×$128$×$32$ \\ \hline

\cite{Ho2019} (\cite{Fu2018}) & Microscopy (Fluorescence, two-photon)  & Yes & Rat kidney  & SpCycleGAN & Adv, Cycle consistency, Spatial & (Pre, Sen, F1) & $128$×$128$×$128$ \\ \hline

\end{tabular}
\end{center}
\end{small}
\end{table*}

\setcounter{table}{2} 
\begin{table*}[!t]
\begin{small}

\begin{center}
\caption{Continuation of the previous Table \ref{tab:table4} - Compact overview of reviewed studies with other modalities, sorted by modality and then by year of publication, indicating the modality, and the datasets used with the respective localisation, if available. It also shows the network used, the loss functions, the evaluation metrics, the resolution of the synthetic data generated, and whether or not it is a medical application. The abbreviations used in this table can be found in List 4 or in section 1.1 (Acronyms and Abbreviations)}

\begin{tabular}{|p{2cm}|p{1.7cm}|p{0.7cm}|p{2.9cm}|p{2cm}|p{2cm}|p{2cm}|p{2cm}|}
\hline

\textbf{Study} & \textbf{Modality} & \textbf{Med} & \textbf{Dataset} & \textbf{Network (G/D)} & \textbf{Loss function} & \textbf{Metrics} & \textbf{Comments} \\ \hline

\cite{Han2019} & Microscopy (Fluorescence, two-photon) & Yes & Rat kidney  & SpCycleGAN & Adv, Cycle consistency, Spatial, (MSE) & (MAPE, F1, Sen, Pre) & $128$×$128$×$128$, $128$×$128$×$32$, $64$×$64$×$64$ \\ \hline

\cite{Baniukiewicz2019}$^q$ & Microscopy (diSPIM light sheet, LSFM, Confocal microscope) & Yes & Cells (several sources$^r$) & cGAN &  Adv, MAE & Visual, (F1, Sen, Pre) & $256$×$256$×$66$ (stack of $66$ images $256$×$256$) \\ \hline

\cite{Liu20212843} & RGB-D & No & ModelNet40$^a$, KinectData$^s$  & DLGAN (ED-GAN) & Adv, Latent vector, Depth & IoU, CE & $64$×$64$×$64$ \\ \hline

\cite{Liu2020O47} & Seismic reflection data & No & Synthetic model, F3 block seismic data$^t$  & semi-supervised GANs & Adv, MSE, (CE) & (Acc, F1) & $64$×$64$×$64$ \\ \hline

\cite{Kench2021299}$^u$ & Synthetic, x-ray, KPFM, SEM & No & Several materials$^v$  & SliceGAN & WGAN & Visual, Volume fraction, Relative surface area, Relative diffusivity & $64$×$64$×$64$ \\ \hline

\cite{Yang2021130}$^w$ & Synthetic (Data simulator, PSF$^x$) & Yes & Synthetic isotropic quad-view embryo and \textit{C. elegans} $^y$  & CycleGAN based (CNN based/multi-scale) & LSGAN, Cycle consistency, MSE & NRMSE, PSNR, SSIM, CC  & $64$×$64$×$64$ \\ \hline

\cite{Nozawa2021} & Synthetic & No & 3D car meshes$^z$, Contour sketches & cGAN & Adv, MAE, CE & Visual & $64$×$64$xViews (reconstruction using several views), Voxel/ PC \\ \hline

\end{tabular}
\end{center}
\end{small}
\end{table*}

\setcounter{table}{2} 
\begin{table*}[!t]
\begin{small}

\begin{center}
\caption{Continuation of the previous Table \ref{tab:table4} - Compact overview of reviewed studies with other modalities, sorted by modality and then by year of publication, indicating the modality, and the datasets used with the respective localisation, if available. It also shows the network used, the loss functions, the evaluation metrics, the resolution of the synthetic data generated, and whether or not it is a medical application. The abbreviations used in this table can be found in List 4 or in section 1.1 (Acronyms and Abbreviations)}

\begin{tabular}{|p{1.6cm}|p{1.7cm}|p{0.7cm}|p{2.9cm}|p{2cm}|p{2cm}|p{2cm}|p{2cm}|}
\hline

\textbf{Study} & \textbf{Modality} & \textbf{Med} & \textbf{Dataset} & \textbf{Network (G/D)} & \textbf{Loss function} & \textbf{Metrics} & \textbf{Comments} \\ \hline

\cite{Shen20213250} & Synthetic & No & 653 3D Strand-level models$^{aa}$ & WGAN-GP & WGAN-GP, Content, Style, Laplacian, Projection, Orientation & Visual, MSE & $128$×$128$×$96$ \\ \hline

\cite{Danu2019662} & Synthetic & Yes & 2D/3D vessel-like structures, 3D stenosed blood vessel segments & GAN (N/D) & Adv & Visual & $128$×$128$, $64$×$64$×$64$, $128$×$32$×$32$, Meshes /Voxel \\ \hline

\cite{Halpert20192081} & Synthetic (Earth Model) & No & SEAM Phase I synthetic model$^{ab}$ & DCGAN & Adv & Visual & $32$×$32$×$32$ \\ \hline

\multicolumn{8}{l}{
$^a$ \href{https://modelnet.cs.princeton.edu/}{ModelNet10/40}; 
$^{b}$ \href{https://github.com/CadQuery/cadquery}{CadQuery};
$^{c}$ \href{https://github.com/vlkniaz/Z\_GAN}{Z-GAN};
$^{d}$ \href{http://www.zefirus.org/SyntheticVoxels}{SyntheticVoxels};
$^{e}$ \href{http://www.zefirus.org/Z\_GAN}{VoxelCity};
$^{f}$ \href{https://arxiv.org/pdf/1512.03012.pdf}{ShapeNet};
$^{g}$ \href{https://openaccess.thecvf.com/content\_cvpr\_2018/papers/Sun\_Pix3D\_Dataset\_and\_CVPR\_2018\_paper.pdf}{Pix3D};
} \\

\multicolumn{8}{l}{
$^{h}$ \href{http://ai.stanford.edu/~jkrause/cars/car\_dataset.html}{Stanford car};
$^{i}$ \href{https://authors.library.caltech.edu/27452/1/CUB\_200\_2011.pdf}{CUB-Birds-200-2011};
$^{j}$ \href{https://github.com/Yang7879/3D-RecGAN}{3D-RecGAN};
$^{k}$ \href{https://github.com/Fdevmsy/3D\_shape\_inpainting}{3D Shape Completion};
$^{l}$ \href{https://arxiv.org/pdf/1604.03265.pdf}{Real-world scans};
} \\

\multicolumn{8}{l}{
$^{m}$ \href{https://github.com/harshit-gupta-cor/CryoGAN}{CryoGAN};
$^{n}$ \href{https://pubmed.ncbi.nlm.nih.gov/25953817/}{EMPIAR-10061};
$^{o}$ \href{https://www.sciencedirect.com/science/article/pii/S0969212618301291}{EMPIAR-10028};
$^{p}$ \href{https://www.ncbi.nlm.nih.gov/pmc/articles/PMC4725298/}{Janelia-Fly/Tokyo-Fly};
$^{q}$ \href{https://pilip.lnx.warwick.ac.uk/Frontiers_2019/Net_3D/Figure\%205/pix2pix\_3D\_multichannel.py}{pix2pix\_3D\_multichannel};
} \\

\multicolumn{8}{l}{
$^{r}$ Sources: \href{https://pubmed.ncbi.nlm.nih.gov/24616222/}{here}, \href{https://doi.org/10.1073/pnas.1322291111}{here}, \href{https://pubmed.ncbi.nlm.nih.gov/28751725/}{here};
$^{s}$ \href{http://kinectdata.com/}{KinectData};
$^{t}$ \href{https://drive.google.com/drive/folders/0B7brcf-eGK8CRUhfRW9rSG91bW8}{F3 block seismic data};
$^{u}$ \href{https://github.com/stke9/SliceGAN}{SliceGAN};
$^{v}$ \href{https://www.nature.com/articles/s42256-021-00322-1/tables/3}{See their Table 2};
} \\

\multicolumn{8}{l}{
$^{w}$ \href{https://github.com/stegmaierj/MultiViewFusion}{MultiViewFusion};
$^{x}$ \href{https://www.nature.com/articles/nmeth.2929}{PSF};
$^{y}$ \href{https://www.nature.com/articles/nmeth.2929}{Synthetic isotropic quad-view embryo and C. elegans};
$^{z}$ \href{https://shapenet.org/}{3D car meshes};
}\\

\multicolumn{8}{l}{
$^{aa}$ Strand-level models: \href{http://www.hao-li.com/publications/papers/siggraphAsia2018HSUVVA.pdf}{here}, and \href{http://www.kunzhou.net/2016/autohair.pdf}{here};$^{ab}$ \href{https://library.seg.org/doi/book/10.1190/1.9781560802945}{SEAM Phase I synthetic model};
}

\end{tabular}
\end{center}
\end{small}
\end{table*}

\twocolumn

\subsection{Visual results}
\label{sec:visual_results}
As explained in Appendix \ref{subsub:visual}, visual judgement is currently the most widely used evaluation tool and one of the best for assessing the images produced by GANs.  Therefore, almost every publication presents some real and synthetic images that allow the reader to check the realism of the synthetic data. However, in some cases, it is very difficult to assess this realism because the 2D rendering processes of the paper can change the colour, resolution, and other properties of the images, e.g. MRI or CT scans. Some authors make the trained generator or some synthetic data freely available, but others do not, making it difficult for readers to perform visual assessments.

Fortunately, it is usually possible to check the improvements only on the basis of the images provided, as the authors are aware of these limitations. In this section, we present some selected images from some reviewed papers, as they can best reflect the realism of the generated synthetic data. We have selected images related to different organs (medical domain) and different structures (non-medical domain). 

\begin{figure}
  \centering
  \includegraphics[width=0.7\columnwidth]{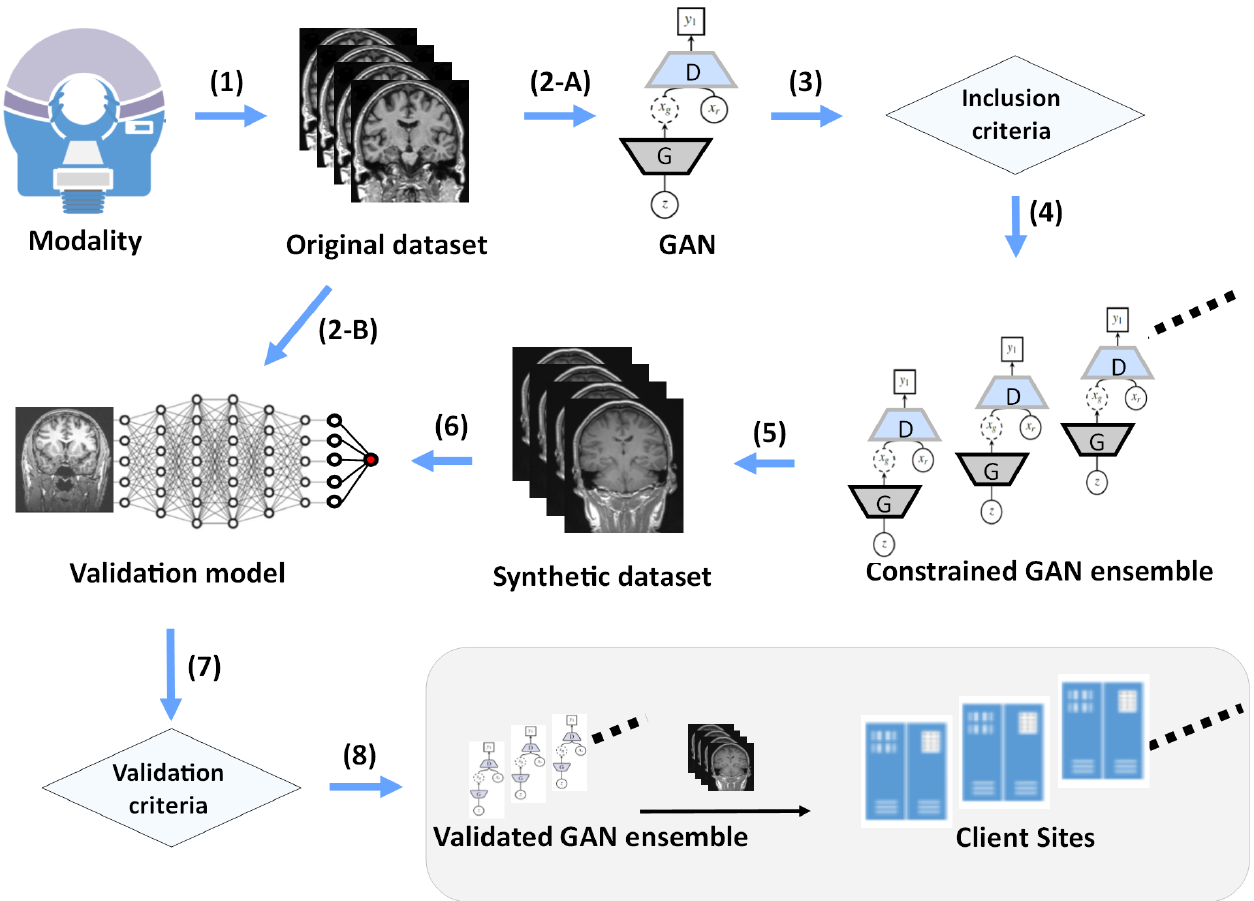}
  \caption{Constrained GAN ensemble for data sharing \cite{Dikici2021}.  Figure reproduced from  \cite{Dikici2021}. \href{https://arxiv.org/pdf/2003.00086.pdf}{Publication link}.
  \label{Fig:Rev:brain}}
\end{figure}

Figure \ref{Fig:Rev:brain} shows the cGANe system developed by \cite{Dikici2021}. They offer a novel protocol for sharing data by generating synthetic data from the original dataset, validating it and making it available at client sites. This inherent anonymization capacity of GANs was already explored by \cite{shin2018medical}. 

\begin{figure}
  \centering
  \includegraphics[width=0.7\columnwidth]{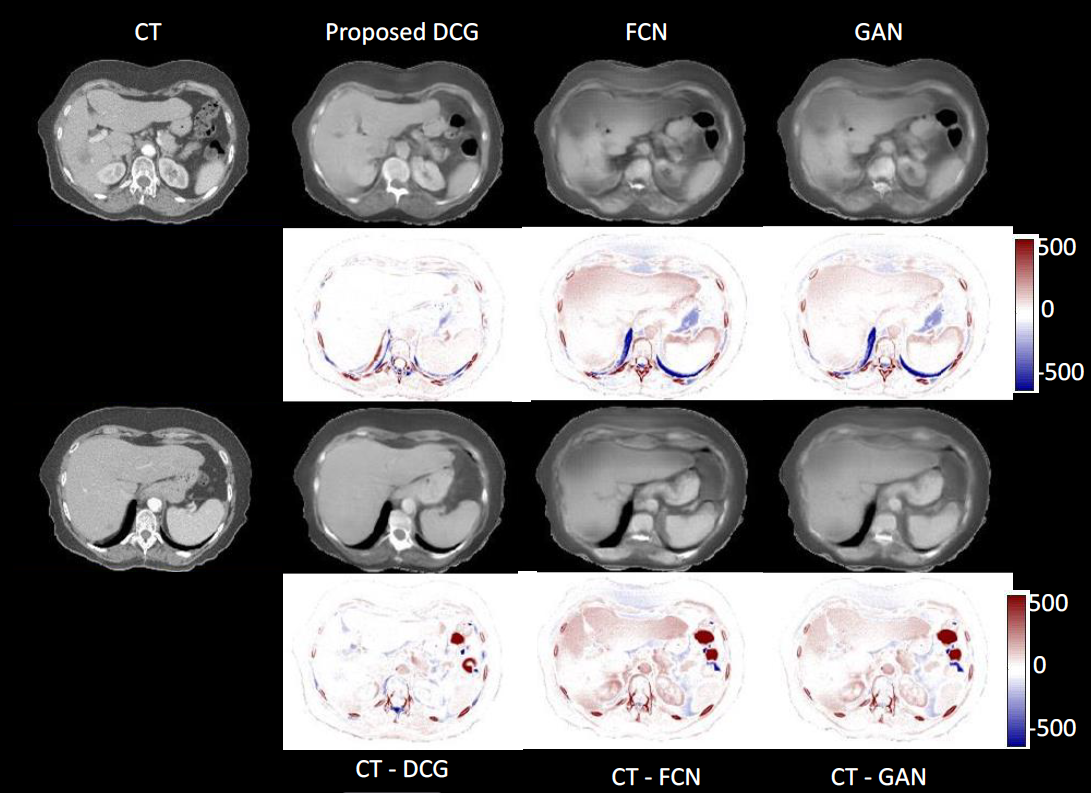}
  \caption{Original axial CT images and synthetic CT images generated by dense-cycle-GAN (DCG) developed by \cite{Liu2020}, 3D Fully Convolutional Neural network (FCN) and  vanilla GAN. The second and last rows are the  difference image between the original CT and the generated CT images. Reprinted from Medical Imaging 2020: Image Processing, Vol. 11313, authors Yingzi Liu and Yang Lei and Tonghe Wang and Jun Zhou and Liyong Lin and Tian Liu and Pretesh Patel and Walter J Curran and Lei Ren and Xiaofeng Yang, "Liver synthetic CT generation based on a dense-CycleGAN for MRI-only treatment planning", pages 659-664, Copyright (2020), with permission from Society of Photo-Optical Instrumentation Engineers (SPIE) and one of the authors. \href{https://doi.org/10.1117/12.2549265}{Publication link}.
  \label{Fig:Rev:liver}}
\end{figure}

\begin{figure}
  \centering
  \includegraphics[width=1\columnwidth]{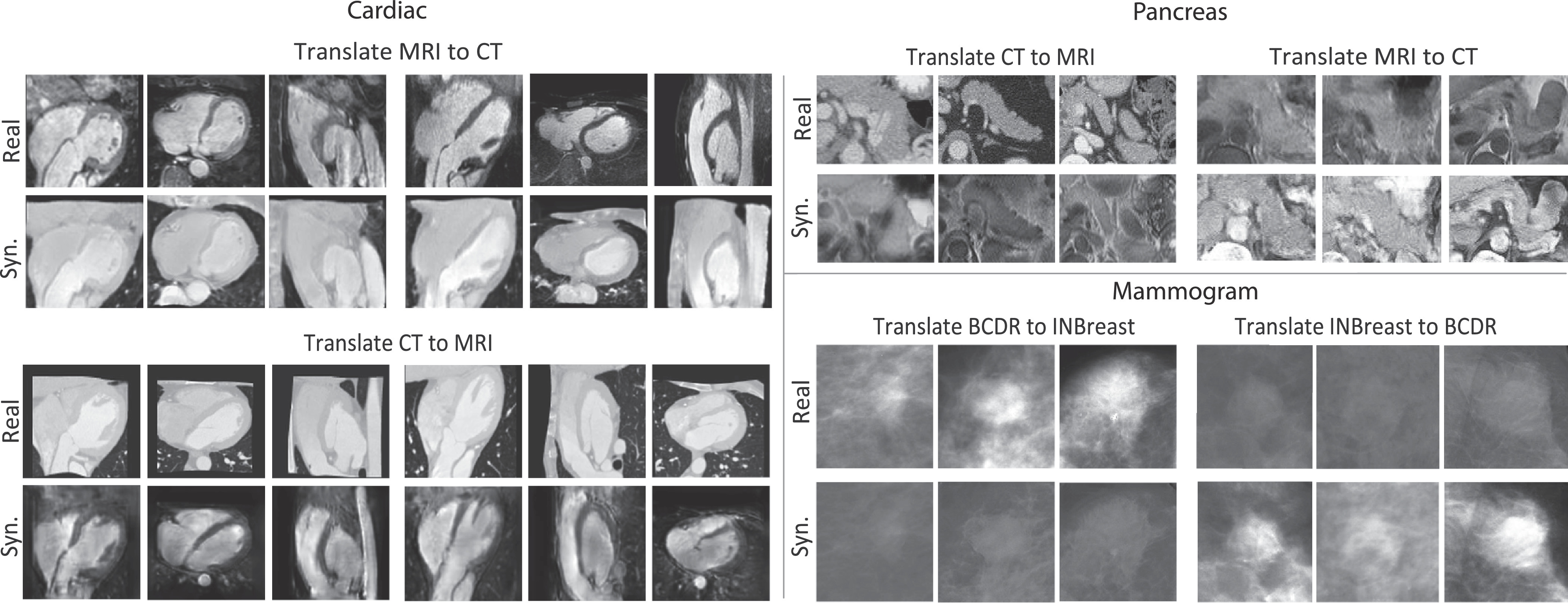}
  \caption{On the left side three orthogonal sections through the centre of the 3D cardiac volumes are shown. The top right displays the pancreatic volumes of the maximum cross-sectional area. The bottom right shows real and synthetic mammographic lesion patches \cite{Cai2019174}. Reprinted from Medical Image Analysis, Vol. 52, authors Jinzheng Cai and Zizhao Zhang and Lei Cui and Yefeng Zheng and Lin Yang, "Towards cross-modal organ translation and segmentation: A cycle- and shape-consistent generative adversarial network", pages 174-184, Copyright (2018), with permission from Elsevier B.V.. \href{https://doi.org/10.1016/j.media.2018.12.002}{Publication link.}}
  \label{Fig:Rev:cardiac_pancreas_mammogram}
\end{figure}

\begin{figure}
  \centering
  \includegraphics[width=1\columnwidth]{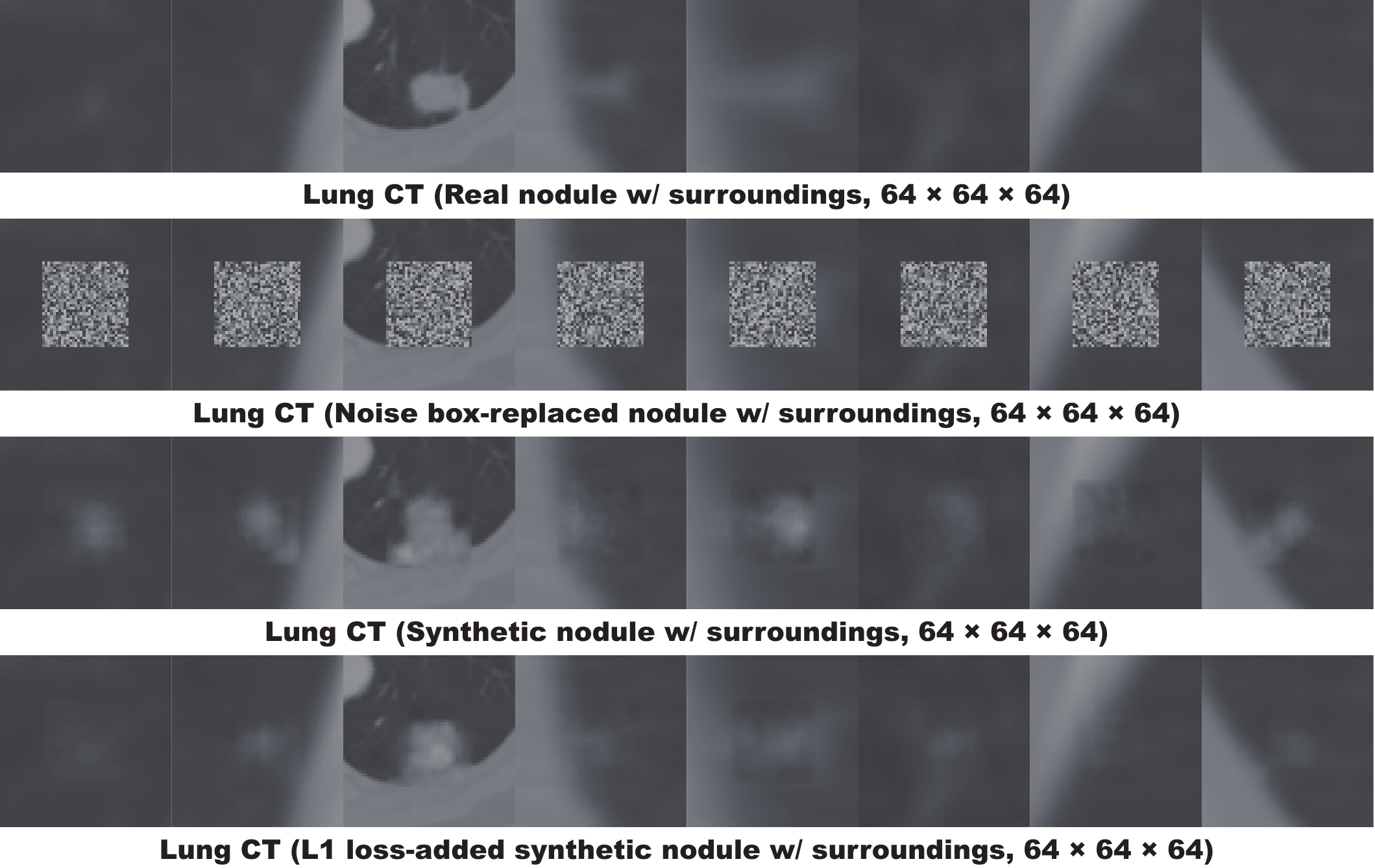}
  \caption{2D axial view of: First row — Real nodules; Second row — Lung nodules replaced by noise of dimension $32$×$32$×$32$; Third row — Synthetic nodules with surrounding tissue; Fourth row — Synthetic nodules using L\textsubscript{1} loss with surrounding tissue \cite{Han2019729}. © 2019 IEEE. Reprinted, with permission, from 2019 International Conference on 3D Vision (3DV), authors Changhee Han and Yoshiro Kitamura and Akira Kudo and Akimichi Ichinose and Leonardo Rundo and Yujiro Furukawa and Kazuki Umemoto and Yuanzhong Li and Hideki Nakayama, "Synthesizing Diverse Lung Nodules Wherever Massively: 3D Multi-Conditional GAN-Based CT Image Augmentation for Object Detection", pages 729-737. \href{https://doi.org/10.1109/3DV.2019.00085}{Publication link}.}
  \label{Fig:Rev:lung}
\end{figure}

The error images are used in Figure \ref{Fig:Rev:liver} to help the readers identify the differences between the ground truth, a scan generated by the proposed method, and two others generated by a 3D Fully Convolutional Neural network and a vanilla GAN. The whiter the error images are, the more similar the synthetic and the original images. In image translation with paired images, this comparison is possible because ground truth is available. However, some authors choose not to use these error images, as can be seen in Figures \ref{Fig:Rev:cardiac_pancreas_mammogram} and \ref{Fig:Rev:lung}. In some situations, it is not possible to create them, e.g., when ground truth is not available \citep{Zeng2019759}.

If the generation of synthetic data is only an intermediate step, this visual analysis is not very important if it is successful in its purpose.
\cite{Han2019729} improved the nodule detection algorithm by using synthetic data. Figure \ref{Fig:Rev:lung} shows the results of generating synthetic nodules and inserting them into surrounding tissue with and without using the L\textsubscript{1} loss. In this case, it is possible to see the differences between the two approaches.

\begin{figure}
  \centering
  \includegraphics[width=0.8\columnwidth]{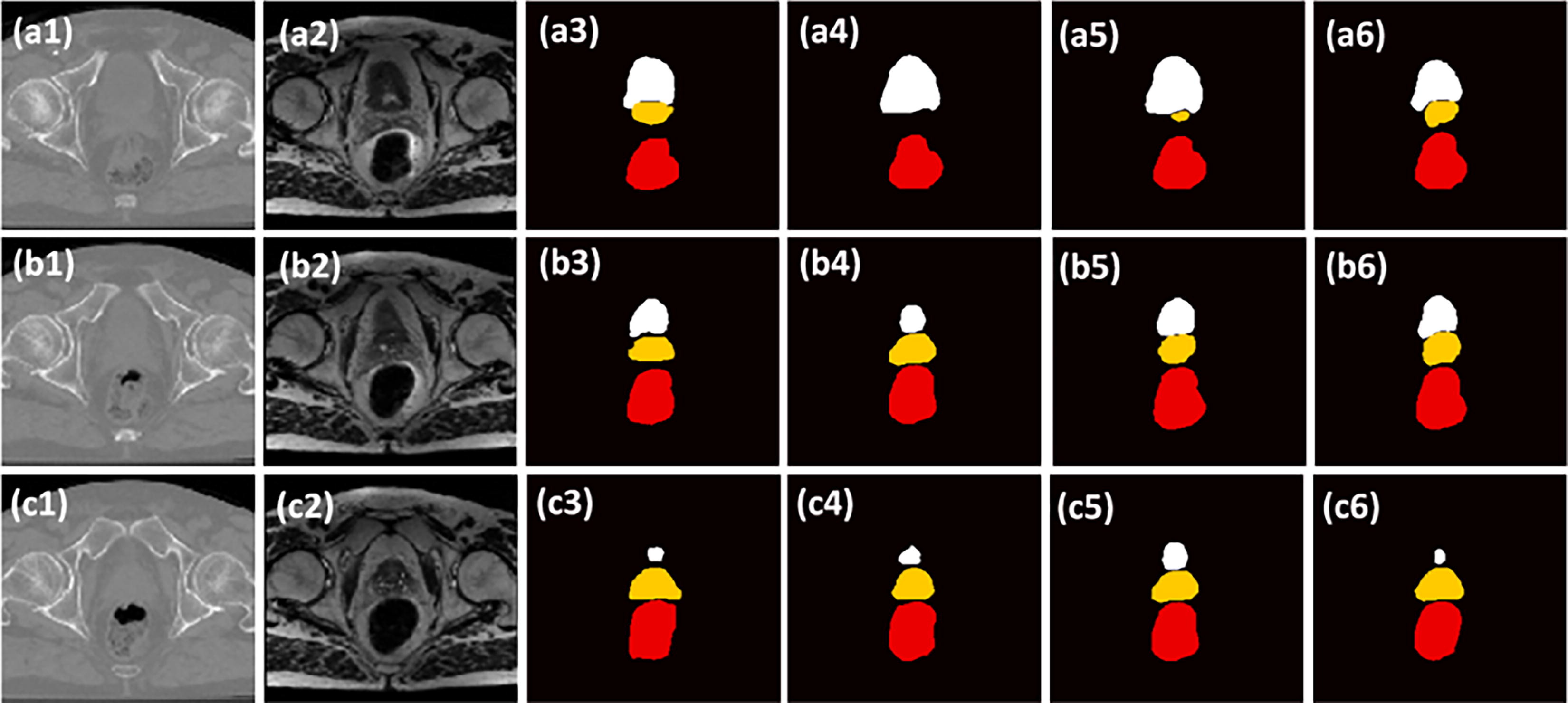}
  \caption{Comparison between original and synthetic scans generated by the method proposed by \citep{dong2019synthetic}, and segmentations: (1) CT  image; (2) synthetic MRI; (3) manual segmentation of bladder (white), prostate (yellow), and rectum (red); (4) segmentation by DSUnet; (5) segmentation by DAUnet trained on CT data; (6) segmentation by DAUnet trained on synthetic MRI data. (a), (b) and (c) are the results on three distinct patients. Reprinted from Radiotherapy and Oncology, Vol. 141, authors Xue Dong and Yang Lei and Sibo Tian and Tonghe Wang and Pretesh Patel and Walter J. Curran and Ashesh B. Jani and Tian Liu and Xiaofeng Yang, "Synthetic MRI-aided multi-organ segmentation on male pelvic CT using cycle consistent deep attention network," pages 192-199, Copyright (2019), with permission from Elsevier B.V.. \href{https://doi.org/10.1016/j.radonc.2019.09.028}{Publication link.}
  \label{Fig:Rev:prostate_bladder_rectum}}
\end{figure}

\cite{Lei2020a} trained a CycleGAN-based architecture to perform image translation from CT to MRI to achieve better soft tissue contrast. Again, the focus of this work is not on the realism of synthetic MRIs, but on improving segmentation when synthetic data are available. The same authors published a work similar \citep{dong2019synthetic}, where in Figure \ref{Fig:Rev:prostate_bladder_rectum} it can be seen that the segmentation performed with synthetic data outperforms the other two when compared to manual segmentation (a3).

\begin{figure}
  \centering
  \includegraphics[width=0.7\columnwidth]{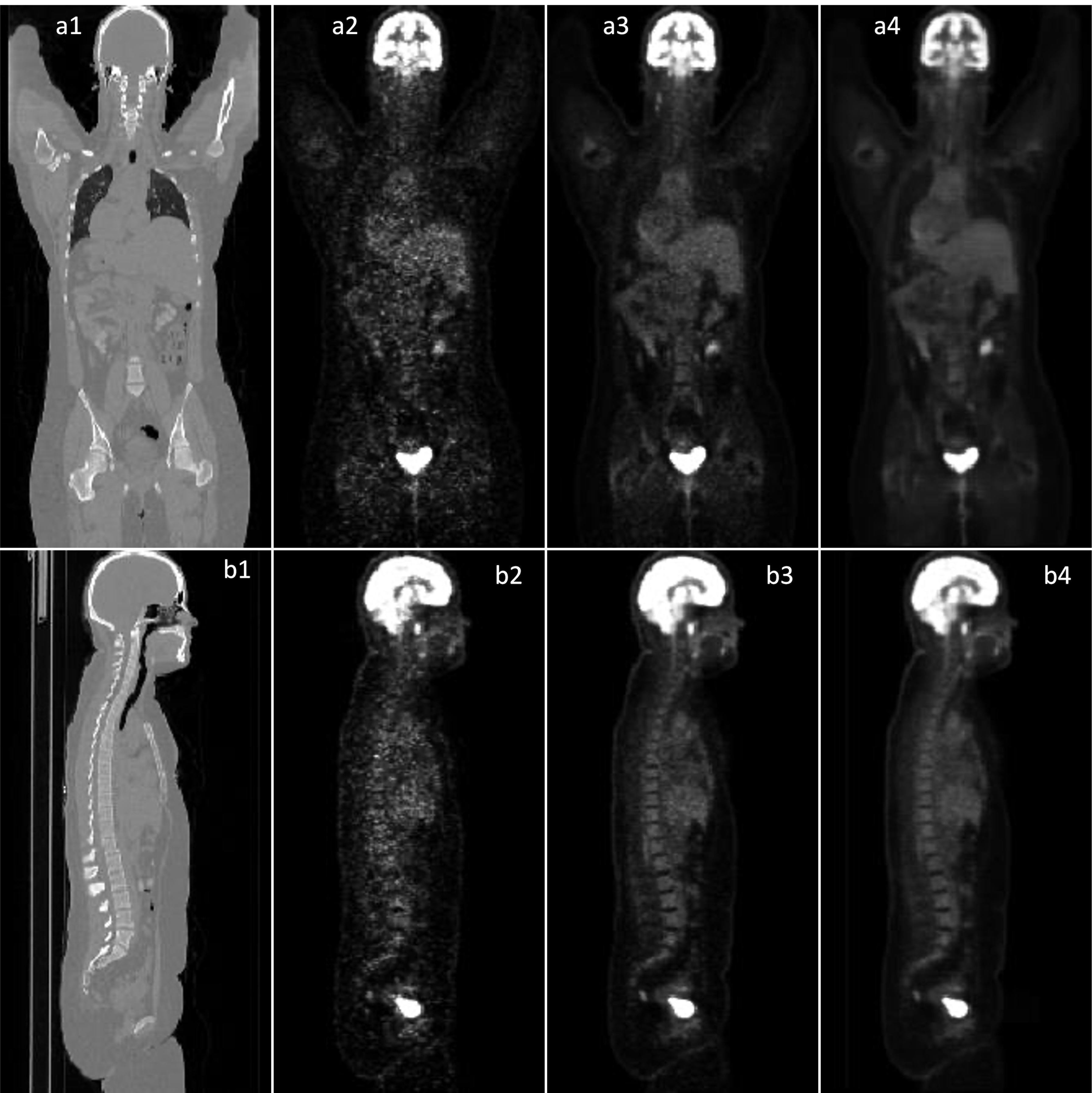}
  \caption{Visual results. (a1-a4) shows the sagittal views of CT, low count PET, full count PET and synthetic full count PET images. (b1-b4) shows the coronal views \cite{Lei2020b}. Reprinted from Medical Imaging 2020: Physics of Medical Imaging, Vol. 11312, authors Yang Lei and Tonghe Wang and Xue Dong and Kristin Higgins and Tian Liu and Walter J Curran and Hui Mao and Jonathon A Nye and Xiaofeng Yang, "Low dose PET imaging with CT-aided cycle-consistent adversarial networks", pages 1043-1049, Copyright (2020), with permission from Society of Photo-Optical Instrumentation Engineers (SPIE) and one of the authors. \href{https://doi.org/10.1117/12.2549386}{Publication link}.}
  \label{Fig:Rev:whole_body}
\end{figure}

The CycleGAN architecture is so versatile that it can also be used for image translation in small structures such as the heart, pancreas, and mammograms, as seen in Figure \ref{Fig:Rev:cardiac_pancreas_mammogram} by \cite{Cai2019174}. It can even be used for translating whole body images, as shown in Figure \ref{Fig:Rev:whole_body} developed by \cite{Lei2020b}.

\begin{figure}
\centering
\includegraphics[width=0.6\linewidth]{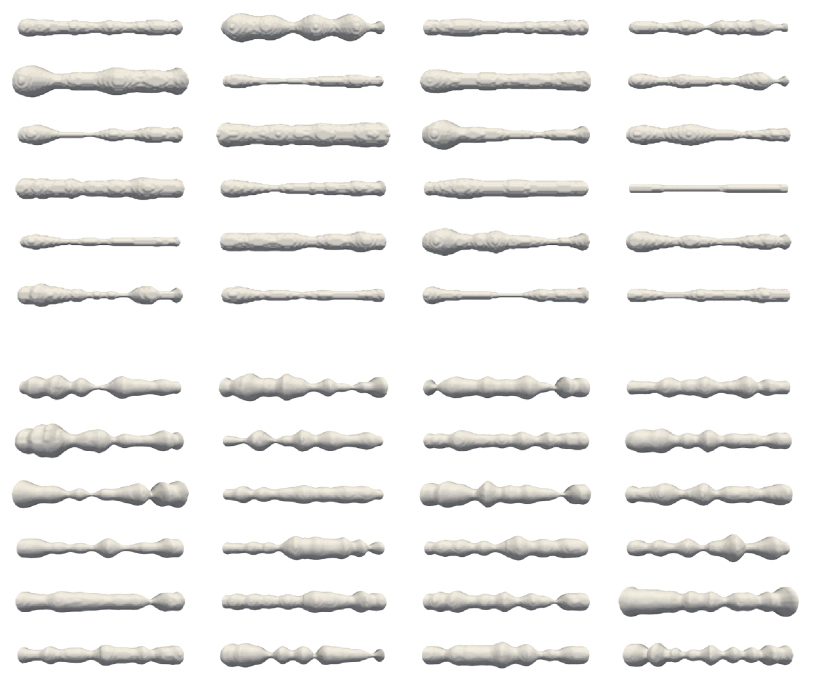}
  \caption{Stenosed segments of coronary arteries \cite{Danu2019662}: Top — training surfaces;  Bottom — synthetically generated using the GAN model. © 2019 IEEE. Reprinted, with permission, from 2019 23rd International Conference on System Theory, Control and Computing (ICSTCC), authors Manuela Danu and Cosmin-Ioan Nita and Anamaria Vizitiu and Constantin Suciu and Lucian Mihai Itu, "Deep learning based generation of synthetic blood vessel surfaces", pages 662-667. \href{https://doi.org/10.1109/ICSTCC.2019.8885576}{Publication link.}}
  \label{Fig:Rev:vessels}
\end{figure}

\begin{figure}
  \centering
  \includegraphics[width=0.6\linewidth]{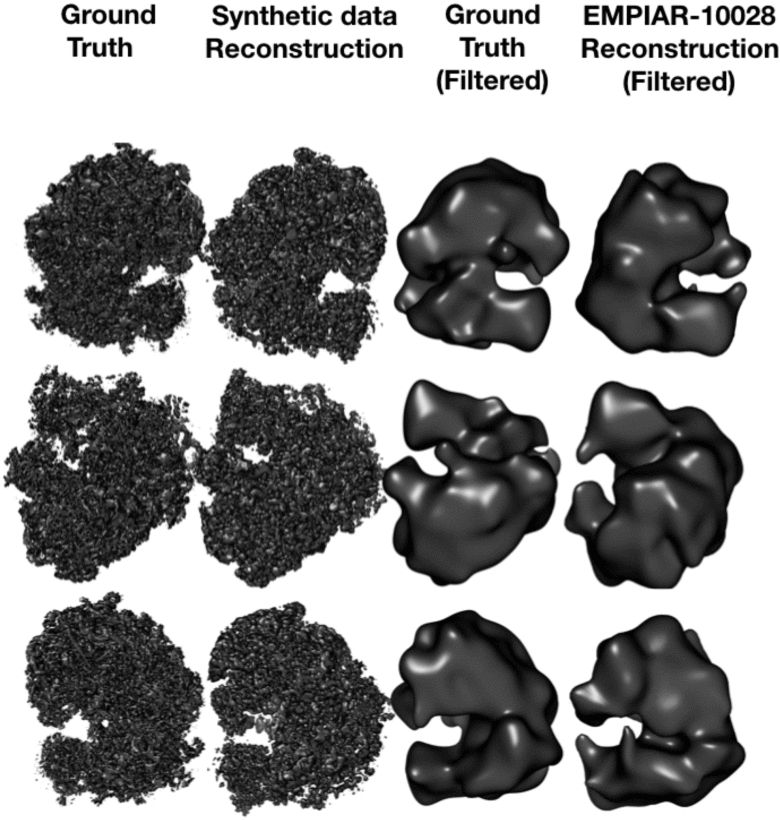}
  \caption{Different views of the 80S Ribosome and reconstruction using CryoGAN \cite{Gupta2021759}. First column — 80S Ribosome from EMPIAR-10028; Second column — CryoGAN reconstruction from synthetic data with; Third column — 80S Ribosome filtered; Fourth column — CryoGAN reconstruction filtered. Figure reproduced from  \cite{Gupta2021759}. \href{https://ieeexplore.ieee.org/document/9483649}{Publication link}. 
  \href{https://creativecommons.org/licenses/by/4.0/legalcode.en}{Licence CC BY 4.0 DEED}.
}
  \label{Fig:Rev:biomolecule}
\end{figure}

Still in the medical field, GANs have been used successfully by \cite{Danu2019662} to generate stenosed segments of coronary arteries and by \cite{Gupta2021759} to generate reconstructions of the 80S Ribosome, as seen in Figures \ref{Fig:Rev:vessels} and \ref{Fig:Rev:biomolecule}. This last work differs from the others because a CryoEM physics simulator was used as a generator instead of a deep learning architecture.

\begin{figure}
  \centering
  \includegraphics[width=1\columnwidth]{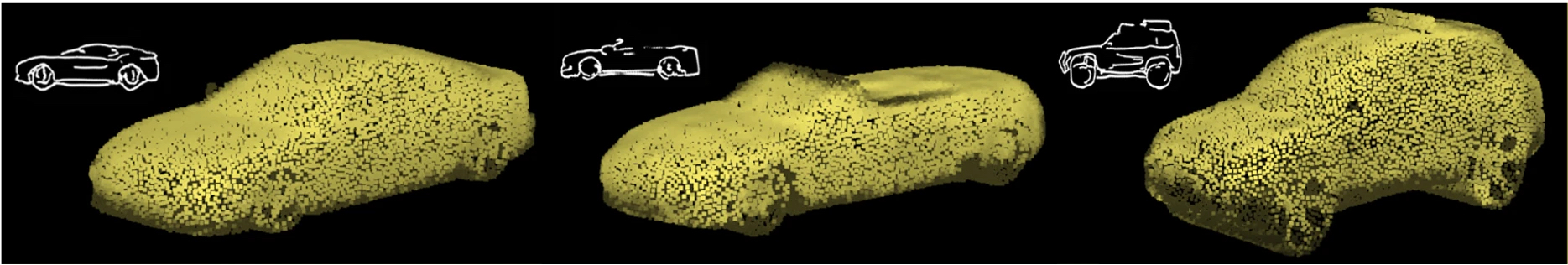}
  \caption{Examples of 3D car shapes generated by the system created by \cite{Nozawa2021} with side-view sketches. Top left white lines are input contour sketches; yellow point clouds are corresponding outputs. Figure reproduced from  \cite{Nozawa2021}. \href{https://link.springer.com/article/10.1007/s00371-020-02024-y}{Publication link}. \href{https://creativecommons.org/licenses/by/4.0/legalcode.en}{Licence CC BY 4.0 DEED}.}
  \label{Fig:Rev:cars}
\end{figure}

\begin{figure}
  \centering
  \includegraphics[width=1\columnwidth]{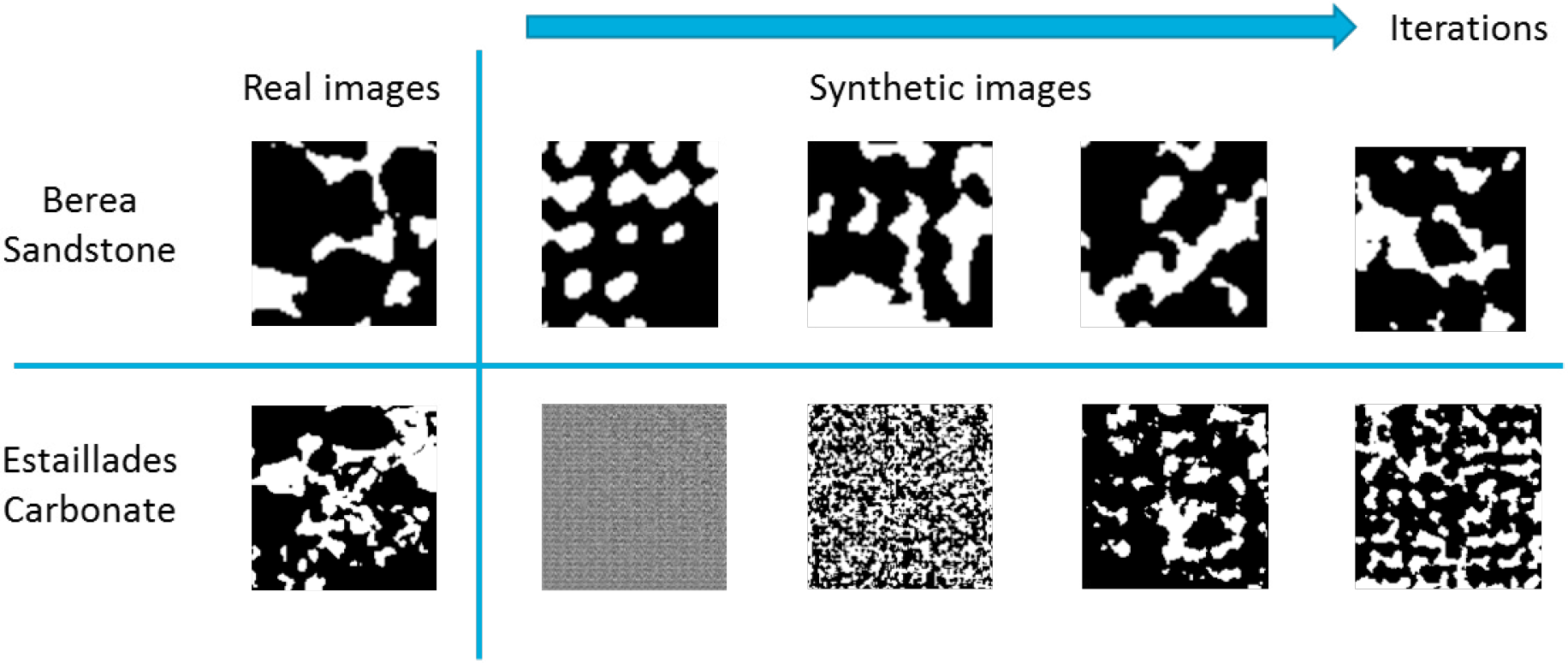}
  \caption{Real and generated images during training process for Berea sandstone and Estaillades carbonate sample \cite{Liu20196164}. Reprinted from Energy Procedia, Vol. 158, authors Siyan Liu and Zhi Zhong and Ali Takbiri-Borujeni and Mohammad Kazemi and Qinwen Fu and Yuhao Yang, "A case study on homogeneous and heterogeneous reservoir porous media reconstruction by using generative adversarial networks", pages 6164-6169, Copyright (2019), with permission from Elsevier B.V.. \href{https://doi.org/10.1016/j.egypro.2019.01.493}{Publication link}. Licence \href{https://creativecommons.org/licenses/by-nc-nd/4.0/}{CC BY-NC-ND 4.0 DEED}.}
  \label{Fig:Rev:porous_media}
\end{figure}

The use of GANs to generate 3D data is not limited to medical imaging, as explained earlier. They can also be used to generate synthetic porous media or 3D car shapes from sketches, as shown in Figures \ref{Fig:Rev:cars} and \ref{Fig:Rev:porous_media}, or even various everyday objects \citep{Muzahid202120}. These works demonstrate the wide versatility of GAN architectures.

It would be beneficial to the paper if in addition to the error images, the authors also represented the evolution of training, as made by \cite{Liu20196164} and shown in Figure \ref{Fig:Rev:porous_media}. It is known that the GANs training is very unstable and during training the output may deteriorate, but if no training breakdown occurs, such as mode collapse, the discriminator will eventually improve the output of the generator.

\onecolumn

\twocolumn

\section{Discussion and Conclusion}
\label{sec:Conclusion_and_Discussion}

Using GANs to generate volumetric data has become a highly researched topic, although it was not explored much in the first two years of the existence of GANs. This review has shown that GANs are capable of generating synthetic data that can be used for various medical and non-medical purposes. The generation of volumetric data is of great importance for the medical field, as volumetric medical imaging such as CT, MRI, and PET are increasingly being used, and the need for suitable pipelines to process these data is growing. 

GANs have already been used for tasks such as classification, segmentation or even reconstruction, both for medical and non-medical fields. This overview contains lists of abbreviations/acronyms, common loss functions, evaluation metrics, applications and architectures. It is important to have this overview because of the large number of publications, the variability between papers and terms, and their major impact on the generation of synthetic data with GANs. When available, implementation code is also provided so that new researchers can build their pipeline more quickly.

\subsection{Other papers}

The authors are aware of a few relevant papers that were not covered by the systematic review search. However, they are worth mentioning because of their novel results. 

\cite{ferreira2022IGI} and \cite{ferreira2022generation} developed a pipeline for generating synthetic rat brain MRIs by relying on the $\alpha$-GAN architecture \citep{rosca2017variational, kwon2019generation}, but with two different goals in mind. The first work addressed the need of large datasets to train deep learning models and a solution to this challenge, i.e. the use of GANs. They proved that it is possible to use synthetic data generated by GANs to improve existing segmentation algorithms. The second paper went even further and proved that synthetic data can outperform the use of conventional data augmentation and showed that in this particular case conventional data augmentation is actually suboptimal. As far as we know, these are the first papers in which synthetic MRI scans of rat brains have been generated. This work is the evidence that GANs can also be used in preclinical studies.

The $\alpha$-GAN architecture proposed by \cite{rosca2017variational} may solve both blur and mode collapse by using perceptual similarity metrics and adversarial loss. It has also been shown that the use of spectral normalisation \citep{Miyato2018} can stabilise the training of the discriminator, which helps to avoid mode collapse and non-convergence. 
WGAN-GP \citep{Gulrajani2017} can also be a good strategy to achieve Lipschitz continuity and, again, avoid mode collapse, non-convergence and vanishing.
All of these techniques were successfully applied in \cite{ferreira2022generation}.

\cite{sun2020mm} proposes an architecture based on GANs to solve the limited amount of labelled datasets. The MM-GAN can translate label maps into 3D MRI without distorting the pathology. They proved the extensibility of their approach by using the BraTS17 dataset and a new dataset (LIVER100).

\subsection{Dealing with limited data}
\label{sub:Dealing_with_limited_data}
As already mentioned, the generation of synthetic data using GANs is a good approach for data argumentation when the amount of data is not sufficient for the downstream task. However, if the amount of data is already too small, how can a GAN be trained? It is well known that GANs require large datasets to produce realistic samples, and large volumetric datasets are not freely available. Training a GAN depends on the type and complexity of the data to be generated, thus there is no general approach to train a GAN perfectly with any dataset. 

The discriminator is a deep learning classifier. Therefore, if there is not enough data, overfitting can easily occur when no precaution is taken. Even with large datasets, overfitting can occur if the discriminator network is large enough. Overfitting the discriminator is a problem that is usually not addressed, but can lead to poor results because an overfitted discriminator does not provide meaningful feedback to the generator. This can be detected by using a validation dataset consisting of real images that were not seen in the training process of the GAN. If the discriminator classifies most of these images as fakes, it means that the discriminator is overfitted.

To address this issue, \cite{karras2020training} have conducted a study on how the size of the dataset affects the quality of the generated data and propose the use of an adaptive augmentation mechanism that stabilises the training and allows the use of smaller datasets without overfitting the discriminator. Traditional data augmentations, i.e. geometric transformation (rotate, flip, scale, shift), colour transformation, noise addition, image filtering and cropping, have been shown to be beneficial for the generality of deep learning tasks. Therefore, these augmentations can be used for training the discriminator. However, such transformations should not be learned by the generator, i.e. the augmentations cannot leak because the augmentation version does not correspond to reality. The probability ($p$) of a transformation occurring should be lower than the probability of the transformation not occurring. This makes the training dependent on the original data without transformation, which is actually learned by the generator. It ensures that if $\tau_x=\tau_y$, then $x=y$, where $\tau$ is the stack of invertible transformations, $x$ and $y$ are the real and fake distributions. In order to make this work, all data (real and fake) are transformed before being fed into the discriminator, i.e., $D(\tau (G(z)))$ and $D(\tau (x))$, where $z$ is the random vector. $p$ is adjusted depending on the overlap between the real and fake distributions. If the percentage of correct predictions on the real data is too high, $p$ increases, and in the opposite case it decreases. Although this study was performed on 2D images, the same approach can be applied to volumetric data, as it does not depend on the loss function or architecture of the GAN.

Besides data augmentation, other strategies can be used to avoid overfitting the discriminator. Transfer learning has been shown to improve the results of deep learning tasks when a small dataset is available. However, large pre-trained models, e.g. VGG, ResNet50, InceptionV3, only work with 2D data and not volumetric data.  The lack of such a large model pre-trained on volumetric data makes this approach more difficult. However, even if no large pre-trained model is available, it is possible to address such a task with similar datasets \citep{chen2019med3d}. For example, if the goal is to generate synthetic MRI scans of the brain for Alzheimer's disease classification in a personal small dataset, the discriminator can first be pre-trained with the BraTS and/or ADNI datasets using a strategy called self-supervised learning \citep{rani2023self}. Subsequently, this model can be used as a discriminator by replacing the last layers with a sigmoid and freezing the first layers. The rest of the training process of the GAN is done as usual. However, if no data is available, a technique called weight inflation can be used to perform transfer learning from 2D models to 3D models \citep{liu2022inflating}. This technique is very recent and little explored for GANs, but it could be interesting for further research. Zero-shot learning \citep{xian2018zero} has also been explored to overcome the lack of data of particular classes for positron image denoising \citep{zhu2023generative}, image-to-image translation \citep{lin2021zstgan},  feature generation \citep{gao2020zero}, and others. No work has been found in which zero-shot learning has been applied to volumetric data, but such an approach could be adapted to volumetric data for further investigations.

However, what should be done when transfer learning is not possible? In cases where no other data can be found that is similar to the dataset to be used, efforts should focus on the discriminator's architecture. No deep learning architecture will overfit if its capacity is not large enough, so reducing the number of learning parameters of the discriminator can help. Adding dropout layers \citep{srivastava2014dropout} and using weight decay could also be helpful. Focusing on hyperparameter tuning is also a good strategy as, e.g., an incorrect learning rate can cause the network to collapse. 

Volumetric data is more complex than two-dimensional data because of the presence of volumetric information. However, some types of volumetric datasets might be easier to learn and would require fewer training samples than 2D data. For example, medical data of the same modality and body part is very identical, which is not the case for datasets of natural images, such as ImageNet. This low intravariability is advantageous for training GANs because the data distribution is more restricted, which makes it easier for the generator to produce realistic samples. Therefore, in some cases, a few hundred cases may be sufficient to train a GAN properly.

\subsection{Differences between volumetric and 2D data generation}
\label{sub:Differences_2D_3D}
Although a lot of research has been conducted in this area, many challenges remain, such as those mentioned in section \ref{sec:GANs}. Instability, mode collapse, non-convergence, and overfitting are even more difficult to overcome when 3D data are used instead of 2D data. Experiments with 3D data also take longer and there is no large dataset freely available, which explains the fact that the most important discoveries of GANs have been made for 2D data first.
In \cite{egger2022medical}, the main differences between using 2D and 3D are discussed, highlighting the complexity of dealing with an extra dimension, the difficulty of capturing large datasets among other issues.

Volumetric data generation is not just an extension of 2D approaches, as some architectures, loss functions and evaluation metrics are incapable of capturing the contextual and spatial information. 2D approaches are not able to use the context of adjacent layers since they cannot learn the spatial relationship between voxels, which is important for accurate and realistic generation. Such an approach can lead to shape discontinuities across adjacent layers and no consistency in voxel intensity among layers. Continuity between layers is very important in some downstream tasks, e.g. the winners of the 2020 \cite{isensee2021nnu} and 2021 \cite{futrega2022optimized} BraTS Challenges used a volumetric architecture for brain tumour segmentation. Therefore, synthetic volumetric data must contain this information, if not, it might be useless or even harmful for the downstream task.

\textbf{Network designs:} 3D convolution based networks are often used to generate volumetric data. These differ from 2D convolutions in that they are able to capture the depth information of volumetric data by learning dependencies between layers. Some of the architectures mentioned in section \ref{sub:architectures} are derived from 2D architectures, but with adapted hyperparameters and particular specifications to handle the additional difficulty of dealing with the complexity of volumetric data. Depending on the task, the network can process both 2D and 3D data (sections \ref{subsub:3Dfrom2D}), accept volumetric data as input (section \ref{sub:Vanilla_GAN_based} and \ref{subsub:conditional_GAN_based}) or even use special networks to control the main GAN pipeline, e.g. a DSNN (section \ref{subsub:Feature-consistent}). Therefore, these networks must be able to extract all features contained in volumetric data. Volumetric data also requires computers with higher processing and storage capacity than 2D data. Some works address this last challenge and attempt to solve it. \cite{Nozawa2021} use multiview depth images as an intermediate representation, and then create the volume by recomposing multiple views. \cite{Jung202079} replaced the 3D generator with a 2D generator, keeping a 3D and a 2D discriminator, which allowed faster training with lower memory consumption. \cite{Pesaranghader202167} has been able to create volumes with dimensions greater than $224$×$224$×$224$ by generating multiple slices and then assembling them, rather than processing the entire volume at once. \cite{Harms2020} uses patches as input and generates full resolution by stacking these patches. However, these approaches do not ensure that depth information and some features that might be important for some tasks, such as symmetry or long dependencies, are captured. \cite{Wang20172317} employ LRCN to increase the resolution of the generator's output, which makes GAN training faster and more stable by using lower resolutions, achieving higher resolutions even with memory constraints.

\textbf{Loss functions:} Some researchers have proposed their own loss functions to take advantage of volumetric information and produce more realistic data. For example, perceptual loss, which is used in several works on 2D images and does not take into account the dependencies between layers, is replaced by volumetrically trained networks, such as in the loss functions feature consistent, latent vector, content and style. The GANs’ objective functions  have already been discussed in \ref{subsub:sum_adv_loss_functions}, where it is concluded that there are no major differences between them, but that the vanilla GAN and the WGAN-GP are the most commonly used and that the WGAN-GP may perform better on more complex data. More detailed information on loss functions for volumetric data can be found in the following sections: latent vector loss (section \ref{subsub:latent}), feature consistent loss (section \ref{subsub:Feature-consistent}), content and style loss (section \ref{subsub:content_style}) Gradient difference loss (section \ref{subsub:GDL_GMD}), identity loss (section \ref{subsub:Identity}), Laplacian, projection and orientation losses (section \ref{subsub:laplacian_projection_orientation}), depth loss (section \ref{subsub:depth}), shape consistency and spatial (section \ref{subsub:shape_spatial}), border and volume losses (section \ref{subsub:border_volume}).

\textbf{Evaluation metrics:}
A consistent metric for comparing different models for volumetric data generation is also crucial. For example, if two papers use the same dataset to generate new data, it is very difficult to compare which model is better. For reconstruction or noise reduction problems, this problem is alleviated by the fact that ground truth is available, e.g. in \cite{Harms2020, Lei2020b, Liu2020, Li2020}, since the use of PSNR and MS-SSIM/SSIM is widely accepted. However, in modality translation it is very difficult to evaluate with unpaired data only, or when the goal is to generate completely new volumes \citep{Xu201962}. 
The generation of 3D data has not been sufficiently researched compared to the generation of 2D data. The most commonly used metric for comparing models is FID, which only works with 2D data. Several adaptations of FID to 3D data have been made, but none of them are optimal, as the volumetric context is lost (section \ref{subsub:FID_IS}). Metrics such as SIS and S-score should be considered instead of FID for volumetric data.
Even quality control through human judgement is a bottleneck, as the developer may not have the required expertise in the field in question, e.g. medical image analysis, relying on expert feedback, which is difficult to obtain, very expensive and time-consuming. Nevertheless, the visual Turing test is one of the best metrics for evaluating synthetic data as well as applying the data to a downstream task. It is curious, though, that so few papers use the visual Turing test. For a more detailed discussion of the best evaluation metric, refer to \ref{sec:Evaluation_Metrics_Generation_Task}. No work was found that uses maximum mean discrepancy (MMD). MMD is able to measure how similar two distributions are, which can be used to assess whether the real and fake data follow the same distribution, as used in \cite{ferreira2022generation}.

There is no specific strategy that works in all cases, not even for 2D images. For volumetric data, the difficulty is greater due to the additional dimension. However, following the recommendations given in section \ref{subsub:sum_adv_loss_functions} will help greatly to achieve satisfactory results. Continuous monitoring of the loss plots is helpful to detect instability and mode collapse; choosing a low learning rate makes the training more stable; using normalisation layers, e.g., spectral normalisation, batch normalisation or instance normalisation, also stabilises the training; test the WGAN-GP objective function if the vanilla GAN does not produce satisfactory results; ensure that the networks are correct and that the pipeline has no errors; find the GAN network that is best suited to the problem at hand, e.g. CycleGAN for image translation and cGAN for reconstructions; use other loss functions to support the objective function of the GAN as described in section \ref{sec:Loss_Functions}, e.g., identity, feature consistency, voxel-wise, among others; choose an appropriate evaluation metric as described in section \ref{sec:Evaluation_Metrics} and visually evaluate the quality of the generated volumes frequently. Stable training is very important to avoid mode collapse. It is important to note that the size of the input vector of the generator should be adapted to the complexity of the data, as volumetric data is to be used, the size should be increased. Problems with the GAN's training may also be related to the size of the dataset, see section \ref{sub:Dealing_with_limited_data}.

The use of more efficient coding methods is also essential to make the best use of the available hardware. MONAI \citep{monai_consortium_2022_7459814} has done a great effort in this respect, using best practices in volumetric data processing and offering a wide range of features to support the user.

The storage, access, and visualisation of volumetric data is also different and more complex than 2D imagery, which requires specialised pipelines. 
2D images are typically stored and loaded as JPG or PNG, whereas volumetric data encompasses a wider range of formats: DICOM, NIfTI (generality of medical data), CN3D \footnote{\href{https://pubmed.ncbi.nlm.nih.gov/25953817/}{EMPIAR-10061}}, OFF \footnote{\href{https://modelnet.cs.princeton.edu/}{ModelNet10/40}}, OBJ+MTL \footnote{\href{https://arxiv.org/pdf/1512.03012.pdf}{ShapeNet}}, among many others.

Summarising, the generation of 3D data compared to 2D data, for both medical and non-medical purposes, faces several challenges related to the limitations of the machines, i.e., need of more memory, more powerful GPUs and more energy consumption, as well as the lack of robust and consistent metrics to assess the quality of the new data, and the lack of large, freely available volumetric datasets.

\subsection{Concerns with synthetic data}
\label{Concerns_with_synthetic_data}

GANs can also be a threat, as they can be used to deceive people. Recently, there have been increasing reports of the use of synthetic videos and voices in video calls \footnote{\href{https://www.businessinsider.com/people-applying-remote-tech-jobs-using-deepfakes-fbi-2022-6}{Deepfakes new (accessed [30-06-2022])}}. These threats only use 2D data and video, and no problems with volumetric data have been reported yet. However, biases in medical data could be amplified, rare diseases are not generated if they are not present in the original dataset (or have not been appropriately processed), the correct phenotype could be missing,  artefacts/hallucinated details can be misleading or lead to (falsely) overconfident interpretation by human experts, among other problems.
In more extreme cases, it may be possible with this technology to fake examinations, e.g. remove brain tumours from images or create synthetic tumours in healthy brains so that patients have to pay more for medical treatment. 

Furthermore, there are no methods sufficiently robust and objective to determine whether a synthetic dataset differs from the genuine original dataset to be classified as truly anonymous, leaving a dangerous gap in sensitive data sharing legislation.
As mentioned in \cite{chen2021synthetic}, the trained models can have vulnerabilities that attackers can explore, such as information leakage and re-identification.

The generation of volumetric synthetic data comes with several other challenges, such as reliability, the need for experts to generate the data and assess its realism, and the possibility of real-world outliers being excluded. It is clear that the lack of proper metrics to ensure quality, realism and anonymization are among the biggest problems in using synthetic data for medical proposes. 

\subsection{Research Questions}
\begin{enumerate}
    \item What are the different applications of GANs in the generation of volumetric data?

GANs are used in various fields, whereby a distinction can be made between medical and non-medical fields, since most works are in the medical field. They can be used for classification, reconstruction, denoising, nuclei counting, segmentation, image translation and general purposes, as can be seen in Tables \ref{tab:table5} and \ref{tab:table6}. 

    \item What are the methods most frequently or successfully employed by GANs in the generation of volumetric data?

Since it is not possible to make a good comparison between the models, it is not easy to assert which GANs are more successful in generating volumetric data. However, CycleGAN-based architectures are chosen very often and have achieved good results, followed by architectures based on cGAN, DCGAN, and WGAN / WGAN-GP, as can be seen in Figure \ref{Fig:ArchToApp} and Tables \ref{tab:table2}, \ref{tab:table3} and \ref{tab:table4}. The best architecture also depends heavily on the task, e.g. CycleGAN should be used for image translation but probably not for reconstruction.

    \item What are the strengths and limitations of these methods?

The use of GANs has proven to be a powerful tool for handling multimodal data, for data augmentation, and for anonymisation. However, there are several problems that have already been mentioned in the Sections \ref{sub:Dealing_with_limited_data} and \ref{sub:Differences_2D_3D}. The lack of a consensus metric to evaluate GANs, the instability of training, hardware limitation, and lack of large datasets remain major challenges, especially when dealing with volumetric data.

    \item What improvements are sought through the use of this technology?

This technology is widely used to generate anonymised synthetic data for data augmentation, as shown in this review, due to the inherent anonymisation capacity of GANs and because of its realism compared to other generative approaches. The synthetic data can be used to improve other deep learning algorithms by solving the problem of lack of data and being a good aid for other data augmentation techniques.
In addition, GANs can also be used for denoising, reconstruction and modality translation problems, making these architectures a powerful tool to have always ready when other technologies are not satisfactory.

\end{enumerate}

\subsection{Tendencies of 3D volumetric data generating with GANs}
It is noted that most of the works use improved adaptations of existing GAN architectures for the problem at hand. Architectures such as CycleGAN, cGAN, WGAN/WGAN-GP, PGGAN, and DCGAN were first developed for 2D data generation and then adapted to volumetric data. As mentioned earlier, the most investigated GAN architectures, loss functions and evaluation metrics are mainly 2D related. These works are so remarkable that researchers usually decide to adapt them for volumetric generation instead of trying to achieve such results from scratch. 

One of the architectures that has attracted the most interest from researchers is the CycleGAN. The use of CycleGAN-based architectures is a preferred choice for many applications, namely image translation, nuclei counting, reconstruction, and segmentation,  as it is not limited to the use of only one modality. Its ability to handle multimodality is a valuable asset in many applications, especially for medical data. It is well known that the use of more than one medical modality for diagnosis is essential, as it offers more information. It is highly desirable to have pipelines that are able to mimic physicians who use more than one modality when treating patients \citep{heiliger2022beyond}. As shown in \cite{huang2020fusion}, using multimodal data can lead to higher accuracy and better imitate human experts. Therefore, it is only natural that versatile mechanisms capable of handling them are the next steps in which the CycleGAN-based architecture can have a crucial role.

In works such as \cite{Zhang20189242, Zhang2019183}, this technology is used to generate MRI scans from CT and vice versa to increase the amount of data available for multiclass segmentation of the heart. Generating one modality from another is also applied to the liver, brain or even the whole body, as can be seen in table \ref{tab:table5} (section \ref{sec:Applications}), showing the versatility of these architectures. In works such as \cite{Schaefferkoetter20213817, Zeng2019759, Zhang2019183, Zhang20189242}, the use of GANs has proven to be beneficial for exploiting multimodal medical datasets, even when only unpaired data exists. This architecture, together with the loss function cycle consistency, feature consistency, identity, shape consistency and spatial consistency, allows researchers to obtain high quality results. These consistency losses (except for cycle consistency) are a good example of an improvement over an existing GAN architecture, making it more suitable for volumetric data. It is expected that the use of such specific loss functions will increase in the future, as each problem is different, so there should be cost functions adapted to each dataset. 

For non-medical applications, cGAN and DCGAN have been the preferred choice. From this review, it can be seen that volumetric data generation has been applied more to medical data than to non-medical data, limiting the development of the latter. In the majority of papers, authors preferred to keep the use of GAN networks at a simpler level without using more complex networks. While this is not a shortcoming, it would be interesting to use more complex networks to provide deeper conclusions. The use of the Minkowski function is another example of what is expected in the future, i.e. the use of specific loss functions that are able to capture the essential features of the dataset.

cGAN is also much explored, e.g. in the work developed by \cite{Han2019729}. They created a pipeline to generate lung nodules in the desired position, size and attenuation by creating small volumes ($32$×$32$×$32$) with nodules and adding them to the surrounding tissue, creating $64$×$64$×$64$ volumes. This technique can also be extended to various other types of lesions in other organs by converting the scans from healthy patients to diseased ones. It can also be applied beyond medicine, for example to create a specific porous media between other materials or to create synthetic fractures in materials. This allows saving computational resources and a more stable training due to the small output size of the generator. cGAN is also a good option for future work, as generating data conditioned by labelling reduces the time and cost of obtaining and labelling large datasets. It is therefore expected that more work will be done in the future with conditions for generating synthetic volumetric data.

It is also noted that latent space has not been explored as much as it deserves. As explained in \cite{fragemann2022review}, the "black box" inherent in generative models should be eliminated by better understanding the models and exploring techniques such as disentanglement. Therefore, this deep understanding and exploration can also be the next approach to the use of GANs, rather than creating new architectures. Explainable AI applied to volumetric data generation will also be of great interest, as understanding how the data was generated can help improve existing mechanisms.

It is noted that researchers are aware of the limitations of 2D data generation approaches for volumetric data as they strive to use architectures, loss functions and evaluation metrics that can consider the additional complexity of such data. No major development is expected in terms of architectures for generating voxel grid data, but most likely the creation/use of loss functions and evaluation metrics that are able to account for all dimensions and the dependencies that voxels have with each other.

As explained in section \ref{sub:Types_of_3D_data_representation}, there are other ways to explore volumetric data besides using voxel grids. In the works found in this systematic review, the authors avoided using point clouds or meshes by converting them to voxel grids, as this type of data is better studied and therefore less difficult to process. The most studied approaches for generating synthetic volumetric data are specific to voxel grids, so the data must first be converted to use such approaches. However, converting to a voxel grid has some disadvantages, such as losing information and adding more constraints to the data. Hence, it is expected that architectures such as PointNet and PC-GAN will be explored more in the future. Research into architectures that can directly process meshes and point clouds would be of great interest. 

\subsection{Conclusion}
In this systematic review, we give an overview of GAN-based approaches for the generation of realistic 3D volumetric data. In doing so, we group the screened works by the underlying data (dimension), which can be 2D or 3D, and remove the 2D data works. Then, we further group them by the targeted modalities, CT, MRI, PET, or combinations of modalities. We furthermore extracted the used datasets and if they are public ones that are available to the research community, or if these are so-called private or in-house repositories that are not (yet) available to the community. Note that even if making datasets available alongside publications is becoming more and more common these days, there may be reasons for keeping data under lock. Making data, especially medical data, public requires much more efforts and extended or additional ethics approvals. It can also be that the institution that generated the data, e.g. a hospital, allows studying the data but forbids its publication, or that the disease is so rare in the institution that it does not make sense to publish it because only a few cases are involved. Another reason can be, that the authors want first to exploit the data for further studies and publications, e.g. for rare disease cases, which can be understandable, because creating data collections requires a certain amount of effort, especially large ones, including further manual processing such as creating ground truth segmentations. Getting an internal review board to share the data with internal and external partners involved in the study also takes time and a lot of work, with regard to preparation and anonymisation.

In addition to the datasets and modalities, we extracted the data types, the network architectures, the loss functions, evaluation metrics and resolutions for every reviewed work. It was concluded that the main modality used was CT, followed by MRI, and that the use of multimodal data is widely used especially in the medical context. \textit{Image translation} was the main use of GANs, followed by \textit{Reconstruction}, which explains the extensive use of CycleGAN-based and cGAN-based architectures. It can be seen that, especially in the non-medical context, the use of synthetically generated datasets to train GANs was a trend in the last year (2021). In the medical context, the use of multimodal data has been explored every year, proving the importance and need for appropriate pipelines to process this type of data.

\subsubsection{Research Opportunities}
Comparable to the study of \cite{lucic2018gans}, it would be interesting to investigate different GAN architectures (including vanilla GAN, PGGAN, CycleGAN) and objective functions (including vanilla GAN, LSGAN, HingeGAN, WGAN and WGAN-GP) with different datasets of volumetric data to investigate whether one is better than the others. Such a study would help to determine with greater clarity whether a particular architecture and/or objective function is preferable over the others, or whether they do not show any difference in results. 

Due to the inherent anonymisation capability of GANs, they could be used to generate synthetic head scans of faces, as the defacement of head scans is not always an option, e.g. for medical augmented reality \citep{gsaxner2019facial} and facial implants \citep{memon2021review}. In these works, it is not possible to develop algorithms with satisfactory results because it is very difficult to obtain large datasets with the facial part, especially with the soft tissue. All data must be approved so that it can be published and freely shared. However, such data is very difficult to be approved because it is not anonymous as it contains the face part. Therefore, creating synthetic faces for these scans would be a good solution to this challenge. 

The creation and deployment of a full data anonymisation framework could be a good solution to data sharing problems, as proposed by \cite{shin2018medical}. It would also be very interesting to develop a tool that is able to automate the whole process of training and generating synthetic data, similar to the nnUNet \citep{isensee2021nnu}, which is able to configure itself and choose the best network architecture, training, pre- and post-processing for each new task. Such an automatic tool would be very difficult to design, as each dataset is very different from the others, but for generating data from the same source, e.g. medical volumetric data, the development of such a tool would be possible and very useful.

The use of multimodal data as input is a rarely addressed topic. An example of the benefits of using multimodal data is \cite{Lei2020b}, which generates a full PET count from low-count PET and CT scans. There are several works that include multimodal data in the training dataset, but they are used to train an image translator and do not exploit the full potential that this dataset can offer, i.e. they only use one modality at a time. For example, the use of PET+CT scans to produce MRI could be relevant.

In order to improve GANs, a uniform evaluation metric should be established to allow better comparison between models from different studies. It would also be very important to find a metric that is close to human judgment to improve the results of GANs, especially when volumetric data is used. FID adaptations to volumetric problems are not generally accepted because it is computationally intensive and lacks spatial context. Research in this area is more important than the development of new GAN architectures, as several different architectures are already available that show great potential, but lack appropriate metrics and loss functions. The next developments of GANs can also proceed through the exploration of latent space and use more techniques such as disentanglement. In a medical context, the best approaches will probably be the use of multimodal data, which makes CycleGAN-based architectures highly preferable.

The creation of large databases of volumetric data that are freely available and easy to use would also be of great interest for pretraining models (as explained in section \ref{sub:Dealing_with_limited_data}). The existence of a database where data can be retrieved via a query would make it easier for the researcher to download the data needed instead of searching for it in different places or wasting time requesting it.

The use of attention mechanisms to generate synthetic data will certainly increase, due to the fast development of architectures based on Transformers \citep{vaswani2017attention}, such Vision Transformer (ViT) \citep{dosovitskiy2020image}, Swin UNETR \citep{hatamizadeh2022swin}. Also, the fast development of hardware that can process large amounts of data will encourage research into volumetric data generation and the employment of more complex/computationally demanding approaches.

The generation of other volumetric data representations, such as point clouds and meshes, has not been explored much. It would be of great benefit to investigate ways to generate such data instead of having to convert them into voxel grids first.

It is undeniable that volumetric data generation is having a greater impact on the medical field, so research in this area is expected to increase. There has also been an increase in the number of FDA-approved AI to support healthcare decision-making, which could lead to greater investment in this area so that more data is available to create more reliable models.

Volumetric 3D data must be explored as a separate problem from 2D data. The distance from 3D to 2D can be understood as the distance from 2D to 1D data. There are special networks for text processing data (1D data) and images (2D data), as both have a very different complexity and distribution. Volumetric data has an extra dimension compared to 2D data, which gives it complexity and additional features that are not present in 2D images.

\section*{Acknowledgments}
André Ferreira was supported by a scholarship from Fundação para a Ciência e Tecnologia (FCT), Portugal (Scholarship number 2022.11928.BD), Ministry of Science, Technology and Higher Education (MCTES), Fundo Social Europeu (FSE) and European Union. This work was supported by FCT within the R\&D Units Project Scope:
UIDB/00319/2020.
It was also supported by by the Advanced Research
Opportunities Program (AROP) of RWTH Aachen University.
This work received funding from enFaced (FWF KLI 678), enFaced 2.0 (FWF KLI 1044) and KITE (Plattform für KI-Translation Essen) from the REACT-EU initiative (EFRE-0801977, \url{https://kite.ikim.nrw/}).

\setcounter{section}{0}
\renewcommand\thesection{\Alph{section}}

\setcounter{equation}{0}
\renewcommand{\theequation}{\Alph{section}.\arabic{equation}}

\section{Appendix - Loss functions}
\label{app:loss_functions}
\subsection{Adversarial loss}
\label{app:sub_Adversarial_loss}
From GANs' first paper \citep{Goodfellow2014}, the adversarial loss is given by the discriminator's correct classification of the realism of the data received. If the discriminator gives a correct prediction, the generator is penalised; if it gives an incorrect one, the discriminator itself is penalised. This is also known as the "minmax-game", originally expressed in equation \ref{eq:GAN}, where the output of the discriminator is compared with the real label (true or fake) using BCE. Several alternatives to compute this "game" have been purposed.  

\subsubsection{\textbf{Jensen-Shannon (JS) and Kullback–Leibler divergence (KL-divergence)}}
\label{subsub:JSD_KLD}
The JS loss is based on the KL divergence to measure the distance between two probability distributions. The adversarial loss (minmax-game, described by the equation \ref{eq:GAN}) is also defined in the original paper \citep{Goodfellow2014} by the JS divergence between the real data distribution and the distribution of the generator's parameters, where $min_Gmax_D$ is given by the equation \ref{eq:JS}:

\begin{equation} \label{eq:JS}
\begin{split}
JS(P_X||P_\theta)=KL(P_X||\frac{P_X+P_\theta}{2}+KL(P_\theta||\frac{P_X+P_\theta}{2})
\end{split}
\end{equation}

where $KL$ is the Kullback-Leiber divergence (equation \ref{eq:KL}), $P_X$ the real data distribution, and $P_\theta$ the distribution of the weights of the generator. \citep{Goodfellow2014} shows that the optimal solution, i.e., $P_X=P_\theta$, is obtained when the generator consistently outputs $\frac{1}{2}$, i.e., $D_G^*(x)=\frac{1}{2}$ (with a fixed G).

\begin{equation} \label{eq:KL}
\begin{split}
KL(P_X||P_\theta) = \int p_X(x)log(\frac{p_X(x)}{p_\theta(x)})dx
\end{split}
\end{equation}

where $p_X$ and $p_\theta$ are the probability densities of $P_X$ and $P_\theta$.

\cite{Pesaranghader202167} is one of the few papers that mentions the use of JS divergence for calculating adversarial loss, but it is assumed that the papers that do not mention any other method for calculating adversarial loss (e.g. Wasserstein or least squares) used the original approach, i.e. JS divergence computed with the BCE (section \ref{subsub:CE}). Thus, if "JS" appears in the column "Loss function" of the tables below (\ref{tab:table2}, \ref{tab:table3} and \ref{tab:table4}), it means that the use of this function is mentioned in the paper, and just "Adv" otherwise.
\cite{Pesaranghader202167} did not find any drastic changes in their experiments by using Wasserstein loss or JS. However, \cite{arjovsky2017wasserstein} has proved that using Wasserstein loss (section \ref{subsub:WGAN}) is preferable over JS, since KL yields infinity when two distributions are disjoint. Even when two distributions have no intersection (which is common for low-dimensional manifolds), Wasserstein loss can still provide a reasonable representation of the distance between the two distributions, without sudden changes, which is very important for a stable learning process. 

The KL-divergence (also known as relative entropy) is used in \cite{Hu2022145} to reduce the distribution divergence between the encoded latent vector (encoded by a ResNet encoder) and the sampled latent vector (from the normal distribution) in the purposed GAN (BMGAN).

\subsubsection{\textbf{Least Squares (LSGAN)}}
\label{subsub:LSGAN}
The LSGAN loss \citep{Mao2017} (equations \ref{eq:LSGAN_D} and \ref{eq:LSGAN_G}) uses the least squares loss function instead of the BCE (section \ref{subsub:CE}) loss function to avoid vanishing gradients. \citep{Mao2017} argue that BCE leads to the problem of vanishing gradients when classifying fake samples that are on the correct side of the boundary but still far from reality. LSGAN even penalises the correctly classified by forcing the generator to produce samples in the direction of the decision boundary (i.e. towards the manifold of the real data). It also provides a more stable training process by mitigating the problem of vanishing gradients, i.e. the BCE saturates more easily, resulting in non-meaningful feedback. 

\begin{equation} \label{eq:LSGAN_D}
\begin{split}
min_{D}V_{LSGAN}(D)=\frac{1}{2}\mathbb{E}_{x \sim p_{data}(x)}[(D(x)-b)^2] \\+\frac{1}{2}\mathbb{E}_{z \sim p_{z}(z)}[(D(G(z))-a)^2]
\end{split}
\end{equation}
\begin{equation} \label{eq:LSGAN_G}
\begin{split}
min_{G}V_{LSGAN}(G)=\frac{1}{2}\mathbb{E}_{z \sim p_{z}(z)}[(D(G(z))-c)^2]
\end{split}
\end{equation}
where $a$, $b$ and $c$ are the labels for fake, real and the value the generator ($G$) wants the discriminator ($D$) to believe for the generated data, respectively. The values of $a$, $b$ and $c$ should satisfy the conditions $b-c=1$ and $b-a=2$ or b=c=1 and a=0, as both performed similarly in \citep{Mao2017}.
It is used in for the generation of volumetric data in \cite{Slossberg2019498, Han2019729, Xu201962, Cai2019174, Hu2022145, Yang2021130}.

\subsubsection{\textbf{Hinge}}
\label{subsub:Hinge}
The hinge loss (equation \ref{eq:hinge}), as explained in \cite{rosasco2004loss}, is a loss function widely used for classification. In this case, it was used in \cite{greminger2020generative} to measure the distance between the discriminator output and the label. It was shown in \cite{lim2017geometric} that the use of hinge losses leads to a Nash equilibrium between the generator and the discriminator. \cite{Miyato2018} claims that using this loss instead of the vanilla GAN loss leads to better results in their spectral normalisation GAN. \cite{jolicoeur2018relativistic} compares the vanilla GAN, LSGAN, HingeGAN (GAN with hinge loss) and WGAN-GP, but no conclusions can be drawn from their experiments, as WGAN-GP performed better with the CIFAR-10 dataset \citep{krizhevsky2009learning}, but HingeGAN performed better with the CAT dataset \citep{zhang2008cat}. However, it is interesting to note that vanilla GAN and LSGAN had problems with convergence on higher resolution images, while WGAN-GP was able to converge. Unfortunately, no experiments were conducted with HingeGAN for higher resolutions. 
Therefore, this loss can be used instead of BCE (section \ref{subsub:CE}) or least squares for GAN loss (section \ref{subsub:LSGAN}). However, no study has been conducted with 3D GANs comparing these metrics, so it is recommended to try this loss and evaluate if more training stability and more realistic generations are archived.

\begin{equation} \label{eq:hinge}
\begin{split}
L_{hinge}=\sum_{i=1}^{n}max(0,1-\hat{y_l}\cdot y_i)
\end{split}
\end{equation}
where $\hat{y}$ is the output of the discriminator, $y$ the label, and $n$ the number of samples. The real and the fake labels are 1 and -1, respectively.

\subsubsection{\textbf{Wasserstein GAN (WGAN) and  WGAN with Gradient Penalty (WGAN-GP)}}
\label{subsub:WGAN}
The WGAN \citep{Arjovsky2017} uses the Wasserstein distance (i.e., earth-mover (EM) distance) instead of the JSD (section \ref{subsub:JSD_KLD}) to measure the divergence between two distribution. The EM distance is defined by equation \ref{eq:EM}:

\begin{equation} \label{eq:EM}
\begin{split}
W(\mathbb{P}_r, \mathbb{P}_g)=inf_{\gamma \in \Pi(\mathbb{P}_r, \mathbb{P}_g)} \mathbb{E}_{(x,\hat{x})\sim \gamma}[||x-\hat{x}||]
\end{split}
\end{equation}
where $\Pi(\mathbb{P}_r, \mathbb{P}_g)$ is all possible joint probability distributions between $\mathbb{P}_r$ and $\mathbb{P}_g$,  $\gamma(x,\hat{x})$ indicates how much "mass" to transport from $x$ to $\hat{x}$ in order to transform $\mathbb{P}_r$ into $\mathbb{P}_g$.

The $inf$ is an intractable problem, however, based on the Kantorovich-Rubinstein duality \citep{cedric2008optimal}, equation  \ref{eq:EM} can be transformed into the equation \ref{eq:EM_sup}:

\begin{equation} \label{eq:EM_sup}
\begin{split}
W(\mathbb{P}_r, \mathbb{P}_\theta)=sup_{||f||_L\leq 1} \mathbb{E}_{x\sim \mathbb{P}_r} [f(x)]-\mathbb{E}_{x\sim \mathbb{P}_\theta} [f(x)]
\end{split}
\end{equation}

where $sup$ is the 1-Lipschitz functions $f : \chi \rightarrow \mathbb{P}$. Supposing that $f$ comes from a family of K-Lipschitz continuous functions $\{f_w\}_{w\in W}$, the optimization function can be defined by the equation \ref{eq:EM_final}:

\begin{equation} \label{eq:EM_final}
\begin{split}
W(\mathbb{P}_r, \mathbb{P}_\theta)=max_{w\in W} \mathbb{E}_{x\sim \mathbb{P}_r}[f_w(x)] - \mathbb{E}_{z \sim p(z)}[f_w(g_\theta (z))]
\end{split}
\end{equation}
where $w$ are the parameters to be learned by the discriminator. The discriminator is called a critic because he evaluates the authenticity of the image instead of saying whether the image is real or fake. However, WGAN suffers from unstable training, slow convergence and vanishing gradients due to the weight clipping used to preserve the Lipschitz continuity, which even the authors describe as "a clearly terrible way to enforce a Lipschitz constraint".

The WGAN-GP \citep{Gulrajani2017} is an improvement over the original WGAN by replacing the weight clipping with gradient penalty (defined in equation \ref{eq:GP}). 

\begin{equation} \label{eq:GP}
\begin{split}
L_{GP}(\mathbb{P}_{\tilde{x}}) = \mathbb{E}_{\tilde{x}\sim \mathbb{P}_{\tilde{x}}}[(||\nabla_{\tilde{x}}D(\tilde{x})||_2-1)^2]
\end{split}
\end{equation}
where $\mathbb{P}_{\tilde{x}}$ is sampled uniformly along straight lines between the data distribution ($\mathbb{P}_{r}$) and generator distribution ($\mathbb{P}_{\theta}$). Since $f$ is 1-Lipschitz if it has gradients with norm at most 1 everywhere, the  model is penalized by the GP when the gradient norm moves away from 1. The final critic (i.e., "discriminator") loss is defined by equation \ref{eq:WGAN-GP}.

\begin{equation} \label{eq:WGAN-GP}
\begin{split}
L_{WGAN-GP}(\mathbb{P}_r, \mathbb{P}_\theta, \mathbb{P}_{\tilde{x}}) = -W(\mathbb{P}_r, \mathbb{P}_\theta) + \lambda L_{GP}(\mathbb{P}_{\tilde{x}})
\end{split}
\end{equation}
with $\lambda$ is the weight of the $GP$ (usually 10), and $-W(\mathbb{P}_r, \mathbb{P}_\theta)$ since in training the loss is minimized.

Using WGAN-GP makes it possible to achieve Lipschitz continuity with almost no hyperparameter tuning. Batch normalisation cannot be used because the objective of WGAN-GP is not valid in such an environment due to the creation of correlations between samples. It should therefore be replaced by layer normalisation \citep{ba2016layer}. 
The WGAN loss is used in \cite{Chen2020, Li2020, Kench2021299}, and the WGAN-GP loss is used in \cite{Slossberg2019498, Pesaranghader202167, Han2019729, Jung202079, Zhuang2019, Baumgartner20188309, Yang2017679, Gupta2021759, Shen20213250} to generate synthetic volumetric data.

\subsubsection{\textbf{Cycle consistency}}
\label{subsub:Cycle_consistenc}
Cycle consistency loss can be defined by the equation \ref{eq:cycle_GAN}:

\begin{equation} \label{eq:cycle_GAN}
\begin{split}
L_{cycle}(G,F)=\mathbb{E}_{x\sim p_{data}(x)}[||F(G(x))-x||_1]\\ + \mathbb{E}_{y\sim p_{data}(y)}[||G(F(y))-y||_1]
\end{split}
\end{equation}
where $||\cdot||_1$ is the MAE (section \ref{subsub:MAE_MSE}), $G$ the forward generator, $F$ the backward generator, $x$ the input image of $G$ and $y$ the input image of $F$. 

This loss ensures that the image generated by the $G$, when used as input of $F$ is similar to the input of $G$, i.e,  $F(G(x))\approx x$, and vice-versa, i.e.,  $G(F(y))\approx y$.
It is mainly used for CycleGAN \citep{CycleGAN2017} training and is widely used for image translation when paired data is not available, as it does not require a comparison between the generated image and the ground truth in the intermediate step, i.e., $||G(x)-y||_1$ and $||F(y)-x||_1$. 

However, the cycle consistency loss does not ensure good quality of the images generated in the intermediate step. This is done by using two discriminators (one for each intermediate step), which can lead to unrealistic synthetic data. When ground truth is available, i.e. the real images of the intermediate step, the use of shape consistency, spatial consistency (section \ref{subsub:shape_spatial}), identity consistency (section \ref{subsub:Identity}) or feature consistency (section \ref{subsub:Feature-consistent}) is recommended to control the middle state of CycleGAN, which is not controlled by the cycle consistency.  Without such control, G(x) and F(y) might have artefacts that only serve to trick the discriminators, not to look realistic.

The comparison between the two images is made per-pixel using MAE as in the original paper, or other metrics such as MSE or CE (section \ref{subsub:CE}). \citep{CycleGAN2017} tested the use of adversarial loss instead of MAE, but no improvements were observed, so MAE should be used as it provides more stability in training, ensures that the generated images are close to input and is less computationally intensive.
It is used in \cite{Jung202079, Harms2020, Lei2020a, Schaefferkoetter20213817, Lin2021, Chen2021961, Han2019, Yang2021130} for the generation of volumetric data.

\subsection{Loss functions to explore the intermediate layers}
\label{app:sub_Loss_functions_to_explore_the_intermediate_layers}

\subsubsection{\textbf{Latent vector loss}}
\label{subsub:latent}
Latent vector loss as well as depth loss (section \ref{subsub:depth}) are used by \cite{Liu20212843} to increase the fidelity of 3D reconstruction from monocular depth images of objects.

The latent vector loss (equation \ref{eq:locgen}) minimises the distance between the latent vector of a trained autoencoder $z_{dae}$ (trained to learn the representation of the ground truth 3D voxel grid) and the latent vector of the proposed generator $z_{edgan}$ (which is also an autoencoder). This loss can be understood as a way to force the generator's encoder to produce latent vectors (local-global latent vectors) identical to those of the autoencoder, which has already learned to correctly encode the ground truth (global-global latent vectors). This loss can be used not only for 3D reconstruction, but also for denoising, segmentation or even image translation as long as the ground truth is available and a bottleneck architecture, i.e., with a latent vector (e.g. autoencoder, variational autoencoder) is used.

\begin{equation} \label{eq:locgen}
\begin{split}
L_{locgen} = |\mathbb{E}(z_{dae})-\mathbb{E}(z_{edgan})|
\end{split}
\end{equation}

\subsubsection{\textbf{Feature-consistent}}
\label{subsub:Feature-consistent}
Feature-consistent loss (equation \ref{eq:featcons}) is used in \cite{Pan2019137} to ensure that the synthetic and real scans of the same modality have the same disease pattern. For this purpose, a disease-specific neural network (DSNN) is trained for disease classification. The features extracted by the model from a real and a generated scan (of the same modality) should be identical. This loss induces the CycleGAN to focus more on the disease-specific features, which is particularly important for disease-specific reconstruction or image translation. Such a loss would not be appropriate for healthy brains, in this case the DSNN should be replaced by another trained model with healthy brains.

This loss is similar to the latent vector loss (section \ref{subsub:latent}), but here the loss is measured with the output features of several layers, not just the latent vector. This loss function is also very familiar to the loss of cycle consistency, since the goal is to minimise the distance between the ground truth and the cycle reconstruction, i.e. $F(G(x))\approx x$, where $x$ is the input, $G$ is the generator that transforms $x$ to the other modality, and $F$ is the generator that transforms $G(x)$ back to the original modality. 

This loss is more specialised than identity loss (section \ref{subsub:Identity}) because it compares specific disease features in the intermediate state instead of just comparing voxel values. It is also more suitable for situations where the ground truth of the translated scan is not available. Both are important to complement cycle consistency loss by controlling the intermediate generation.

\begin{equation} \label{eq:featcons}
\begin{split}
L_{featcons} = ||F_M(G_P(X_P))-F_M(X_M)|| + \\ ||F_P(G_M(X_M))-F_P(X_P)||
\end{split}
\end{equation}
where $F_M$ and $F_P$ are the features extracted from MRI and PET scans by the DSNN, $G_P$ and $G_M$ the generators that transform PET into MRI and vice versa, and $X_M$ and $X_P$ the original MRI and PET scans.

\subsubsection{\textbf{Content, Style}}
\label{subsub:content_style}
Content and style as well as Laplacian, projection, and orientation (section \ref{subsub:laplacian_projection_orientation}) loss functions are used in \cite{Shen20213250} and are essential for their style transfer work from 2D to 3D.  The content loss (equation \ref{eq:content}) is the squared error between two feature representations, i.e. the squared error between the discriminator features in layer $l$ of the real and the fake samples. Higher-level layer (closer to the highest-resolution image) capture high-level information, e.g., entire objects, and lower-level layers represent voxel values, so for the content loss only high-level layers are used. 

\begin{equation} \label{eq:content}
\begin{split}
L_{content}(l)=\frac{1}{2}\sum_{i,j}[F_{l,i,j}(\hat{X}, I_S, I_M)-F_{l,i,j}(X, I_S, I_M)]^2
\end{split}
\end{equation}  
where $F_{l,i,j}$ represents the discriminator features of the $i^{th}$ filter at position $j$ in the layer $l$, $X$ and $\hat{X}$ the real and generated samples, $I_S$ and $I_M$ the sketch image and a mask image, and $\hat{X}=G_s(I_S,I_M)$.

The style loss (equation \ref{eq:style}) is the squared error between the Gram matrices in layer $l$ of the real and the fake samples. This serves to capture the texture and/or colour information, working on lower-level features, i.e., voxel level features.

\begin{equation} \label{eq:style}
\begin{split}
L_{style}(l)=\\ \frac{1}{4N^2_lM^2_l}\sum_{i,j}[A_{l,i,j}(\hat{X}, I_S, I_M)-A_{l,i,j}(X, I_S, I_M)]^2
\end{split}
\end{equation}  
where $A_l$ is the Gram matrices (defined in Appendix \ref{subsub:Perceptual} by equation \ref{eq:Gram_matrix}), $A_{l,i,j}$ is the inner
product between the vectorized feature maps $i$ and $j$ in the
$l^{th}$ layer ($A_{l,i,j}=\sum_{k}F_{l,i,k}F_{l,j,k}$), $N_l$ the number of feature channels, and $M_l$ the size of feature tensors.

Both are based on \cite{gatys2016image} which explains them in more detail. Both are comparisons that focus on different levels of the convolutional network and control the network at multiple levels, not just in the output. These are highly recommended for style transfer tasks that have a relevant impact on the generation of new styles and new features of computed generated volumes (e.g., for video games \citep{shi2020neutral} or for animated films).

\subsubsection{\textbf{Perceptual}}
\label{subsub:Perceptual}
Perceptual loss uses a pre-trained deep convolutional neural network to measure the perceptual differences between two images. In the first publication \citep{Johnson2016}, a VGG-16 (Visual Geometry Group with 16 layers) network was used \citep{simonyan2014very}, pre-trained on the ImageNet dataset \citep{russakovsky2015imagenet}. However, this concept can be extended to other pre-trained networks, such as VGG-19. This metric was used in one of the state-of-the-art publications on GANs \citep{Karras2019}. The perceptual loss is defined by two losses, feature reconstruction loss and style reconstruction loss. 

The feature reconstruction loss (equation \ref{eq:feature}) measures the MSE between the features extracted from the different layers (of the VGG network) of the generated image $\hat{x}$ and the target image $x$. The features extracted from higher layers contain information about the image content and spatial structure, but colour, shape and texture are lost. Therefore, the higher layers are used to calculate this metric by comparing the image $\hat{x}$ with the target content image ($x_c$).

\begin{equation} \label{eq:feature}
\begin{split}
l_{feat}^{\phi ,j}(\hat{x},x)=\frac{1}{C_jH_jW_j}||\phi _j(\hat{x}) - \phi _j(x))||_2^2
\end{split}
\end{equation}
where $\phi _j(\cdot)$ are the activations (i.e., feature map) of the layer $j$ of the VGG network ($\phi$), $C_j$×$H_j$×$W_j$ the shape of the $\phi _j(\cdot)$. 

The style reconstruction loss penalises the style differences (e.g., colour, patters, and texture) between $\hat{x}$ and the style target ($x_s$). Defining the \textit{Gram matrix} by the equation \ref{eq:Gram_matrix}, the style loss is the squared Frobenius norm difference between $\hat{x}$ and $x_s$ in several layers of the VGG network (equation \ref{eq:style_Vgg}). In \citep{Johnson2016}, features are extracted from higher and lower layers, extracting structural and style-related features, respectively.

\begin{equation} \label{eq:Gram_matrix}
\begin{split}
G_{j}^{\phi}(x)=\frac{\psi \psi ^T}{C_jH_jW_j}
\end{split}
\end{equation}
where $\psi$ is the $\phi _j(x)$ reshaped to $C_j$×$H_jW_j$

\begin{equation} \label{eq:style_Vgg}
\begin{split}
l_{style}^{\phi ,j}(\hat{x},x)=\||G_j^\phi (\hat{x}) - G^\phi _j(x))||_F^2
\end{split}
\end{equation}
where, $|| \cdot ||_F^2$ is the squared Frobenius norm, i.e., the MSE between two matrices. 

It was also used in for volumetric data by \cite{Hu2022145} (VGG-16). It is assumed that the pre-trained network has already learned how to encode the perceptual and semantic information, allowing it to give a meaningful feedback to the generator. However, this loss is rarely used with volumetric data, especially medical data, because the pre-trained networks do not accept volumetric inputs, i.e. the volumes would have to be sliced to fit the pre-trained networks, resulting in a loss of contextual and spatial information, and because of the large perceptual gap between the datasets used for pre-training (natural images) and the target datasets. Therefore, this metric is not recommended for volumetric architectures.
However, some other approaches are similar to this, such using a pre-trained segmentation model to assess if the features extracted from the original data are the same as the synthetic data, e.g., feature consistent (section \ref{subsub:Feature-consistent}), and latent vector loss (section \ref{subsub:latent}).

\subsection{Pixel/Voxel-wise loss functions}
\label{app:sub_Pixel/Voxel-wise_loss_functions}

\subsubsection{\textbf{Mean Absolute Error (MAE or L\textsubscript{1}), Mean Squared Error (MSE or L\textsubscript{2})}}
\label{subsub:MAE_MSE}
 MAE \citep{MAEref} (equation \ref{eq:MAE}) and MSE \citep{MSEref} (equation \ref{eq:MSE}) are often used in regression problems to compare the distance between two images/volumes. MAE is the absolute average difference between the predicted ($\hat{x}$) and the expected ($x$) values, and MSE is the average of the squared difference thereof. Both are often used to stabilise GAN training. MAE is used more frequently than MSE in the papers of this review (Figure \ref{Fig:loss_function_3D}), suggesting that the use of MAE may lead to more realistic volumetric data. \cite{isola2017image} and \cite{Zhang20189242} claim that MAE encourages less blurring and better visual results, but \cite{pathak2016context} did not find significant difference between both. Therefore, both metrics should be tested to find out which metric gives the best results, or use the $L_pnorm$ (section \ref{subsub:L_p}).
 Normally, MSE is used to compare pixel/voxel values, but \cite{Liu2020O47} calculates the MSE between feature vectors extracted from the discriminator. MAE and MSE are two very versatile metrics that can be used in different settings, e.g. cycle consistency (section \ref{subsub:Cycle_consistenc}), comparison in the frequency domain (section \ref{subsub:frequency_domain}), calculation of identity loss (section \ref{subsub:Identity}), among others.
 MSE and MAE loss functions can be used for both generation and downstream tasks.

\begin{equation} \label{eq:MAE}
\begin{split}
MAE=\frac{1}{n}\sum_{i=1}^{n}|x_i-\hat{x}_i|
\end{split}
\end{equation}
\begin{equation} \label{eq:MSE}
\begin{split}
MSE=\frac{1}{n}\sum_{i=1}^{n}(x_i-\hat{x}_i)^{2}
\end{split}
\end{equation}

\subsubsection{\textbf{L\textsubscript{p}-norm}}
\label{subsub:L_p}
The L\textsubscript{p} (equation \ref{eq:L_p-norm}) proposed in \cite{rahimi2013norm} is used to overcome the limitations of MSE (L\textsubscript{2}, blurry images) and MAE (L\textsubscript{1}, misclassification), as claimed by \cite{Liu2020, Harms2020}, who set $p=1.5$.

\begin{equation} \label{eq:L_p-norm}
\begin{split}
L_pnorm=\frac{1}{n}(\sum_{i=1}^{n}|x_i-\hat{x}_i|^p)^{1/p}
\end{split}
\end{equation}

where $n$ is the number of samples, $x$ and $\hat{x}$ the real and generated data.
This loss function is recommended when the use of $L_2$ produces too smooth boundaries and $l_1$ leads to misclassifications, as empirically found by \cite{Harms2020}.

\subsubsection{\textbf{Cross Entropy (CE)}}
\label{subsub:CE}
CE, also referred to as reconstruction loss defined in equation \ref{eq:CE}, is used to measure the difference between two distributions. This loss was used in \cite{Zhang2021} for the downstream tasks (segmentation), in \cite{Jung202079, Liu2020O47, Muzahid202120} to train the discriminator, which is also the classifier, i.e. the discriminator has two output channels instead of only one related to the realism of the input, and in \cite{Wang20172317} for the object reconstruction loss, i.e., to compare the generated reconstruction with the ground truth (voxel by voxel).

Binary Cross Entropy (BCE) \citep{e20030208, ruby2020binary}, also called sigmoid CE, is usually used to train a binary classifier. Often used as the criterion of comparison between the discriminator's classification and the real label, as mentioned in \citep{Zhang2021, Gayon-Lombardo2020, Dikici2021, Lei2020b}. 
When authors do not specify the use of other criteria and only mention the vanilla loss function (equation \ref{eq:GAN}), this usually means that BCE was used for the comparison between the discriminator output and reality. Higher values of BCE correspond to higher uncertainty of the discriminator, and therefore higher difference between distributions. Using BCE for this purpose makes the discriminator feedback more understandable to the user, more meaningful to the generator, as opposed to using the absolute value (directly using the output of the discriminator, as in WGAN).
It is also used in \cite{Nozawa2021} to measure the reconstruction loss between the generated masks and the ground truths.

A modified CE is used in \cite{Yang2017679} to strongly penalise false positives compared to false negatives due to the properties of the data used, and in \cite{Zhang2021} a weighted multi-class CE is used to ensure that all classes are learned equally. Such changes make the model less influenced by the classes that are easier to learn and less biased.

\begin{equation} \label{eq:CE}
\begin{split}
L_{CE} = - \sum_{i=1}^{n}t_{i}log(p_i)
\end{split}
\end{equation}  
where $n$ is the number of classes, $t_i$ the true class label, and $log(p_i)$ the predicted probability of class $i$.

\subsubsection{\textbf{Gradient difference, Gradient Magnitude Distance (GMD)}}
\label{subsub:GDL_GMD}
Gradient difference loss (GDL) were found in the literature, defined in two distinct ways. GDL is described as another voxel-wise metric for measuring the distance between two volumes, enhancing edge sharpness. A Sobel operator (edge detecting) \citep{kanopoulos1988design} is used by \cite{Ma2020} to measure the gradient difference between the generated image and the target image (equation \ref{eq:gd}).

\begin{equation} \label{eq:gd}
\begin{split}
L_{gd}(G)=\mathbb{E}_{i,j,k \sim  P_{data}(i,j,k)}[||gd(x)-gd(G(x))||_{MAE}]
\end{split}
\end{equation}
where $gd(\cdot)$ is the function that calculates the gradient using a Sobel operator, and $i$, $j$, $k$ the three axis.

The GDL can also be calculated using the equation \ref{eq:GMD}, which is referred to as the GMD loss in \cite{Harms2020} but was first introduced as GDL in \cite{Mathieu2016}. Instead of comparing the whole volume, it is sliced in all three axis and the squared errors between two real and generated sequential slices are computed. This ensures continuity between the two layers situated next to each other. \cite{Harms2020} do not discuss the importance of the loss, but it is assumed that this loss can lead to sharper and continuous consecutive slices instead of calculating the difference of only two slices, as explained by \cite{Mathieu2016}. It is also assumed that using one GDL or the other would give similar results, but no experiments comparing these two losses have been found. Therefore, GDL is recommended when edge sharpness is critical to the realism of the generated volume and the blur margins created by using MSE or MAE need to be reduced. It is expected that the use of such a metric would be beneficial for all image generation tasks, especially super-resolution, reconstruction, image translation and noise reduction.

\begin{equation} \label{eq:GMD}
\begin{split}
GMD(X,\hat{X}) = \sum_{i,j,k} \{(|X_{i,j,k}-X_{i-1,j,k}|-|\hat{X}_{i,j,k}-\hat{X}_{i-1,j,k}|)^2 \\+(|X_{i,j,k}-X_{i,j-1,k}|-|\hat{X}_{i,j,k}-\hat{X}_{i,j-1,k}|)^2 \\+ (|X_{i,j,k}-X_{i,j,k-1}|-|\hat{X}_{i,j,k}-\hat{X}_{i,j,k-1}|)^2\}
\end{split}
\end{equation}

where X and $\hat{X}$ are two volumes to be compared, and $i$, $j$ and $k$ the three orientations.

Gradient difference is also mentioned in \cite{Liu2020} but no further discussion and explanation about the metric is given.

\subsubsection{\textbf{Identity}}
\label{subsub:Identity}
The identity loss used in \cite{Schaefferkoetter20213817}, as explained in \cite{CycleGAN2017}, uses the MAE distance to measure the distance between the input and the corresponding output of the generator and works as a cycle consistency loss (section \ref{subsub:Cycle_consistenc}) helper in the CycleGAN architecture to avoid unnecessary changes in the intermediate step. The use of this loss is only possible when the images of the target modality are available. This loss is recommended to avoid unnecessary changes in the intermediated state of the CycleGAN training. Without such loss, G and F would be free to generate unrealistic intermediate scans, which is not controlled by the cycle consistency loss. Therefore, its use is recommended when CycleGAN-based architectures are used and when the intermediate target is available. In cases where the intermediate target is not available, feature-consistent loss (section \ref{subsub:Feature-consistent}) is recommended.

\begin{equation} \label{eq:identity}
\begin{split}
L_{identity}(G,F)=\mathbb{E}_{y\sim p_{data}(y)}[||G(y)-y||_{MAE}] \\+ \mathbb{E}_{x\sim p_{data}(x)}[||F(x)-x||_{MAE}]
\end{split}
\end{equation}
where $G$ is the generator that transforms one modality into the other (e.g. MRI to PET), $F$ is the inverse of the generator $G$, i.e. it transforms back into the original modality (e.g. PET to MRI), $x$ and $y$ are the original MRI and PET scans.

\cite{Schaefferkoetter20213817} trains a 3D CycleGAN to generate CT scans from MRI scans. They also compare the use of a 2D network, a 3D network and the 3D network with cycle consistency and identity losses as generator, concluding that the use of a 3D network leads to substantial improved quality in the inference, and that cycle consistency and identity losses were able to further improve the results (specially noticeable in the hands' reconstruction).

\subsubsection{\textbf{Laplacian, Projection, Orientation}}
\label{subsub:laplacian_projection_orientation}
Laplacian, projection, and orientation as well as content and style (section \ref{subsub:content_style}) loss functions are used in \cite{Shen20213250} and, as mentioned above, are essential for their style transfer work from 2D to 3D.

The projection loss (equation \ref{eq:proj}) is the squared error of the 2D projection of the generated volume with  the 2D image used as input. This is particularly useful to ensure fidelity between the input 2D image and the generated volume only in the visible part. 

\begin{equation} \label{eq:proj}
\begin{split}
L_{proj}=\sum_{i\in I_\Theta}(Proj(y)^i -\tilde{X_i})^2
\end{split}
\end{equation}  
where $I_\Theta$ represents the 2D region, $Proj(y)$ the 2D projection of the 3D orientation field $y$ in $I_\Theta$, and $\tilde{X_i}$ the fake 2D orientation map.

However, this does not take into account the entire 3D orientation. For this reason, the Laplacian loss (equation \ref{eq:lap}) helps to diffuse the constraints of the entire 3D orientation and not just the visible area. This loss is the representation of the gradient divergence difference between the ground truth and the generated 3D orientation. 

\begin{equation} \label{eq:lap}
\begin{split}
L_{lap}=\sum_{i}(\Delta(\tilde{y_i})-\Delta(y_i))^2
\end{split}
\end{equation}  

where $\tilde{y}$ is the generated 3D orientation, and $y$ the ground truth.

Orientation loss (reference \ref{eq:ori}) is used to create volumes, taking into account the style of other volumes. This is very similar to projection loss, but instead of comparing pixels, it compares voxels, which is specific to 3D data.

\begin{equation} \label{eq:ori}
\begin{split}
L_{ori}=\sum_{i\in \Gamma} (\tilde{y_i}^*-R(\tilde{y})_i)^2
\end{split}
\end{equation}
where $\Gamma$ are the visible 3D voxels, $R(\tilde{y})$ the rotated 3D orientation field, and $\tilde{y^*}$ the generated volume.

\subsubsection{\textbf{Depth}}
\label{subsub:depth}
The depth loss (equation \ref{eq:depth}) penalises the difference between the generated 3D voxel and the input 2.5D voxel grid (used by \cite{Liu20212843}). This ensures that the predicted values corresponding to the 2.5D voxel must be 1 and the values before that must be 0. The hidden voxels (not visible in the 2.5D grid) are weighted less because multiple solutions are possible. This ensures the fidelity of the visible parts of the input in the reconstructed volume. This loss is particularly relevant in 3D reconstruction problems from 2D or 2.5D images.

\begin{equation} \label{eq:depth}
\begin{split}
L_{depth}=\sum_{i=1}^{I}\sum_{j=1}^{J}((1-y_{edgan}^{ijk})^2+\sum_{k=1}^{k-1}(y_{edgan}^{ijk})^2), \\
k=argmin \{1\leq k \leq K|X_{in}^{ijk}=1\}
\end{split}
\end{equation}

where $y_{edgan}$ is the predicted 3D voxel grid by the proposed 3D GAN (EDGAN), ($I$, $J$, $K$) the size of the voxel grid, and $X_{in}$ the input 2.5D voxel grid.

\subsubsection{\textbf{Frequency domain}}
\label{subsub:frequency_domain}
Frequency domain loss (equation \ref{eq:freq}) was used by \cite{Ma2020} to reduce blur. It uses the Fast Fourier Transform to transfer the target image and the generated image into the frequency domain and calculate the MAE in the frequency domain. This loss can be used in various problems to increase the visual realism of the generated images if ground truth is available. It can be added or even replace other voxel-wise functions, giving similar results and stabilising the training of the generator. 

\begin{equation} \label{eq:freq}
\begin{split}
L_{freq}(G)=\mathbb{E}_{i,j,k \sim  P_{data}(i,j,k)}[||fft(y)-fft(G(x))||_{MAE}]
\end{split}
\end{equation}
where $fft$ is the Fast Fourier Transform, $x$ and $G(x)$ are the target and generated images. Note that this equation has been adapted to 3D from the original version \citep{Ma2020}.

\subsubsection{\textbf{Shape consistency and spatial}}
\label{subsub:shape_spatial}
Shape consistency loss (equation \ref{eq:shape}) is used by \cite{Zhang2019183, Zhang20189242, Cai2019174} to ensure anatomical structure invariance of a CycleGAN through the use of segmentation networks, i.e. if a scan of a certain modality (e.g., MRI) has a certain semantic label, the generated scan of the other modality (e.g., CT) must have the same semantic label. The CE (section \ref{subsub:CE}) is used to compare both segmentations.

\begin{equation} \label{eq:shape}
\begin{split}
L_{shape}(S_A,S_B,G_A,G_B)=\\ \mathbb{E}_{x_B \sim p_d(x_B)}[-\frac{1}{N}\sum_{i=1}^Ny^i_A log(S_A(G_A(x_B))_i)] \\ + \mathbb{E}_{x_A \sim p_d(x_A)}[-\frac{1}{N}\sum_{i=1}^Ny^i_B log(S_B(G_B(x_A))_i)]
\end{split}
\end{equation}
where $x_A$ and $x_B$ are the input scans of the modality $A$ and $B$, $y_A$ and $y_B$ are the respective ground truth labels, $S_A$ and $S_B$ the semantic segmentation networks, $G_A$ and $G_B$ the generators, and $A$ and $B$ the two modalities (CT and MRI).   

The spatial loss \citep{Fu2018} (equation \ref{eq:spatial}) is also related to the cycle consistency loss used in \cite{Chen2021961, Han2019, Ho2019} and serves the same purpose, but the segmentation network is trained to generate binary labels instead of semantic labels. MSE is used instead of CE to compare the two segmentations. In their experiments, a spatially constrained CycleGAN (SpCycleGAN) is used where the input is the label ($y$) and the output is the synthetic volumetric nuclei ($x$) and vice versa in the other direction of the CycleGAN. In this case, only $y$ is compared with the output of the segmentation network and not both, as in the case of shape consistency.

\begin{equation} \label{eq:spatial}
\begin{split}
L_{spatial}(S,G)= \mathbb{E}_{x \sim p_d(x)}[-\frac{1}{N}\sum_{i=1}^N||x^i-(S(G(y))_i)||_{MSE}] 
\end{split}
\end{equation}

Shape consistency and Spatial losses are similar to identity (section \ref{subsub:Identity}) and feature-consistent (\ref{subsub:Feature-consistent}) losses, as they are used to regulate the intermediate step of CycleGAN training. Therefore, one of these approaches should be considered to avoid unrealistic generations in the intermediate state of CycleGAN.

\subsection{Other loss functions}
\label{app:sub_Other_loss_functions}

\subsubsection{\textbf{Border and Volume}}
\label{subsub:border_volume}
Implemented in \cite{Momeni2021}, border and volume losses are used to improve the quality of the MRI scans generated for their GAN conditioned by volume and location. Volume losses (equation \ref{eq:volume_loss}) penalise the differences between the input volume (a numerical value) and the volume of the cerebral microbleed generated. Border loss (equation \ref{eq:border_loss}) forces the edges of the generated scan to be 1, making the generated microbleed blend with the surrounding tissue.
These losses are particularly important for this cGAN, which involves creating lesions with specific volumes, and it is crucial that they blend correctly into the surroundings. However, the border loss does not take into account that the voxels next to the border may have random values, so that discontinuity (no correct blending) may be possible. 
Such losses are not necessary if the generated lesions are not intended to fuse with the surrounding tissue or if the volume of the lesion is not important, i.e. the generator is not conditioned by the volume of the lesion. 

\begin{equation} \label{eq:volume_loss}
\begin{split}
Loss_{volume}=\frac{1}{N}\sum_{i=1}^{N}\left |\frac{V_{fake}^{i}-V_{in}^{i}}{V_{in}^{i}}  \right |
\end{split}
\end{equation}
where $N$ is the number of samples in the dataset, $V_{fake}^{i}$ and $V_{in}^{i}$ are the fake and real volumes.

\begin{equation} \label{eq:border_loss}
\begin{split}
Loss_{border}=\frac{1}{K}\sum_{(i,j,k)\in B}\left |1-G(z)_{ijk} \right |
\end{split}
\end{equation}  
where $B$ are the voxels at the edge,  $G(z)_{ijk}$ the generated lesion and $K$ the number of samples

\subsubsection{\textbf{Minkowski functional}}
\label{subsub:Minkowski}
The Minkowski functional is used in \cite{Chen2020} as a loss function to measure the distance between geometric features of volumes. For an $n$-dimensional space, there are $n+1$ measurements, namely porosity, specific surface area, average width and Euler's number for volumes (3D). 
This is used to assess if GANs captured correctly the geometric information of the images. For this, a second discriminator is trained to distinguish between Minkowski functional extracted from real or generated volumes. \cite{Chen2020} approach is similar to extract the features of medical images (using radiomics \citep{van2020radiomics} or like \cite{Pan2019137} (section \ref{subsub:Feature-consistent}) who uses a DL network to extract disease features) and use a discriminator to distinguish between features extracted from real and generated images.
Minkowski functional is a precise technique for analysing geometric structures that can also be used for biomedical image analysis \citep{depeursinge2014three, boehm2008automated} (the last explains these measures in more detail).

The volume ($V$), surface area ($S$), average width ($M$) and Euler's number ($X$) are defined in \cite{Chen2020} by the equation \ref{eq:Minkowski_Chen}. The porosity is defined by $V_p/V$ where $V_p$ is the connected pore space volume and $V$ the total volume.

\begin{equation} \label{eq:Minkowski_Chen}
\begin{split}
V=V(y) \\
S=\int_{\partial y} ds \\
M=\int_{\partial y} \frac{1}{2}[\frac{1}{r_1(s)}+\frac{1}{r_2(s)}]ds \\
X=\int_{\partial y} \frac{1}{r_1(s)r_2(s)}ds
\end{split}
\end{equation}

where $r_1(s)$ and $r_2(s)$ represents the maximum and minimum curvature on the surface, and $y$ the data (3D volume). 

The $V$, $S$, $M$ and $X$ can also be defined by equation \ref{eq:Minkowski_boehm} \citep{boehm2008automated}. 

\begin{equation} \label{eq:Minkowski_boehm}
\begin{split}
V=n_{voxeis} \\
S=-6V+2n_{faces} \\
M=3V-2n_{faces}+n_{edges} \\
X=-V+n_{faces}-n_{edges}+n_{vertices}
\end{split}
\end{equation}
where $n_{voxeis}$, $n_{faces}$, $n_{edges}$ $n_{vertices}$ are the number of voxels, faces, edges and vertices, respectively.

The Minkowski functional is more commonly used for evaluating geometric features, but when the features it defines are important for a realistic generation, the use of such metrics is recommended. \cite{Chen2020} could also have chosen to compare the real and fake features using other metrics, e.g. MAE, MSE or CE, instead of training a discriminator, which would reduce the computational power required, but the loss could not be as meaningful as using the discriminator. However, no conclusion can be drawn as this approach has not been researched in depth. 

\section{Appendix - Evaluation metrics}
\label{app:sec_Evaluation_Metrics}
\subsection{Generation Task}
\label{app:sec_Evaluation_Metrics_Generation_Task}
\subsubsection{\textbf{Correlation Coefficient (CC), Normalised Cross Correlation (NCC), Pearson CC (PCC)}}
\label{subsub:CC_NCC_PCC}
The CC and NCC \citep{PCCref} measure the linear association of two variables. They are used in \cite{Yang2021130, Baumgartner20188309, Harms2020, Lei2020b, Liu2020} to evaluate the quality of the generated volumes. The main problem with CC is that it depends on the amplitude of the variables being compared, i.e. the size of the signals, which can be a problem when comparing scans from different instruments, for example. By using the normalised version, i.e. the NCC, this dependency is eliminated \citep{Baumgartner20188309}. Its calculation is very fast and efficient, so it is recommended for real-time systems. Therefore, the use of NCC is highly recommended to statistically evaluate the agreement between real and synthetically generated datasets. NCC is generally defined by the equation \ref{eq:NCC}, and by the equation \ref{eq:NCC_image} when adapted to volume. 

\begin{equation} \label{eq:NCC}
\begin{split}
NCC=cos\theta=\frac{a\cdot b}{|a||b|}=\frac{\sum_ia_ib_i}{\sqrt{\sum_ia_i^2}\sqrt{\sum_ib_i^2}}
\end{split}
\end{equation}

where $\theta$ is the angle between the two vectors ($a$ and $b$), and -1 corresponds to not correlated and 1 to highly correlated.

In the context of image comparison:
\begin{equation} \label{eq:NCC_image}
\begin{split}
NCC=\frac{1}{n_i n_j n_k} \sum_{i,j,k}^{n_i n_j n_k}\frac{1}{\sigma_x\sigma_{\hat{x}}}[x_{i,j,k} \times \hat{x}_{i,j,k}]
\end{split}
\end{equation}
where $n_i$, $n_j$ and $n_k$ the dimensions in the $x$, $y$ and $z$ axis, i.e., $n_i\times n_j \times n_k$ is the number of voxels, $\sigma_x$ and $\sigma_{\hat{x}}$ the standard deviation of each volume, $x_{i,j,k}$ and $\hat{x}_{i,j,k}$ the voxel values of each volume.

PCC is also the correlation between two datasets, such as CC and NCC, and in the context of image comparison they are all equivalent, but NCC is recommended in this context due to the magnitude invariance. PCC is mentioned in \cite{Schaefferkoetter20213817} to evaluate the downstream task.

We decided to include these metrics in the voxel-wise evaluation section, since they utilize the values of each voxel to measure the linear association between the real and fake data distributions.
 
\subsubsection{\textbf{Cross Entropy (CE)}}
\label{subsub:CE_eval}
CE has already been explained in section \ref{subsub:CE}. Since this metric is able to compare two distributions, it is a good metric to compare the distribution of the real dataset and the generator. Lower values of CE correspond to better results, i.e. a distribution that is closer to the real distribution. It was used to assess the generated data in \cite{Yang2017679, Liu20212843} and to evaluate the downstream task in \cite{Wang20172317}.

\subsubsection{\textbf{MAE, Mean Absolute Percentage Error (MAPE)}}
\label{subsub:MAE_MAPE}
The MAE is already explained in section \ref{subsub:MAE_MSE}. 
MAE is used in \cite{Lei2020b, Rusak202011, Qin2021S98, Harms2020, Yang2021415, Liu2020, Zeng2019759, Hu2022145, Pan2019137} for the assessment step.
MAPE (equation \ref{eq:MAPE}) is similar to the MAE, but the values are different because MAPE is a percentage and the value is relative to the real value \citep{de2016mean}. It is used in \cite{Han2019} to evaluate the results of the nuclei counter GAN-based model.

MAPE is better suited for comparisons between models, even if they have been trained with different datasets, as the error is always relative to that dataset. However, comparing such models is not good practice because the different datasets are of different complexity. MAPE values are also easier to understand than MAE values. Although MAE is more commonly used, MAPE is also highly recommended.

\begin{equation} \label{eq:MAPE}
\begin{split}
MAPE=\frac{100\%}{n}\sum^n_{i=1}|\frac{\hat{x}-x}{x}|
\end{split}
\end{equation}
where $\hat{x}$ and $x$ are the output and ground truth.

\subsubsection{\textbf{MSE, Normalised MSE (NMSE), Normalised Root MSE (NRMSE)}}
\label{subsub:MSE_NMSE_NRMSE}
The MSE is already explained in section \ref{subsub:MAE_MSE}. The MSE is used in \cite{Rusak202011, Yang2019181, Shen20213250} to assess the quality of the generated data and in \cite{Schaefferkoetter20213817} in the downstream task.
The NMSE is a normalised version of the MSE with respect to the signal intensity used in \cite{Moghari2019, Lei2020b}.
The NRMSE (equation \ref{eq:RMSE_NRMSE}) is a normalised version of the Root MSE  used in \cite{Yang2021130}. Both NMSE and NRMSE are suitable for comparing models with different scales, i.e. comparing datasets from different sources, as is MAPE, which can be useful for comparing synthetic data that is intended to have different scales from the real dataset, e.g. generating PET scans from low-count PET, or CT scans from low-dose CT. However, as mentioned in section \ref{subsub:HU_SNU}, other metrics must be used to ensure that the intensity of the voxels matches the real tissue HU values.

\begin{equation} \label{eq:RMSE_NRMSE}
\begin{split}
RMSE=\sqrt{\frac{\sum_{i=1}^n(x_i-\hat{x})^2}{n}} \\ NRMSE= \frac{RMSE}{{\omega}} 
\end{split}
\end{equation}
where $\hat{x}$ and $x$ are the output and ground truth, and $\omega \in \{x, mean, std\}$ (depending on what is to be normalised). NMSE is identical to NRMSE, but without the $\sqrt{\cdot}$.

\subsubsection{\textbf{Intersection-over-Union (IoU)}}
\label{subsub:IoU}
The IoU or Jaccard index \citep{taha2015metrics} is used in \cite{Kniaz20203, Yang2017679, Liu20212843} to measure the intersection of two distributions over the union of the two distributions (equation \ref{eq:JAC}). The goal of these works is to produce 3D volumes from 2D/2.5D images. The result is a voxel grid where each voxel is 1 or 0, which allows the calculation of such a metric. This metric cannot be used to evaluate the generation of data where the intensities of the voxels are not binary, e.g. when transferring from MRI to CT. Therefore, this metric is only relevant when the output is a binary grid.

\begin{equation} \label{eq:JAC}
\begin{split}
JAC= \frac{TP}{TP+FP+FN}
\end{split}
\end{equation}
where TP, FP and FN are True positives, false positives and false negatives, respectively. This metric is formally adapted to volumes by equation \ref{eq:JAC_vol}.

\begin{equation} \label{eq:JAC_vol}
\begin{split}
JAC= \frac{\sum_{i=1,j=1,k=1}^{n_i n_j n_k}(\hat{x}\cdot x)}{\sum_{i=1,j=1,k=1}^{n_i n_j n_k}(\hat{x}+ x)}
\end{split}
\end{equation}
where $\hat{x}$ is the generated volume, and $x$ the ground truth.

This is formally related with the Dice score (section \ref{subsub:DSC}) by the equation \ref{eq:JAC_DICE}.

\begin{equation} \label{eq:JAC_DICE}
\begin{split}
JAC=\frac{DICE}{2-DICE}
\end{split}
\end{equation}

Both metrics measure sensibly the same, so there is no need to use both metrics. As can be seen in this review, IoU is used to assess the quality of the generated scans, and the Dice score for the downstream task (frequently used to assess segmentations).

\subsubsection{\textbf{Peak Signal-to-Noise Ratio (PSNR)}}
\label{subsub:PSNR}
The PSNR (equation \ref{eq:PSNR}) measures the proportion of desired signal relative to background noise in decibels, and is commonly used for volumetric medical imaging applications of GANs. This metric is often used to assess the quality of enhanced images or reconstructions. \cite{Li2020} is the only work in which the PSNR is used to evaluate the downstream task and not the generation task, in contrast to \cite{Moghari2019, Rusak202011, Ma2020, Harms2020, Lei2020b, Gu2021, Yang2021415, Liu2020, Yang2019181, Zeng2019759, Hu2022145, Lin2021, Pan2019137, Tang20201775, Yang2021130}. 

\begin{equation} \label{eq:PSNR}
\begin{split}
PSNR=20\cdot log_{10}(\frac{MAX_{I}}{\sqrt{MSE}})
\end{split}
\end{equation}
where MSE is defined in Equation \ref{eq:MSE}, and $MAX_{I}$  the maximum possible intensity of a voxel of the scan. PSNR metric if close to the inverse of MSE (for PSNR the higher, the better), however, PSNR measure the distance between signal intensities and not absolute values. Although this metric is widely used, it does not correlate with human perception of quality, so metrics such as SSIM and MS-SSIM should be used instead. However, it is able to measure the degree of degradation of a scan, which is important for super-resolution and reconstruction problems, e.g. for the generation of high-resolution MRI scans from low-resolution MRI scans to increase acquisition speed.

\subsubsection{\textbf{Structural Similarity Index Measure (SSIM), Multi-Scale Structural Similarity Index Measure (MS-SSIM)}}
\label{subsub:SSIM_MS-SSIM}
The SSIM \citep{wang2004image} is widely used to extract the structural information from digital images and videos, just like human visual perception. SSIM (equation \ref{eq:SSIM_1}) is then composed by three components: luminance, contrast and structure (equation \ref{eq:SSIM_l_c_s}).
This metric differs from MAE, MSE and PSNR because it does not only calculate the absolute error, but also takes into account the dependencies between voxels, especially when they are close to each other. These dependencies contain important information about the structure of the evaluated objects. Its operation is based on the physiology of human perception of quality, which makes this metric very suitable for assessing the quality of generated volumes. This metric needs the original volumes, i.e., the ground truth, to make its comparisons. The comparison is made using a sliding window approach with a window size chosen by the user (e.g. 8×8×8).
It is used in \cite{Moghari2019, Rusak202011, Ma2020, Gu2021, Yang2021415, Lin2021, Pan2019137, Yan2018, Tang20201775, Yang2021130}.

\begin{equation} \label{eq:SSIM_l_c_s}
\begin{split}
l(X,\hat{X})=\frac{2\mu_{X}\mu_{\hat{X}}+C_1}{\mu_X^2+\mu_{\hat{X}}^2+C_1} \\
c(X,\hat{X})=\frac{2\sigma_X\sigma_{\hat{X}}+C_2}{\sigma_X^2+\sigma_{\hat{X}}^2+C_2} \\
s(X,\hat{X})=\frac{\sigma_{X\hat{X}}+C_3}{\sigma_X\sigma_{\hat{X}}+C_3}
\end{split}
\end{equation}

where $X$ and $\hat{X}$ are two discrete non-negative signals aligned with each other, i.e., in this case two volume patches extracted from the same spatial location from the two volumes to be compared, $\mu_{X}$, $\sigma_{X}$ and $\sigma_{X\hat{X}}$ the mean of $X$, variance of $X$ and the covariance of $X$ and $\hat{X}$, and $C_1$, $C_2$ and $C_3$ are constants given by: $C_1=(K_1L)^2$, $C_2=(K_2L)^2$, $C_3=C_2/2$, where $L$ is the dynamic range of the voxel values, and $K_1,K_2\ll1$ two scalar constants.

\begin{equation} \label{eq:SSIM_1}
\begin{split}
SSIM(X,\hat{X})=[l(X,\hat{X})]^{\alpha}\cdot[s(X,\hat{X})]^{\beta}\cdot[s(X,\hat{X})]^{\gamma}
\end{split}
\end{equation}
where the values of $\alpha$, $\beta$ and $\gamma$ depend on the relative importance of luminance, contrast and structure, respectively. When $\alpha=\beta=\gamma=1$, the SSIM is given by equation \ref{eq:SSIM_2}.

\begin{equation} \label{eq:SSIM_2}
\begin{split}
SSIM(X,\hat{X})=\frac{(2\mu_X\mu_{\hat{X}}+c_1)(2\delta_{X\hat{X}}+c_2)}{(\mu_X^2+\mu_{\hat{X}}^2+c_1)(\delta_X^2+\delta_{\hat{X}}^2+c_2)}
\end{split}
\end{equation}

MS-SSIM \citep{wang2003multiscale} is based on SSIM at multiple scales, i.e. the input image is subsampled by a factor of 2 $M$ times and the similarity is calculated by the equation \ref{eq:MS-SSIM}. MS-SSIM has been shown to perform better than the single scale SSIM approach, as it is more robust to variations in viewing conditions. The only disadvantage compared to SSIM is the speed of inference, though it is still very fast.
However, it is intriguing that this metric was only used in \cite{Hu2022145}. Both metrics can be adapted to work with volumes and capture the semantic perception of all three dimensions (i.e. depth features as well), making MS-SSIM highly recommended for evaluating volumes in problems such as reconstruction, as long as ground truth is available. 

\begin{equation} \label{eq:MS-SSIM}
\begin{split}
MSSSIM(X,\hat{X})=\\ [l_m(X,\hat{X})]^{\alpha_M}\cdot\prod_{j=1}^M[c_j(X,\hat{X})]^{\beta_j}\cdot[s_j(X,\hat{X})]^{\gamma_j}
\end{split}
\end{equation}
where $M$ is the highest scale, i.e., smaller volume, and $c_j(X,\hat{Y})$ and $s_j(X,\hat{X})$ the contrast and structure comparison at the $j^{th}$ scale

\subsubsection{\textbf{Visual}}
\label{subsub:visual}
Visual assessment is the most commonly used evaluation to obtain a human perception of the images/scans generated. So far, there is no metric that can truly replace human perception, even though some metrics, e.g., FID or MS-SSIM, try to do so, but still fail in some specific cases.

Normally, researchers perform a visual inspection when training generative models, but experts in the field are needed to accurately assess the quality of the generated data. 
The visual Turing test is a more sophisticated visual assessment in which experts are asked to classify a series of images as real or fake, without prior knowledge of the data. The best result is a percentage of correct answers close to 50\%, which means that the experts were not able to make a distinction and therefore a random choice was made. Instead of a binary selection, a value in a specific range can also be specified, e.g. between 0 and 5, where 0 is certainly fake and 5 is certainly real. After a reasonable number of answers (which depends on the task) have been obtained from different experts (a reasonable number would be e.g. 3), a confusion matrix can be created and the accuracy, sensitivity, specificity, precision and other metrics can be calculated.

As with MS-SSIM, it was expected that more papers would be found using the visual Turing test (which was only used in \cite{Bu2021670, Han2019729}). Although visual assessment should always be an essential approach for evaluating a GAN model, it is subjective. Therefore, researchers opt to use objective metrics such as SSIM or PSNR and evaluate the synthetic data applied to the downstream task.

\subsubsection{\textbf{Clustering/t-Distributed Stochastic Neighbour Embedding (t-SNE)}}
\label{subsub:t-SNE}
Clustering is used to divide a population of data points into a certain number of groups \citep{Clusteringref} specified by the user. One of the most used algorithms for visualization of high-dimensional data distributions is the t-SNE \citep{van2008visualizing}. It is an important technique to visualise whether the distribution of the generated data follows the real data distribution. 

It was used in \cite{Muzahid202120} to test whether the class distribution of the generated samples could be easily split, i.e. whether the generator was able to learn to generate data from each class, and in \cite{Han2019729} to assess whether the distribution of the generator was similar to the distribution of the real data. t-SNE is then a qualitative metric that does not provide an absolute value of the quality of the synthetic data, but shows the user how the distribution of the synthetic and real data overlap in an easy-to-understand 2D graph. Therefore, this is recommended for any task of GANs, provided it is advantageous to have a generator that follows the same distribution as the real dataset, or when different classes are to be learned from the GAN, e.g. in a cGAN.

\subsubsection{\textbf{Semantic Interpretability Score (SIS), Shape-score(S-score)}}
\label{subsub:SIS_S-score}
The SIS \citep{seitzer2018adversarial} and S-score \citep{Zhang20189242} are inspired by the Inception score (section \ref{subsub:FID_IS}). SIS is the Dice (section \ref{subsub:DSC}) overlap between the segmentation of a reconstructed image using a pre-trained segmentation network and the ground truth segmentation. The S-Score is the same, but uses a multiclass dice score instead. These metrics assume that scans with better visual quality, e.g. with defined boundaries and fewer distortions, are better segmented by the pre-trained model. These are similar to IS, FID, perceptual, latent vector and feature consistency losses in that they use a pre-trained network to evaluate the generated images. 

However, as mentioned in the section \ref{subsub:summary_intermediate_layers}, the perceptual loss as well as IS and FID should be avoided as they use a model trained on a dataset consisting of different classes than those used to train the GAN. SIS and S-score are particularly interesting because they are trained on the same dataset used to train the GAN. Therefore, the segmentation networks have been trained to extract the correct features for segmentation in that specific data distribution. However, it is not guaranteed that good results correspond to high realism, as the generator may learn to accurately produce only the part to be segmented and neglect the remaining part. Nevertheless, SIS and S-score are preferable to IS and FID. The authors claim that these metrics are more related to human judgement than SSIM and PSNR, and better scores correspond to less geometric distortion. SIS is only used by \cite{Ma2020}, and the S-score is used by \cite{Zhang2019183, Zhang20189242, Cai2019174}.

\subsubsection{\textbf{Fréchet Inception Distance (FID), Inception Score (IS)}}
\label{subsub:FID_IS}
The IS is known to correlate with human judgement \citep{salimans2016improved}. To calculate this metric, first the Inception v3 model \citep{szegedy2016rethinking} trained on the ImageNet dataset is needed. The generated images are fed into the model, which gives a score as output, i.e., the label distribution ($p(y|x)$, where $y$ is a set of labels and $x$ the image). A high score means that the images are varied and look realistic, i.e. they resemble other images. The inception model assigns the images to a certain class, i.e. it is a classifier. For a synthetic image to receive a high score, it must be clearly assigned to a certain class. A good score means that the collection of synthetic images has low entropy, i.e., all the images are strongly assigned to a specific class. After $p(y|x)$ is determined, the KL divergence for all generated images is calculated and the IS is determined using the equation \ref{eq:IS}.

\begin{equation} \label{eq:IS}
\begin{split}
IS=e^{\mathbb{E}_xKL(p(y|x)||p(y))}\\
KL=p(y|x)\cdot (log(p(y|x))-log(p(y)))
\end{split}
\end{equation}
where $||$ is the divergence between two distributions, which in this case is calculated by the KL divergence, and $p(y)$ the marginal distribution.

FID is a metric widely used to compare the distribution of  real images with images generated by GANs. Introduced by \cite{Heusel2017}, who claims that the FID metric captures the similarity between generated images and real images better than IS, as IS does not compare the statistics of real and synthetic data, making it the most commonly used metric to assess the quality of GANs (for 2D).

FID uses the feature vector before calculating the probabilities (before the last layer of the inception model). These activation values (the feature vector) are then summarised as a multivariate Gaussian with mean and covariance. These statistics are calculated for all images in the real and generated datasets, and finally the Fréchet distance (Wasserstein-2 distance) between the two distributions is calculated (equation \ref{eq:FID}).

\begin{equation} \label{eq:FID}
\begin{split}
d^2((m,C),(m_w,C_w))=||m-m_w||^2_2+Tr(C+C_w-2(CC_W)^{1/2})
\end{split}
\end{equation}
where $p(\cdot)$ and $p_w(\cdot)$ are the probabilities of real world data and generated data, $m$, $C$ the mean and covariance from $p(\cdot)$ and $m_w$, $C_w$ from $p_w(\cdot)$, $Tr$ is the trace of the matrix (i.e., sum of its diagonal entries from upper left to down right), and $||m-m_w||^2_2$ the square of the difference of the two matrices. 

\cite{Heusel2017} shows that FID is consistent with human judgement and with the increase of disturbances on the original images, but it is not commonly used for volumetric assessment, nor is IS. For its calculation, the data must be 2D, i.e. volumetric data cannot be used because the inception model is trained on 2D data. In order for them to be used, the volumes are usually sliced in all three axes to allow the use of the FID, but this practice is not well accepted and therefore not widely used.

These two metrics have some more problems that have already been mentioned in the \ref{subsub:summary_intermediate_layers} section. They assume that the inception model is able to correctly classify any existing image. However, in cases where a class is to be generated that is not represented in the dataset used to train the inception model, the score will always be low, even if the generated images are of good quality, i.e., the dataset used to train the GAN must contain the same or fewer classes than the dataset used to train the inception model, with no difference between these classes. Therefore, the use of such metrics to assess the quality of synthetic medical data should be avoided.

Since the inception model is a CNN classifier, the classification may rely only on texture and low-level features, rather than also relying on high-level features such as whole objects, which may result in a high score for images that are clearly not realistic. These metrics are also not able to detect mode collapse or overfitting. Other metrics must be used for this, e.g. MS-SSIM and visual assessment are always the most reliable choice, specially for volumetric data.

The IS \citep{salimans2016improved} is used in \cite{Pesaranghader202167}.
The FID is used in \cite{Pesaranghader202167, Dikici2021, Jung202079, Hu2022145, Li20195530}.

\subsubsection{\textbf{Absolute permeability, Minkowski functional}}
\label{subsub:abs_perm_mink}
 Absolute permeability is used in \cite{Krutko2019} to assess the quality of generated porous media volumes. It is usually used in conjunction with Minkowski functional. This metric is calculated using a pore network model, which is not explained in detail in the paper, but it suggests that it is a computer simulator that accepts 3D volumes of porous media consisting of voxels as input.
The Minkowski functional is already explained in the section \ref{subsub:Minkowski}. \cite{Chen2020, Krutko2019, Mosser2017} use it in the evaluation step. These metrics measure important characteristics of porous media that must be learned by the GAN in order to reconstruct volumes that are as realistic as possible. Therefore, in these cases, using these metrics to evaluate and compare the trained GANs is the best approach.

\subsubsection{\textbf{Compliance, Fraction of Unmachinable Voxels}}
\label{subsub:comp_fraction_un_voxels}
The compliance and fraction of unmachinable voxels are used in \cite{greminger2020generative} to measure the difference between machinable designs generated by GANs versus the traditional topology optimisation algorithm. Compliance roughly stands for the inverse of stiffness, so the goal of \cite{greminger2020generative} is to archive designs with minimum compliance, i.e. a design that maximises stiffness for a given target mass of the object. The minimum compliance for the GAN-based solution is given by the equation \ref{eq:compliance}. For the sake of brevity, please refer to \cite{greminger2020generative}, where it is explained in more detail how the compliance is calculated.

\begin{equation} \label{eq:compliance}
\begin{split}
c(z)=\frac{1}{2}u^TK(e)u, \frac{V(G(z))}{V_0}=v_f
\end{split}
\end{equation}
where $z$ is the input latent vector of the generator, $u$ is the vector of nodal displacements of the finite element model, $K(e)$ the global stiffness matrix, $e$ the vector of element modulus values calculated from $z$, $V(G(z))$ the volume of the generated model, $V_0$ the volume of the model domain, and $v_f$ the target volume fraction.

The fraction of unmachinable voxels is the fraction of voxels that cannot be reached in the machining process. The volumes produced are to be machined with a 3-axis vertical milling machine with a single setup, and it is imperative that each voxel is reachable in this process so that their production is possible. The proportion of voxels that cannot be machined is therefore given by the number of voxels that cannot be reached / total number of voxels. 

These metrics are very specific to this problem, but they are highly recommended for any related task that physically involves the production of such volumes. The use of metrics that express geometric features and physical constraints are essential for evaluating the volumes produced. For example, it is not useful to use these metrics to access the results of an image translation task (e.g. CT to MRI).

\subsubsection{\textbf{Hounsfield Unit (HU), Spatial Non-Uniformity (SNU)}}
\label{subsub:HU_SNU}
HU \citep{kamalian2016computed} is a relative scale to measure the attenuation of the X-ray beam in a given voxel, calculated by equation \ref{eq:HU}. HU is approximately linearly proportional to the physical density of the material (with the exception of some substances such as iodine), but relative to the intensity of the X-rays. Therefore, in contrast to MRI, HU describes the actual density of the object under investigation, be it a porous media or a human organ, e.g. bones have higher HU values than fat, and white matter lower values than grey matter.

\begin{equation} \label{eq:HU}
\begin{split}
HU=\frac{\mu_{material}-\mu_{water}}{\mu_{water}}*1000
\end{split}
\end{equation}
where $\mu$ corresponds to the attenuation of the value of the X-ray bean in a given material, i.e., in a given voxel. Higher HU values appear brighter and lower HU values appear darker. 

This is why calculating the difference between a real image and a reconstructed image using these values is an important method for evaluating the quality of the generated CT data. \cite{Liu2020} want to generate synthetic CT scans based on MRI scans. Therefore, the HU values must represent the real properties of each tissue, so the MAE value between the real CT scans and the synthetic CT scans is calculated to evaluate the realism of the data. The histogram of HU values is also used, which shows a good overlap of the values. This work has shown that GANs can be used to predict the HU values of tissues, which can further increase the speed of treatments.

HU is also used to evaluate the synthetic CT scans in \cite{Zhang2021}, which computes the error (mean and standard deviation) of each organ between the real value and the value of the generated images. 

The SNU is nothing more than the difference of HU in the same material, i.e., the difference of HU of a specific tissue between the generated scans, mentioned in \cite{Harms2020}.

These metrics are no different from other voxel-wise comparisons, but the authors have placed a higher value on them given their importance for CT reconstruction.

\subsubsection{\textbf{Reconstruction resolution}}
\label{subsub:recons_reso}
The reconstruction resolution is used to assess the quality of reconstruction of single particles from noisy projections. This metric is very specific to the work of \cite{Gupta2021759}, who uses a cryo-EM physics simulator as a generator, and depends on the noise level of the input. If this simulator is not used, this metric is not relevant.

\subsubsection{\textbf{Volume fractions, Triple Phase Boundary/Double Phase Boundary (TPB/DPB) densities, relative surface area, relative diffusivity, surface area, Two-Point Correlation Function (TPCF)}}  
\label{subsub:TPB/DPB}
All these metrics are used for microstructural characterisation. 
The average volume fractions, volume fraction variations, and TPB/DPB densities are measurements used in \cite{Sciazko20211363} to evaluate the quality of 3D fuel cell electrode microstructures. Volume fraction, relative surface area and relative diffusivity are measurements used in \cite{Kench2021299} to evaluate the quality of various microstructures, including a battery cathode. The phase volume fraction, relative diffusivity, specific surface area, TPCF and TPB are used in \cite{Gayon-Lombardo2020} to measure the quality of electrode microstructures.

These are just characteristics inherent to the volume they intend to generate synthetically, not really comparison metrics. These cases are important to mention because they use a specific characteristic of the object to assess the quality of the generated volumes, instead of using just voxel-wise metrics. Thus, the mean and variance of these characteristics are compared between the real and generated data using graphs, where a higher overlap corresponds to a higher degree of fidelity of the synthetic data compared to the real data.

\subsection{Downstream task}
\label{app:sec_Evaluation_Metrics_Downstream_Task}
\subsubsection{\textbf{Accuracy, Sensitivity, Specificity, Precision, Average False Positives (AFP), Type-I/Type-II errors}}
\label{subsub:acc_sen_spe_pre_AFP_type_I_II}
The metrics accuracy, sensitivity, specificity \citep{baratloo2015part}, precision \citep{Preref} and AFP are often used in machine learning (equation \ref{eq:Acc_sen_spe_pre}). To calculate these metrics, a confusion matrix \citep{confusionmatrixref} is usually created. The metrics used should be chosen carefully depending on the problem, as high performance on one of these metrics does not necessarily mean a good model, e.g. when classifying a patient as positive or negative for a particular disease, high sensitivity is preferable to high specificity, as it is usually better to classify someone as sick who does not actually have a disease than the opposite. 

These metrics are mainly used to evaluate other downstream tasks instead of the generation task. Accuracy is measured in \cite{Pesaranghader202167, Jung202079, Zhuang2019, Wei2020, Yan2018, Muzahid202120, Wang20172317, Chen2021961, Liu2020O47}.

Sensitivity, also known as recall, is measured in \cite{Bu2021670, Momeni2021, Zhuang2019, Lei2020a, Chen2021961, Ho2019, Han2019, Baniukiewicz2019}. 
Specificity is measured only in \cite{Lei2020a}.

Precision, also known as positive predictive valve, is measured in \cite{Zhuang2019, Chen2021961, Ho2019, Han2019, Baniukiewicz2019}.

The AFP metric is used in \cite{Dikici2021}, which is the average of false positives metastases detected per patient.  

Type-I and Type-II errors, also known as false positive and false negative errors, are used in \cite{Chen2021961}. It can be also be used as $1-Spe$ and $1-Sen$, respectively. Type-I error is rejecting the null hypothesis when this is actually true, and the Type-II is accepting the opposite, i.e., accepting the null hypothesis when this is actually false.

\begin{equation} \label{eq:Acc_sen_spe_pre}
\begin{split}
Acc=\frac{TP+TN}{TP+TN+FP+FN}\\
Sen=\frac{TP}{TP+FN} \\
Spe=\frac{TN}{TN+FP} \\
Pre=\frac{TP}{TP+FP} 
\end{split}
\end{equation}
where TP, TN, FP and FN are true positive, true negative, false positive and false negative respectively.

\subsubsection{\textbf{Area Under the Curve (AUC), Receiver Operating Characteristic curve (ROC)}}
\label{subsub:AUC_ROC}
AUC \citep{AUCref} is the area under the ROC curve \citep{ROCref} and is intrinsically related to the true positive rate (TPR, also known as $Sen$) and the false positive rate (FPR, $1-Spe$ or equation \ref{eq:FPR}). The ROC curve is the graph of TPR on the y-axis and FPR on the x-axis, to evaluate binary classifications. A high AUC value means that the model can distinguish well between positive and negative classes. If the value is close to 0.5, the model is not able to distinguish between them.

The AUC is used in \cite{Momeni2021, Pan2019137, Yan2018}. 
This metric is also used to indirectly assess the quality of the volumes generated, by measuring the performance of the downstream task where synthetic data was used, e.g., a classifier trained with synthetic data.

\begin{equation} \label{eq:FPR}
\begin{split}
FPR=\frac{FP}{FP+TN}
\end{split}
\end{equation}

\subsubsection{\textbf{Attenuation Correction (AC)}}
Attenuation in PET scans is the loss of detection due to absorption caused by photon-tissue interactions in the body and scattering outside the detector field, resulting in artefacts. The attenuation is given by the equation \ref{eq:attenuation}.

\begin{equation} \label{eq:attenuation}
\begin{split}
\frac{I}{I_0}=e^{-\mu L}
\end{split}
\end{equation}
where $I$ and $I_0$ are the unattenuated and attenuated PET signals, $\mu$ and $L$ the linear attenuation coefficient (LAC) and thickness of the tissue. The LAC (also known as $\mu$ map) is calculated using other imaging sources such as CT scans where the HU values are transformed into PET LAC values to perform AC \citep{carney2006method}. 

AC can also be performed with MRI scans, but using MRI has problems with certain body structures, e.g. the bones \citep{chen2017attenuation}. \cite{Schaefferkoetter20213817} use synthetic CT scans generated from MRI scans to perform AC and compare them with MRI-based correction, i.e. the relative differences between the $\mu$ maps generated with PET/CT, PET/MRI and PET/synCT (the synthetically generated scans). The purpose of this comparison is not to evaluate the generation itself, but how significantly the synthetic CT scans were able to improve the $\mu$ maps and the corresponding PET scans compared to the use of MRI scans.

\subsubsection{\textbf{Average Hausdorff Distance (HD) and average surface HD}}
The average HD (equation \ref{eq:AHD}) and the average surface HD (equation \ref{eq:SHD}) are used in \cite{Zhang2021, Lei2020a} to measure the distance between voxel sets in ground truth and prediction \citep{kazemifar2018segmentation}. The HD is the greatest distance between a voxel in one volume and the closest voxel in the other volume in millimetres.

\begin{equation} \label{eq:AHD}
\begin{split}
AHD(X,\hat{X})= \\ max(\frac{1}{|X|}\sum_{x\in X}min_{\hat{x}\in \hat{X}}d(x,\hat{x}),\frac{1}{|\hat{X}|}\sum_{\hat{x}\in \hat{X}}min_{x\in X}d(\hat{x},x))
\end{split}
\end{equation}

\begin{equation} \label{eq:SHD}
\begin{split}
ASHD(X,\hat{X})=\\ \frac{1}{2}(\frac{1}{|X|}\sum_{x\in \hat{X}}min_{\hat{x}\in \hat{X}}d(x,\hat{x})+\frac{1}{|\hat{X}|}\sum_{\hat{x}\in Y}min_{x\in X}d(\hat{x},x))
\end{split}
\end{equation}
where $X$ and $\hat{X}$ are the voxel sets of the ground-truth and generated segmentation, and $d(x,\hat{x})$ the Euclidean distance from voxel $x$ to voxel $\hat{x}$.

For example, in \cite{Zhang2021}, the average HD and surface HD are used to evaluate the performance of prostate and at-risk organ segmentation, but not the generation task. As mentioned earlier, the better performance using synthetic data compared to not using such data suggests that the synthetic data is of good quality and provides an advantage for other deep learning algorithms.

\subsubsection{\textbf{Center of Mass Distance (CMD), Percentage Volume Difference (PVD), Mean Surface Distance (MSD), Residual Mean Square Distance (RMSD)}}
All these metrics are used in \cite{Lei2020a}. CMD and PVD are used to measure the distance between the centre distances and volume differences of the automatic and manual segmentation contours. The MSD is used to measure the surface distances of the two contours. The RMSD is similar to the MSE but is calculated between distances rather than absolute voxel values, just as the MSD is similar to the MAE but compares distances between surfaces rather than voxel intensities. CMD is simply the position difference between the real and generated segmentation centroid. CMD, MSD and RMSD are measured in millimetres.

\subsubsection{\textbf{Competition Performance Metric (CPM)}}
The CPM \citep{niemeijer2010combining} is the average sensitivity at seven predefined false positive rates: $1/8$, $1/4$, $1/2$, $1$, $2$, $4$ and $8$ false positives per scan. The sensitivity is explained in section \ref{subsub:acc_sen_spe_pre_AFP_type_I_II}. CPM reduces the free-response receiver operating characteristic (FROC) to single numbers. The FROC is very similar to the ROC curve (see section \ref{subsub:AUC_ROC}), where the sensitivity is on the y-axis and the average number of false positives per volume is on the x-axis.
\cite{Bu2021670, Han2019729} use this metric to compare the results with the lung nodule analysis (LUNA16) challenge \citep{setio2017validation}. When better performance is achieved by using synthetic data, it means that the synthetic data is of good quality and adds important information that is captured by other deep learning algorithms.
 
\subsubsection{\textbf{Dice Similarity Coefficient (DSC), F1-score}}
\label{subsub:DSC}
The DSC \citep{taha2015metrics} (equation \ref{eq:DSC}) and F1-score \citep{f1_scoreref} (equation \ref{eq:F1}) are two metrics that measure the similarity between two samples. Both are equivalent when applied to Boolean data (equation \ref{eq:DSC_F1}). 
DSC is also known as the overlap index, and is the most commonly used metric for evaluating segmentation in volumetric medical imaging. This term is usually used in the context of computer vision. DSC is directly proportional to the Jaccard index, as explained in section \ref{subsub:IoU}.  
The F1-score is used when the focus is on maximising both precision and recall, being a harmonic mean between both. 
The F1-score is mentioned in \cite{Chen2021961, Ho2019, Han2019, Baniukiewicz2019, Liu2020O47}. The macro F1-score is used in \cite{Zhuang2019}. The macro F1-score is a variation of the F1-score, where the macro F1-score is the average of the F1-scores of the individual classes.
The DSC is used in \cite{Zhang2021, Rusak202011, Ma2020, Li2020, Lei2020a, Zhang2019183, Zhang20189242, Cai2019174, Chen2021961}. \cite{Li2020} is a rare example of a work that uses this metric to evaluate the quality of the generator's output, but these metrics are mainly used to evaluate other downstream tasks, such as hippocampal subfield segmentation \citep{Ma2020}.

\begin{equation} \label{eq:DSC}
\begin{split}
DSC=\frac{2|X\cap \hat{X}|}{|X|+|\hat{X}|}
\end{split}
\end{equation}
where $X$ and $\hat{X}$ are the two sets of data to be compared.
\begin{equation} \label{eq:F1}
\begin{split}
F_1=\frac{2\times Precision \times Recall}{Precision+Recall}
\end{split}
\end{equation}
where precision and recall are calculated as defined in section \ref{subsub:acc_sen_spe_pre_AFP_type_I_II}.
\begin{equation} \label{eq:DSC_F1}
\begin{split}
DSC=F_1=\frac{2TP}{2TP+FP+FN}
\end{split}
\end{equation}
where TP, FP and FN represent true positive, false positive and false negative respectively.

\subsubsection{\textbf{Noise SD}}
The noise SD is used in \cite{Li2020} to measure background noise in order to compare different methods of noise reduction. In this case, one GAN was used to generate synthetic noise and another GAN was used to noise reduction, but no metric was used to evaluate the first, only the second. Two denoising models were trained, one with Gaussian noise and the other with noise generated by the GAN. Lower noise SD scores were obtained with the second denoiser, suggesting that using a GAN that learns how to generate noise is a good strategy to increase the amount of data for training deep learning denoisers.

\subsubsection{\textbf{Tumour localisation error}}
Tumour localisation error is used in \cite{Wei2020} to evaluate the performance of the proposed tumour localisation method. \cite{Wei2020} believe that their CNN-based tumour localisation method does not perform better due to the discrepancy between the intensity of the digitally reconstructed radiographs from CT scans and the measured x-ray projections. To solve this problem, cone beam computed tomography (CBCT) scans are used, but not all patients have both CT and CBCT scans. Therefore, a GAN is trained to produce synthetic CBCT scans from CT.
Better tumour localisation performance using synthetic CBCT scans suggests that the generated data is of good quality for the proposed downstream task.
The tumour localisation error is the absolute difference between the actual and the predicted values.

\section{Appendix - Applications}
\label{sec:Applications}

\setcounter{table}{0}
\renewcommand{\thetable}{\Alph{section}.\arabic{table}}

This section summarizes all works in two different tables, sorted by application. Table \ref{tab:table5} contains all medical papers, sorted by application and organ, and Table \ref{tab:table6} contains all non-medical papers, also sorted by application and structure. Animal and cell studies were considered medical applications.

In Table \ref{tab:table5} it is possible to find the columns: \textit{Application}, \textit{Modality}, \textit{Study}, \textit{Network}, \textit{Organ}, and \textit{Notes}. As specified in the \textit{Organ} column, the cells have several sources, such as kidneys, embryos, neurons, and blood vessels. The last column highlights the main contribution of the paper. \textit{Image translation} is the main application of GANs in a medical context, as can be seen in Figure \ref{Fig:Application_nonNmedical}.

Figure \ref{Fig:Organ} shows the number of medical publications sorted by the number of papers on a specific organ. It can be seen that the most studied organ is the brain, followed by the lungs. The study of 3D structures of cells with GANs is also very popular.  Figure \ref{Fig:ArchToApp} shows all GAN architectures used in each application. The cGAN architecture is the most commonly used architecture in various applications, but it can be seen that CycleGAN-based architectures are the preferred choice for image translation and multimodal data. 

The application "General" refers to the use of GANs only for the generation of synthetic data without a specific purpose, or that only proposals are presented but not developed in the papers.

Table \ref{tab:table6} has the columns: \textit{Application}, \textit{Modality}, \textit{Study}, \textit{Network}, and \textit{Structure}. The last column contains the objects used in the study, although in some cases different objects are used, so it is simply written \textit{Objects} or \textit{Models}. The number of papers per application is shown in Figure \ref{Fig:Application_nonNmedical}, where \textit{Reconstruction} is the main application of GANs in a non-medical context.

\onecolumn

\begin{table*}[!t]
\begin{small}

\begin{center}
\caption{Compact overview of reviewed medical papers sorted by application, then by modality and year. The application, the studied organ and the highlights/notes are the main purpose of this table}
\label{tab:table5}

\begin{tabular}{|p{2cm}|p{1.7cm}|p{2cm}|p{2.5cm}|p{1.2cm}|p{6cm}|}
\hline

\textbf{Application} & \textbf{Modality} & \textbf{Study} & \textbf{Network} & \textbf{Organ} & \textbf{Notes} \\ \hline

Classification & CT & \cite{Bu2021670} & cGAN (3D-UNet/ 3D CNN) & Lung & Boost the overall performance of the lung nodule detection network by 2.53\% \\ \hline

Classification & CT & \cite{Pesaranghader202167} & CT-SGAN & Lung & Pretraining the nodule detection classifier on synthetic volumes and fine-tuning on the real data \\ \hline

Classification & CT & \cite{Han2019729} & MCGAN (U-Net/ Pix2Pix) & Lung & Generating lung nodules using bounding box and adding surrounding tissues \\ \hline

Classification & Microscopy & \cite{Baniukiewicz2019} & cGAN & Cells & When 3D label is not available, create pseudo-3D   synthetic cell data from individually generated 2D slices \\ \hline

Classification & MRI & \cite{Dikici2021} & cGANe (DCGAN based) & Brain & Provide a novel data-sharing protocol \\ \hline

Classification & MRI & \cite{Momeni2021} & conditional LesionGAN (cGAN based) & Brain & It can be applied on unseen dataset with different   MRI parameters and diseases \\ \hline

Classification & MRI & \cite{Jung202079} & cGAN & Brain & The classification task improved \\ \hline

Classification & MRI & \cite{Zhuang2019} & ICW-GAN & Brain & It showed that not all data augmentation methods are equally beneficial \\ \hline

Denoising & OCT & \cite{Li2020}  & GAN (N/D) & Brain & This provided more data for noise reduction training \\ \hline

\begin{tabular}[c]{@{}l@{}}Image \\ translation\end{tabular} & CBCT to CT & \cite{Qin2021S98} & GAN (Residual-UNet/CNN) & Lung & Investigated the dose validation accuracy \\ \hline

\begin{tabular}[c]{@{}l@{}}Image \\ translation\end{tabular} & CBCT to CT (corrected CBCT) & \cite{Harms2020} & res-cycle GAN & Brain, Pelvis & CBCT allows for daily 3D imaging, for enhanced image-guided radiation therapy \\ \hline

\begin{tabular}[c]{@{}l@{}}Image \\ translation\end{tabular} & CT to CBCT & \cite{Wei2020} & cGAN & Lung & Handle the discrepancy between the DRR from CT and   an x-ray projection \\ \hline

\begin{tabular}[c]{@{}l@{}}Image \\ translation\end{tabular} & CT and Low count PET to full count PET & \cite{Lei2020b} & CycleGAN & Whole Body & This technique could be a useful tool for low dose   PET imaging \\ \hline

\begin{tabular}[c]{@{}l@{}}Image \\ translation\end{tabular} & CT to MRI & \cite{Yang2021415} & CAE-ACGAN & Brain & Multi-contrast MR synthesis \\ \hline

\begin{tabular}[c]{@{}l@{}}Image \\ translation\end{tabular} & CT to MRI and vice versa & \cite{Gu2021} & Dual3D\& PatchGAN (3DGAN based) & Brain & Creation of a model using transfer learning approach \\ \hline

\end{tabular}
\end{center}
\end{small}
\end{table*}

\setcounter{table}{0} 

\begin{table*}[!t]
\begin{small}

\begin{center}
\caption{Continuation of the previous Table \ref{tab:table5} - Compact overview of reviewed medical papers sorted by application, then by modality and year. The application, the studied organ and the highlights/notes are the main purpose of this table}

\begin{tabular}{|p{2.2cm}|p{1.7cm}|p{2.4cm}|p{2.5cm}|p{1.5cm}|p{6.5cm}|}
\hline

\textbf{Application} & \textbf{Modality} & \textbf{Study} & \textbf{Network} & \textbf{Organ} & \textbf{Notes} \\ \hline

\begin{tabular}[c]{@{}l@{}}Image \\ translation\end{tabular} & CT to MRI & \cite{Lei2020a} & CycleGAN & Prostate, Bladder, Rectum & Facilitate routine prostate-cancer radiotherapy treatment planning \\ \hline

\begin{tabular}[c]{@{}l@{}}Image \\ translation\end{tabular} & CT to MRI and vice versa & \cite{Cai2019174} & CycleGAN (cGAN/Patch-GAN) & Heart, Pancreas, Breast & Cross-modality to improve segmentation of multiple   data sources in an online manner \\ \hline

\begin{tabular}[c]{@{}l@{}}Image \\ translation\end{tabular} & CT to MRI and vice versa & \cite{Zhang20189242, Zhang2019183} & CycleGAN & Heart & Multiclass segmentation using online synthetic data   generation \\ \hline

\begin{tabular}[c]{@{}l@{}}Image \\ translation\end{tabular} & MRI to CT & \cite{Schaefferkoetter20213817} & CycleGAN (Residual-UNet/patchGAN) & Whole Body & The synthetic CT data used for PET attenuation   correction \\ \hline

Image translation & MRI to CT & \cite{Liu2020} & dense CycleGAN & Liver & Use MRI-only photon or proton radiotherapy treatment   planning \\ \hline

Image translation & MRI to CT & \cite{Zeng2019759} & hGAN (CycleGAN   based) & Brain & Less data and computational resources needed due to   hybrid approach \\ \hline

Image translation & MRI to PET & \cite{Hu2022145} & BMGAN(Dense-UNet/Patch-Level) & Brain & Preserves the diverse brain structure details \\ \hline

Image translation & MRI to PET and vice versa & \cite{Lin2021} & RevGAN & Brain & It uses only one bidirectional generator \\ \hline

Image translation & MRI to PET & \cite{Pan2019137} & FGAN & Brain & Achieves the state-of-the-art performance in AD identification   and MCI conversion prediction \\ \hline

Image translation & MRI to PET & \cite{Yan2018} & cGAN & Brain & The MCI progression classification improved 7\% \\ \hline

Nuclei counting & Microscopy & \cite{Han2019} & SpCycleGAN & Cells (Kidney) & The method is capable of counting nuclei in 3D \\ \hline

\begin{tabular}[c]{@{}l@{}} Reconstruction \\ (2D -\textgreater 3D) \end{tabular} & Microscopy & \cite{Gupta2021759} & CryoGAN (cryo-EM physics   simulator) & Biomolecule & CryoGAN currently achieves a 10.8{\AA} resolution on a realistic synthetic dataset \\ \hline

\begin{tabular}[c]{@{}l@{}}Reconstruction \\ (2D -\textgreater 3D) \end{tabular} & Synthetic & \cite{Yang2021130} & CycleGAN based (CNN based/multi-scale) & Cells (Embryos) & 3D multi-view deconvolution and fusion using semi- and self-supervised networks \\ \hline

\begin{tabular}[c]{@{}l@{}} Reconstruction \\ (Low -\textgreater High) \end{tabular} & CT (CTP) & \cite{Moghari2019} & cGAN & Brain & This   method could allow dose reduction in CT Perfusion \\ \hline

Segmentation & CT & \cite{Zhang2021} & PGGAN & Pelvis & Semi-supervised learning for semantic segmentation \\ \hline

Segmentation & Microscopy & \cite{Chen2021961} & SpCycleGAN & Cells (Kidney) & The non-ellipsoidal nuclei approach achieves   improved segmentation on volumes with irregularly shaped nuclei \\ \hline

Segmentation & Microscopy & \cite{Ho2019} & SpCycleGAN & Cells (Kidney) & It can segment nuclei visually and numerically \\ \hline

\end{tabular}
\end{center}
\end{small}
\end{table*}

\setcounter{table}{0} 

\begin{table*}[!t]
\begin{small}

\begin{center}
\caption{Continuation of the previous Table \ref{tab:table5} - Compact overview of reviewed medical papers sorted by application, then by modality and year. The application, the studied organ and the highlights/notes are the main purpose of this table}

\begin{tabular}{|p{2.2cm}|p{1.7cm}|p{2.4cm}|p{2.5cm}|p{1.5cm}|p{6.5cm}|}
\hline

\textbf{Application} & \textbf{Modality} & \textbf{Study} & \textbf{Network} & \textbf{Organ} & \textbf{Notes} \\ \hline

Segmentation & Microscopy & \cite{Tang20201775} & cGAN & Cells (Neuron) & It improves the performance of neuron segmentation \\ \hline

Segmentation & MRI & \cite{Ma2020} & da-GAN & Brain & It can improve Hippocampal subfields segmentation accuracy \\ \hline

Segmentation & MRI and CT & \cite{Yang2019181} & N/D (3D-Unet/3D CNN) & Heart & Helps to achieve better understanding cardiovascular motion \\ \hline

General & CT & \cite{Xu201962} & MCGAN & Lung & More vigorous study is needed to verify \\ \hline

General & MRI & \cite{Rusak202011} & GAN (Pix2Pix based) & Brain & Generate brain MRI with accurate borders between   tissue \\ \hline

General & MRI & \cite{Baumgartner20188309} & VA-GAN (WGAN based 3D U-Net / C3D) & Brain & It can capture multiple regions affected by disease \\ \hline

General & Synthetic & \cite{Danu2019662} & GAN   (N/D) & Cells (Blood vessels) & The synthetic blood vessels are indistinguishable from the real ones \\ \hline

\end{tabular}
\end{center}
\end{small}
\end{table*}

\begin{table*}[!t]
\begin{small}

\begin{center}
\caption{Compact overview of reviewed non-medical papers sorted by application, then by modality and year. The application and the studied structure are the main purpose of this table}
\label{tab:table6}

\begin{tabular}{|p{3.8cm}|p{2cm}|p{3cm}|p{4.5cm}|p{2.8cm}|}
\hline

\textbf{Application} & \textbf{Modality} & \textbf{Study} & \textbf{Network} & \textbf{Structure} \\ \hline

Classification & CAD & \cite{Muzahid202120} & Progressive Conditional GAN (PGGAN/cGAN based) & Diversified objects \\ \hline

Classification & CAD & \cite{greminger2020generative} & MSG-GAN & Models to be manufactured \\ \hline

Classification & Seismic reflection   data & \cite{Liu2020O47} & semi-supervised GANs & Seismic facies \\ \hline

Reconstruction (2D -\textgreater 3D) & CAD & \cite{Kniaz20203} & Z-GAN (pix2pix based) & Cars \\ \hline

Reconstruction (2D -\textgreater 3D) & CAD & \cite{Li20195530} & MP-GAN & Diversified objects \\ \hline

Reconstruction (2D -\textgreater 3D) & CAD & \cite{Yang2017679} & 3D-RecGAN & Diversified objects \\ \hline

Reconstruction (2D -\textgreater 3D) & CT & \cite{Krutko2019} & SPGAN (3D DCGAN based) & Porous Media \\ \hline

Reconstruction (2D -\textgreater 3D) & FIB-SEM & \cite{Sciazko20211363} & GAN2Dto3D & Electrode \\ \hline

Reconstruction (2D -\textgreater 3D) & RGB-D & \cite{Liu20212843} & DLGAN (EDGAN) & Diversified objects \\ \hline

Reconstruction (2D -\textgreater 3D) & Synthetic & \cite{Nozawa2021} & cGAN & Cars \\ \hline

Reconstruction (2D -\textgreater 3D) & Synthetic & \cite{Shen20213250} & WGAN-GP & Hair \\ \hline

Reconstruction (2D -\textgreater 3D) & Synthetic, x-ray,   KPFM, SEM & \cite{Kench2021299} & SliceGAN & Electrode \\ \hline

Reconstruction  (Low -\textgreater High) & CAD, RGB-D & \cite{Wang20172317} & ED-GAN (cGAN   based) & Diversified objects \\ \hline

Reconstruction  (Low -\textgreater High) & Synthetic & \cite{Halpert20192081} & DCGAN & Seismic \\ \hline

General & CT & \cite{Chen2020} & WGAN with two discriminators & Porous Media \\ \hline

General & CT (XCT) & \cite{Gayon-Lombardo2020} & DCGAN & Electrode \\ \hline

General & CT & \cite{Liu20196164} & DCGAN & Porous Media \\ \hline

General & Micro-CT & \cite{Mosser2017} & DCGAN & Porous Media \\ \hline

\end{tabular}
\end{center}
\end{small}
\end{table*}

\twocolumn

\section{Appendix - Supplementary figures}
\label{app:Supplementary_figures}

\setcounter{figure}{0}
\renewcommand{\thefigure}{\Alph{section}.\arabic{figure}}

\begin{figure}[h!]
  \centering
  \includegraphics[width=1\columnwidth]{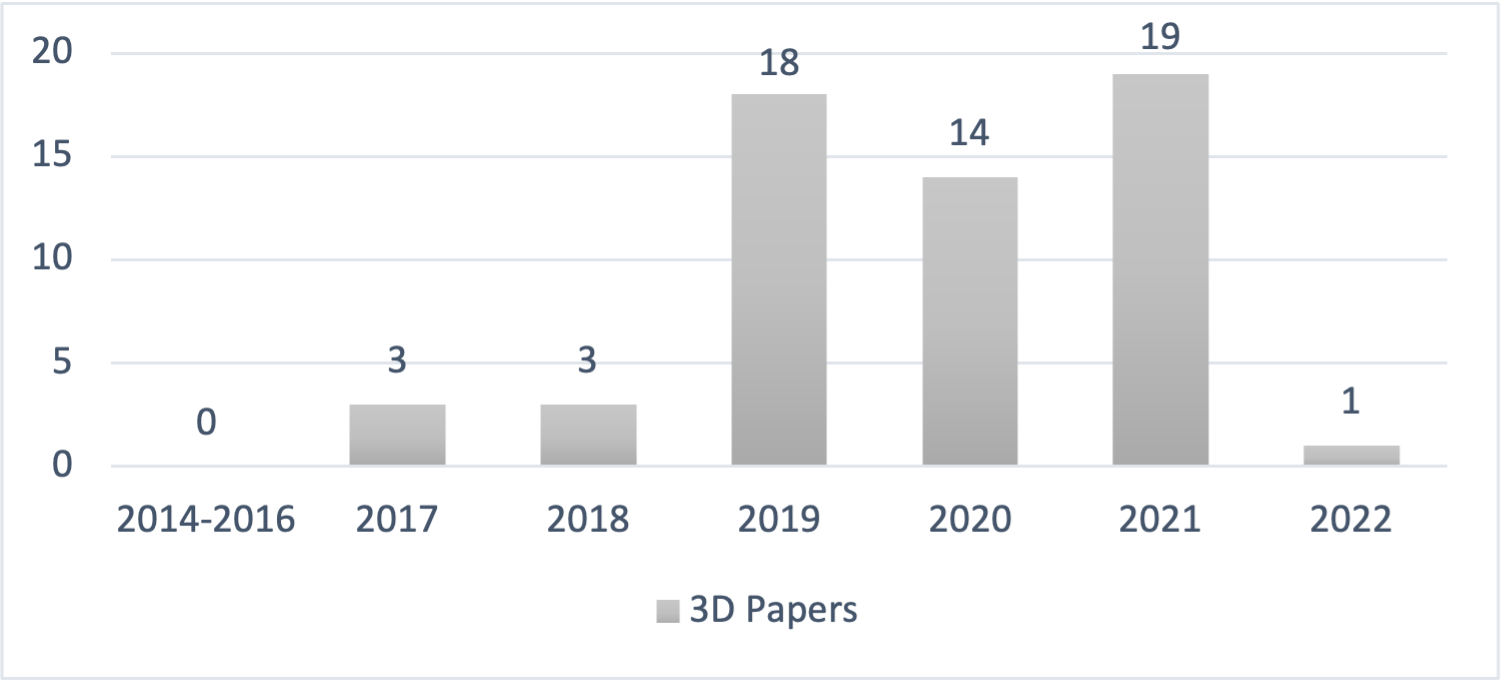}
  \caption{Number of papers per year.}
  \label{Fig:year_all_3d}
\end{figure}%

\begin{figure}[h!]
\centering
\includegraphics[width=1\linewidth]{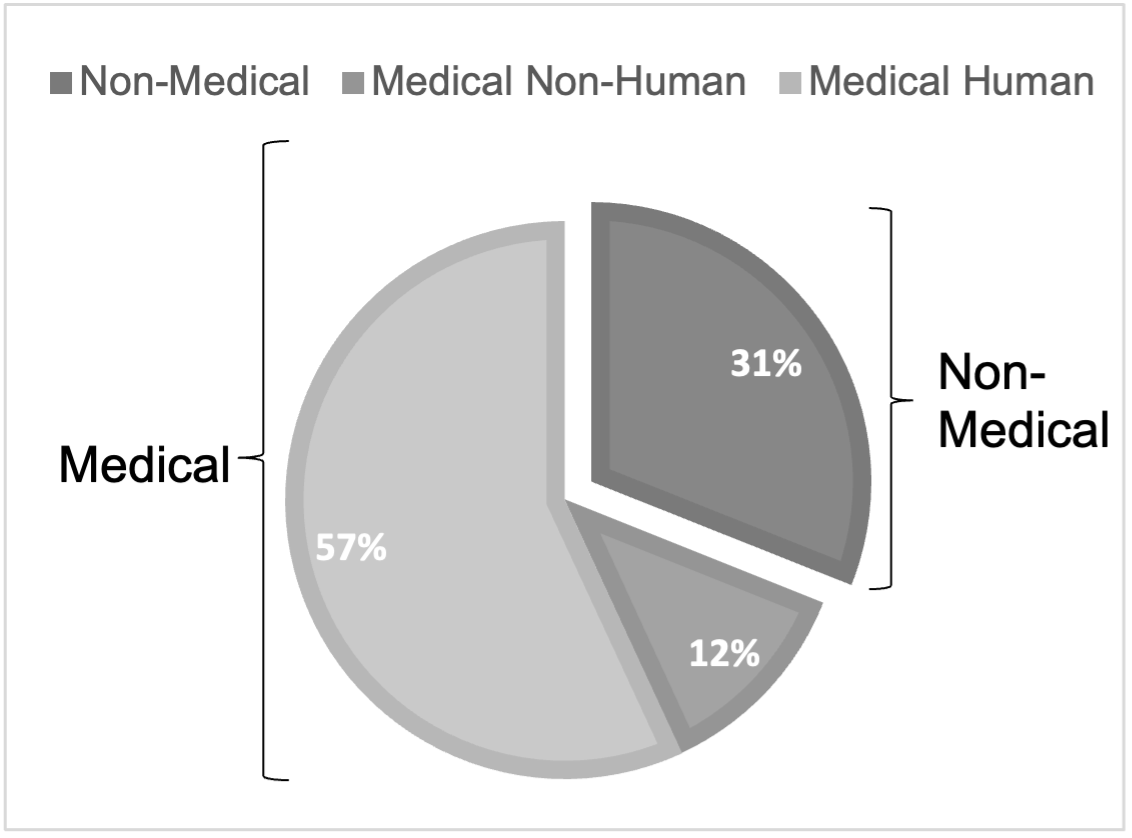}
\caption{Number of papers on volumetric data with medical and non-medical application. The medical papers are divided into human and non-human applications.}
  \label{Fig:Medical_3D_non_human}
\end{figure}

\begin{figure}
    \centering
    \includegraphics[width=1\columnwidth]{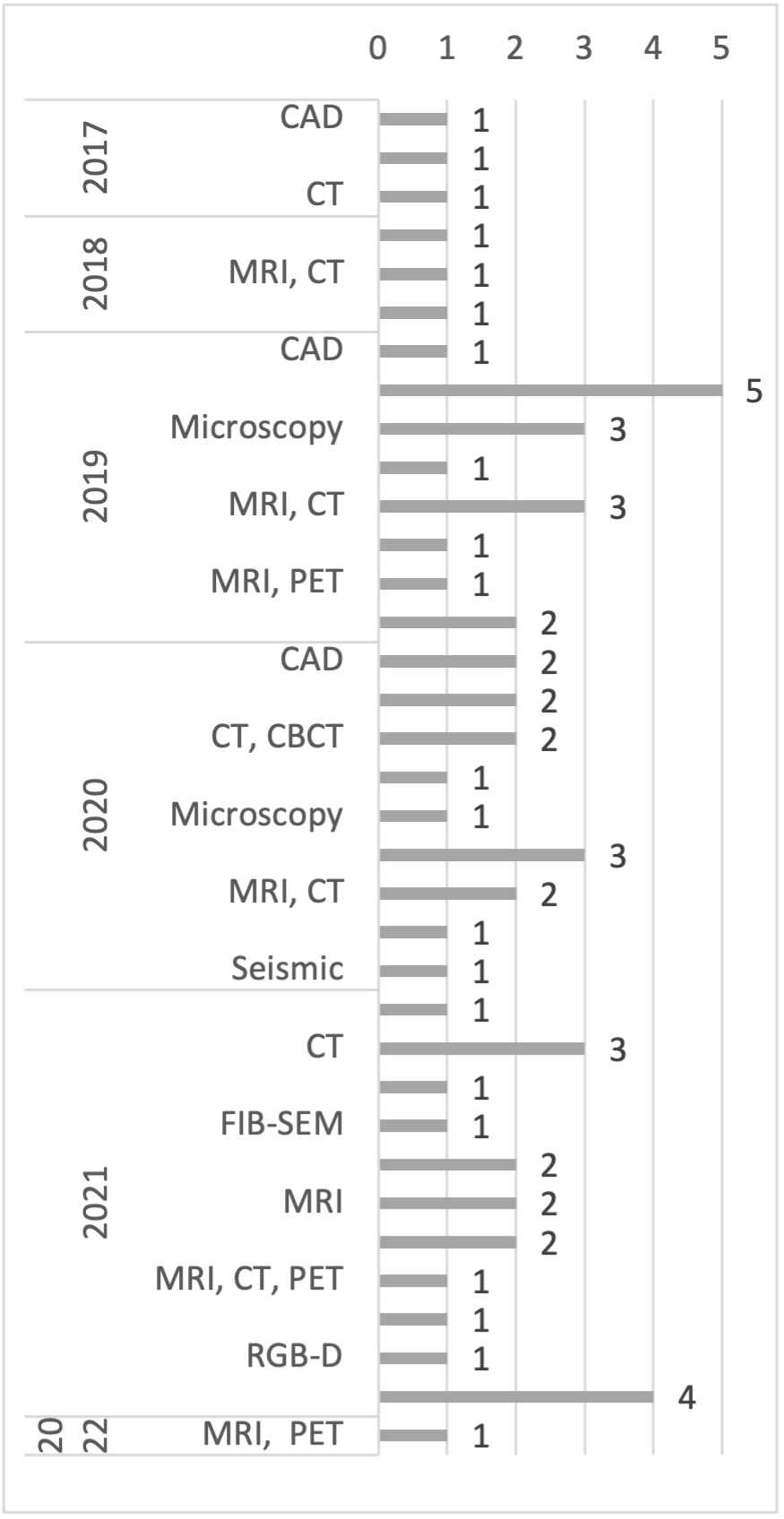}
    \caption{Number of papers per modality per year.}
\label{Fig:Modalities_per_year}
\end{figure}

\bibliographystyle{apacite}
\bibliography{bib}
\end{document}